\documentclass{article}

\usepackage{microtype}
\usepackage{graphicx}
\usepackage{subfigure}
\usepackage{booktabs} 

\usepackage{hyperref}

 \usepackage[accepted]{icml2024}

\usepackage{amsmath}
\usepackage{amssymb}
\usepackage{mathtools}
\usepackage{amsthm}

\usepackage{cuted}
\usepackage{bbm}

\usepackage{multicol}
\usepackage{bm}
\usepackage{dsfont}
\usepackage{amsfonts}
\usepackage{enumerate}
\usepackage{hyperref}
\mathtoolsset{showonlyrefs}
\allowdisplaybreaks

\theoremstyle{plain}
\newtheorem{theorem}{Theorem}[section]
\newtheorem{proposition}[theorem]{Proposition}
\newtheorem{lemma}[theorem]{Lemma}
\newtheorem{corollary}[theorem]{Corollary}
\theoremstyle{definition}

\newtheorem{assumption}[theorem]{Assumption}
\theoremstyle{remark}
\newtheorem{remark}[theorem]{Remark}

\usepackage[textsize=tiny]{todonotes}

\newcommand\smallo{
  \mathchoice
    {{\scriptstyle\mathcal{O}}}
    {{\scriptstyle\mathcal{O}}}
    {{\scriptscriptstyle\mathcal{O}}}
    {\scalebox{.7}{$\scriptscriptstyle\mathcal{O}$}}
}

\newcommand{\wtkappa}{\tilde{\kappa}}
\newcommand{\bx}{\bm{x}}
\newcommand{\bxk}[1]{\bm{x}^{(#1)}}
\newcommand{\tbxk}[1]{\widetilde{\bm{x}}^{(#1)}}

\newcommand{\zk}[1]{z^{(k)}}
\newcommand{\muk}[1]{\mu^{(k)}}
\newcommand{\Lk}[1]{L^{(k)}}
\newcommand{\epsilonk}[1]{\epsilon^{(k)}}
\newcommand{\yk}[1]{y^{(#1)}}
\newcommand{\etak}[1]{\eta^{(k)}}
\newcommand{\pk}[1]{p^{(k)}}
\newcommand{\bw}{\bm{w}}
\newcommand{\bwk}[1]{\bm{w}^{(#1)}}
\newcommand{\wk}[1]{w^{(#1)}}
\newcommand{\wks}[1]{w^{(#1)*}}
\newcommand{\bwks}[1]{\bm{w}^{(#1)*}}
\newcommand{\hw}{\widehat{w}}
\newcommand{\hwkt}[2]{\widehat{w}^{(#1)[#2]}}
\newcommand{\hbwkt}[2]{\widehat{\bm{w}}^{(#1)[#2]}}
\newcommand{\btheta}{\bm{\theta}}
\newcommand{\bthetak}[1]{\bm{\theta}^{(#1)}}
\newcommand{\bthetaks}[1]{\bm{\theta}^{(#1)*}}
\newcommand{\htheta}{\widehat{\bm{\theta}}}
\newcommand{\hthetakt}[2]{\widehat{\bm{\theta}}^{(#1)[#2]}}
\newcommand{\tthetakt}[2]{\widetilde{\bm{\theta}}^{(#1)[#2]}}
\newcommand{\qk}[1]{q^{(#1)}}
\newcommand{\Qk}[1]{Q^{(#1)}}
\newcommand{\hQk}[1]{\widehat{Q}^{(#1)}}
\newcommand{\otheta}{\overline{\bm{\theta}}}

\newcommand{\tP}{\mathbb{P}}
\newcommand{\tE}{\mathbb{E}}

\newcommand{\lambdat}[1]{\lambda^{[#1]}}
\newcommand{\ind}{\perp\!\!\!\!\perp}

\newcommand{\infnorma}[1]{\left\|#1\right\|_{\infty}}

\newcommand{\twonorma}[1]{\left\|#1\right\|_{2}}

\newcommand{\twonorm}[1]{\|#1\|_{2}}

\newcommand{\norma}[1]{\left|#1\right|}
\newcommand{\norm}[1]{|#1|}
\newcommand{\lambdamax}{\lambda_{\textup{max}}}
\newcommand{\lambdamin}{\lambda_{\textup{min}}}

\DeclareMathOperator*{\argmin}{arg\,min}

\makeatletter
\newcommand*{\Rom}[1]{\expandafter\@slowromancap\romannumeral #1@}
\makeatother

\newcommand*{\rom}[1]{\romannumeral #1}

\icmltitlerunning{Towards the Theory of Unsupervised Federated Learning: Non-asymptotic Analysis of Federated EM Algorithms}

\begin{document}

\twocolumn[
\icmltitle{Towards the Theory of Unsupervised Federated Learning: Non-asymptotic Analysis of Federated EM Algorithms}



\icmlsetsymbol{equal}{*}

\begin{icmlauthorlist}
\icmlauthor{Ye Tian}{yyy}
\icmlauthor{Haolei Weng}{comp}
\icmlauthor{Yang Feng}{sch}


\end{icmlauthorlist}

\icmlaffiliation{yyy}{Department of Statistics, Columbia University, New York, USA}
\icmlaffiliation{comp}{Department of Statistics and Probability, Michigan State University, East Lansing, USA}
\icmlaffiliation{sch}{Department of Biostatistics, School of Global Public Health, New York University, New York, USA}

\icmlcorrespondingauthor{Yang Feng}{yang.feng@nyu.edu}

\icmlkeywords{Machine Learning, ICML}

\vskip 0.3in
]



\printAffiliationsAndNotice{}  

\begin{abstract}
While supervised federated learning approaches have enjoyed significant success, the domain of unsupervised federated learning remains relatively underexplored. Several federated EM algorithms have gained popularity in practice, however, their theoretical foundations are often lacking. In this paper, we first introduce a federated gradient EM algorithm (FedGrEM) designed for the unsupervised learning of mixture models, which supplements the existing federated EM algorithms by considering task heterogeneity and potential adversarial attacks. We present a comprehensive finite-sample theory that holds for general mixture models, then apply this general theory on specific statistical models to characterize the explicit estimation error of model parameters and mixture proportions. Our theory elucidates when and how FedGrEM outperforms local single-task learning with insights extending to existing federated EM algorithms. This bridges the gap between their practical success and theoretical understanding. Our numerical results validate our theory, and demonstrate FedGrEM's superiority over existing unsupervised federated learning benchmarks.
\end{abstract}

\section{Introduction}
Federated learning (FDL) is a machine learning paradigm that allows the training of statistical models by leveraging data from various local tasks, while ensuring the data remains decentralized to protect privacy \citep{li2020federatedsurvey}. Introduced a few years ago, notably by Google \citep{konevcny2016federated, mcmahan2017communication}, FDL has witnessed remarkable success in a diverse range of applications, including smartphones \citep{hard2018federated}, healthcare \citep{antunes2022federated}, and the internet of things \citep{nguyen2021federated}. However, it is important to note that a large portion of current FDL research is centered around supervised learning problems. In this paper, we delve into the realm of unsupervised FDL, a scenario in which each task involves a mixture of distributions.

Before proceeding further, we summarize the mathematical notations used in this paper here. $\tP$ and $\tE$ denote the probability and expectation, respectively. For two positive sequences $\{a_n\}$ and $\{b_n\}$, $a_n \ll b_n$ or $a_n = \smallo(b_n)$ means $a_n/b_n \rightarrow 0$, $a_n \lesssim b_n$ or $a_n = \mathcal{O}(b_n)$ means $a_n/b_n \leq C < \infty$, and $a_n \asymp b_n$ means $a_n/b_n, b_n/a_n \leq C< \infty$. $\widetilde{\mathcal{O}}(b_n)$ is the same as $a_n = \mathcal{O}(b_n)$ up to logarithmic factors. For a random variable $x_n$ and a positive sequence $a_n$, $x_n = \mathcal{O}_{\tP}(a_n)$ means that for any $\epsilon > 0$, there exists $M > 0$ such that $\sup_{n}\tP(|x_n/a_n|>M) \leq \epsilon$. $\widetilde{\mathcal{O}}_{\tP}(a_n)$ has a similar meaning up to logarithmic factors in $a_n$. For a vector $\bx \in \mathbb{R}^d$, $\twonorm{x}$ represents its Euclidean norm. For two numbers $a$ and $b$, $a \vee b = \max\{a,b\}$ and $a \wedge b = \min\{a,b\}$. For any positive integer $K$, $1:K$ and $[K]$ stand for the set $\{1, 2, \ldots, K\}$. And for any set $S$, $|S|$ denotes its cardinality and $S^c$ denotes its complement. ``w.p." stands for ``with probability". The absolute constants $c$ and $C$ may vary from line to line.

\section{Federated Learning on Mixture of Distributions}

\subsection{Problem Setting}\label{subsec: setting}
Consider $K$ tasks, where for the $k$-th task, we observe data $\{\bxk{k}_i\}_{i=1}^n \subseteq \mathbb{R}^d$ \footnote{For simplicity, we assume all tasks share the same sample size $n$.  We can easily extend our analysis to the case of heterogeneous task sample sizes with similar theoretical results.}. There exists an \textit{unknown} subset $S \subseteq [K]$, such that each observation in task $k \in S$ comes from a \textit{mixture model} with $R$ components ($R \geq 2$):
\begin{equation}\label{eq: generic model}
	\bxk{k}_i \overset{i.i.d.}{\sim} \sum_{r=1}^R \wks{k}_r\cdot \pk{k}_r(\,\bm{\cdot}\,\,;\bthetaks{k}_r),
\end{equation}
where the \textit{mixture proportion} $\{\wks{k}_r\}_{r=1}^R \subseteq (0, 1)$ with $\sum_{r=1}^R \wks{k}_r = 1$ and $\pk{k}_r(\,\bm{\cdot}\,\,;\bthetaks{k}_r)$ is a Radon-Nikodym density w.r.t. a base measure $\sigma$. $\bthetaks{k}_r \in \mathbb{R}^d$ \footnote{For simplicity, we assume parameters and observations are of the same dimension, but our results can be generalized to the case where the two dimensions are different.} are the parameters that index the distribution $\pk{k}_r$. This is equivalent to
\begin{equation}\label{eq: generic model 2}
	\zk{k}_i \overset{i.i.d.}{\sim} \sum_{r=1}^R \wks{k}_r\cdot \delta_r, \quad \bxk{k}_i|\zk{k}_i = r \overset{i.i.d.}{\sim} \pk{k}_r(\,\bm{\cdot}\,\,;\bthetaks{k}_r),
\end{equation}
for $k \in S$, where $\zk{k}_i$ is the \textit{unobserved} latent cluster label and $\delta_r$ is the point mass at $r$. Here $S$ is the index of similar tasks (unknown), where the parameters $\{\bthetaks{k}_r\}_{k\in S}$ of different tasks are ``similar" to each other, in the sense that
\begin{equation}
	\min_{\otheta \in \mathbb{R}^d}\max_{k \in S}\twonorm{\bthetaks{k}_r - \otheta} \leq h, \quad \forall r \in [R],
\end{equation}
where $h$ is an \textit{unknown} parameter controlling the task similarity level. A small $h$ implies that the tasks are more similar. The data from tasks in set $S^c = [K]\backslash S$ can be \textit{arbitrarily distributed}, i.e., $\{\bxk{k}_i\}_{i\in [n], k \in S^c}$ follow an \textit{arbitrary} joint distribution $\mathbb{Q}_{S^c}$, and we denote the proportion $\epsilon \coloneqq |S^c|/K \in [0, 1)$. Note that $h$, $K$, $R$ and $d$ can change with single-task sample size $n$. 

There are two different interpretations of this setting. The first one is from the perspective of \textit{adversarial attacks/contaminations}, where there is an \textit{adversarial attacker} who can arbitrarily contaminate the data of tasks in an index set $S^c$. The index set $S^c$ and the distribution $\mathbb{Q}_{S^c}$ are picked by the attacker \textit{after} we pick the estimator (hence $S^c$ and $\mathbb{Q}_{S^c}$ are \textit{unknown} to us). A similar setting can be found in \citet{qiao2018outliers},  \citet{konstantinov2019robust}, \citet{konstantinov2020sample}, and \citet{tian2023learning}. 

Alternatively, aside from adversarial attacks, we can interpret the presence of contaminated data sets as a result of \textit{outlier tasks}. In the era of big data, certain collected data sets may exhibit distributions significantly different from others, particularly when dealing with numerous tasks \citep{zhang2021survey}. These data sets within $S^c$ can be viewed as outlier tasks. In practice, detecting outlier tasks is challenging.

In the rest of this paper, we may take the views of ``adversarial attacks" and ``outlier tasks" interchangeably.

Note that in our unsupervised learning setting, similar to \citet{marfoq2021federated} and \citet{wu2023personalized}, we avoid assuming that the mixture proportions $\{\wks{k}_r\}_{k \in S}$ are similar across tasks, which offers more flexibility in practice.

The goal is to develop an algorithm to estimate the mixture proportions $\{\wks{k}_r\}_{k \in S, r \in [R]}$ and the parameters $\{\bthetaks{k}_r\}_{k \in S, r \in [R]}$ simultaneously, which satisfies the following five desired properties:
\begin{enumerate}[(i)]
	\item \textbf{Adaptability} to unknown similarity level $h$: The algorithm should utilize the data from different sources in an ``optimal" way. When $h$ is small, the output estimator should achieve a better convergence rate than the local estimator (or single-task estimator of each task). When $h$ is large, the output estimator should perform no worse than the local estimator.
	\item \textbf{Robustness} against the adversarial attack on a small fraction of sources: The output estimator should maintain a good performance when the contaminated proportion $\epsilon$ is small.
	\item \textbf{Privacy} for local data: The algorithm should avoid transferring raw data out of each task.
	\item \textbf{Computation efficiency} on local servers: The local computational cost should be low.
	\item \textbf{Communication efficiency} between local and global servers: The communicational cost should be low.
\end{enumerate}

\subsection{Related Works}
\textbf{Federated learning (FDL):}  While there exists an extensive body of literature on FDL, the majority of it centers around the supervised learning paradigm.  Notable frameworks within supervised FDL include COCOA \citep{jaggi2014communication}, MOCHA \citep{smith2017federated}, and FedAvg \citep{mcmahan2017communication}. To accommodate varying task characteristics, FedProx was introduced by \citet{li2020federated}. See \citet{yang2019federated} and \citet{li2020federatedsurvey} for a comprehensive overview of supervised FDL. In contrast to supervised FDL, much less attention has been given to unsupervised FDL with mixture models. \citet{marfoq2021federated} examined a similar FDL problem presented in this paper and introduced a federated EM algorithm without exploring the estimation error of the parameter estimators. \citet{dieuleveut2021federated} also proposed a federated EM algorithm that supports communication compression and partial participation. \citet{wu2023personalized} adapted the EM algorithm to the scenario where predictors are from Gaussian Mixture Models (GMMs) and the regression models can encompass general mixture models. Notably, none of these papers provided finite-sample results for their EM algorithm (which is the key to interpreting their practical successes) nor discussed the adversarial attacks and outlier tasks. Our work complements this line of research by explicitly accommodating outlier tasks, providing comprehensive finite-sample results, and applying the developed theory to illustrative model examples. Our theory illustrates when and how the aforementioned federated EM algorithms \citep{dieuleveut2021federated, marfoq2021federated, wu2023personalized} outperform local single-task learning. Note that there exist works on clustered FDL, wherein tasks are organized into several groups with tasks within each group being identical (task-level mixture) \citep{ghosh2020efficient, kong2020meta, su2022global}. This setting differs from ours, where each task's data originates from a mixture model at the sample level.

\textbf{Multi-task learning (MTL) and transfer learning (TL):} Problems related to federated learning but permitting raw data sharing across tasks include multi-task learning and transfer learning. Analogous to federated learning, a substantial proportion of research in MTL and TL centers on supervised learning. In unsupervised MTL and TL, there have been diverse approaches, including the kernel $k$-means clustering \citep{gu2011learning}, the spectral method \citep{yang2014multitask}, and the penalized optimization \citep{dai2008self, zhang2011multitask, zhang2015smart, zhang2018multi}. Specific mixture models such as Gaussian Mixture Models (GMMs) have also been explored \citep{wang2021general, tian2022unsupervised}. Discussions on outlier tasks, adversarial attacks, and negative transfer in MTL and TL have emerged in various model settings, for example, \citet{qiao2018outliers, konstantinov2019robust, hanneke2020value, konstantinov2020sample, li2021transfer, duan2022adaptive, tian2023learning}.

\textbf{EM algorithm:} The EM algorithm was formalized by \citet{dempster1977maximum}, and there have been intensive studies on the local convergence of the likelihood and the estimator to some stationary point \citep{wu1983convergence, redner1984mixture, meng1994global, mclachlan2007algorithm}. More recently, \citet{xu2016global} established the global convergence of EM algorithm on binary GMMs. Additionally, \citet{balakrishnan2017statistical}, \citet{yan2017convergence}, \citet{cai2019chime}, \citet{kwon2020algorithm}, \citet{kwon2020converges}, and \citet{zhao2020statistical} provided finite-sample convergence results for EM and its variants under certain initialization conditions.

\subsection{A Federated Gradient EM: FedGrEM}

\begin{algorithm*}[tb]
   \caption{FedGrEM: A Federated Gradient EM Algorithm}
   \label{algo: FG-EM}
\begin{algorithmic}
   \STATE {\bfseries Input:} Initializations $\{\hbwkt{k}{0}\}_{k \in [K]}$ and $\{\hthetakt{k}{0}\}_{k \in [K]}$ ($\hbwkt{k}{0} = \{\hwkt{k}{0}_r\}_{r=1}^R$, $\hthetakt{k}{0} = \{\hthetakt{k}{0}_r\}_{r=1}^R$), data $\{\bxk{k}_i\}_{i \in [n], k \in [K]}$, iteration number $T$, penalty parameters $\{\lambdat{t}\}_{t=1}^T$, step sizes $\{\etak{k}_r\}_{k\in [K], r\in [R]}$
   \FOR{$t=1$ {\bfseries to} $T$}
   \STATE \underline{Local update:} For task $k = 1:K$: 
   \begin{itemize}
	\item \underline{E-step:} $\hQk{k}(\btheta|\hbwkt{k}{t-1}, \hthetakt{k}{t-1}) \coloneqq \frac{1}{n}\sum_{i=1}^{n}\sum_{r=1}^R \tP(\zk{k} = r|\bxk{k}_i; \hbwkt{k}{t-1}, \hthetakt{k}{t-1}) \log \pk{k}_r(\bxk{k}_i;\btheta_r)$\\
	\item \underline{M-step:} \,\, $\diamond$ $\hwkt{k}{t}_r = \frac{1}{n}\sum_{i=1}^{n}\tP(\zk{k}=r|\bxk{k}_i; \hbwkt{k}{t-1}, \hthetakt{k}{t-1})$
	
	 \hspace{1.52cm}$\diamond$ $\tthetakt{k}{t}_r = \hthetakt{k}{t-1}_r + \eta_r^{(k)} \cdot \frac{\partial}{\partial \btheta_r}\hQk{k}(\btheta|\hbwkt{k}{t-1}, \hthetakt{k}{t-1})|_{\btheta = \hthetakt{k}{t-1}}$ 
	\end{itemize}
	\STATE \underline{Central update:} $\{\hthetakt{k}{t}_r\}_{k=1}^K, \otheta^{[t]}_r = \argmin\limits_{\{\bm{\nu}^{(k)}\}_{k=1}^K \subseteq \mathbb{R}^d, \overline{\bm{\nu}} \in \mathbb{R}^d}\Big\{\sum_{k=1}^K\big(\frac{n}{2}\twonorm{\bm{\nu}^{(k)} - \tthetakt{k}{t}_r}^2+\sqrt{n}\lambdat{t}\cdot \twonorm{\bm{\nu}^{(k)}-\overline{\bm{\nu}}}\big)\Big\}$, define $\hbwkt{k}{t} = \{\hwkt{k}{t}_r\}_{r=1}^R$, $\hthetakt{k}{t} = \{\hthetakt{k}{t}_r\}_{r=1}^R$ \\
   \ENDFOR
   \STATE {\bfseries Output:} Final estimators $\{\hwkt{k}{T}_r\}_{k \in [K], r \in [R]}$ and $\{\hthetakt{k}{T}_r\}_{k \in [K], r \in [R]}$
\end{algorithmic}
\end{algorithm*}

Before delving into our main algorithm, we first introduce some key notations and definitions. The posterior, i.e. the probability of $\zk{k} = r$ conditioned on the observation $\bxk{k}$, given that $(\bxk{k}, \zk{k})$ is from the mixture model \eqref{eq: generic model 2} with parameters $\bwk{k} = \{\wk{k}_r\}_{r=1}^R$ and $\bthetak{k} = \{\bthetak{k}_r\}_{r=1}^R$, is defined as
\begin{align}
	&\tP(\zk{k} = r|\bxk{k}; \bwk{k}, \bthetak{k}) \\
	&= \frac{\wk{k}_r \times  \pk{k}_r(\bxk{k};\bthetak{k}_r)}{\sum_{r=1}^R \wk{k}_r\times \pk{k}_r(\bxk{k};\bthetak{k}_r)}.
\end{align}

Based on this posterior, we define 
\begin{align}
	\Qk{k}(\btheta|\bw', \btheta') &= \tE\Bigg[\sum_{r=1}^R \tP(\zk{k} = r|\bxk{k}; \bw', \btheta') \\
	&\hspace{2cm}\times \log \pk{k}_r(\bxk{k};\btheta_r)\Bigg],
\end{align}
where $\btheta = \{\btheta_r\}_{r=1}^R$, $\bw' = \{w'_r\}_{r=1}^R$, and $\btheta' = \{\btheta'_r\}_{r=1}^R$. By Jensen's inequality, it can be shown that $\Qk{k}(\btheta|\bw', \btheta')$ is a lower bound of the complete population-level log-likelihood $\tE\log \big[\sum_{r=1}^Rw_r\pk{k}_r(\bxk{k};\btheta_r)\big]$. The latter one is difficult to manage as the summation is within the logarithm. The EM algorithm endeavors to maximize the surrogate $\Qk{k}(\btheta|\bw', \btheta')$ by iteratively alternating with the E-step and M-step. Gradient EM does a one-step gradient ascent in M-step instead of finding the exact optimizer, which can speed up the computation. In practice, we work on a sample-based variant of $\Qk{k}(\btheta|\bw', \btheta')$ as
\begin{align}
	&\hQk{k}\big(\btheta|\bw', \btheta'\big) \\
	&= \frac{1}{n}\sum_{i=1}^{n}\sum_{r=1}^R \tP(\zk{k} = r|\bxk{k}_i; \bw', \btheta') \times \log \pk{k}_r(\bxk{k}_i;\btheta_r).
\end{align}
\vspace{-0.8cm}

\begin{figure}[!h]
	\centering
	\includegraphics[width=0.5\textwidth]{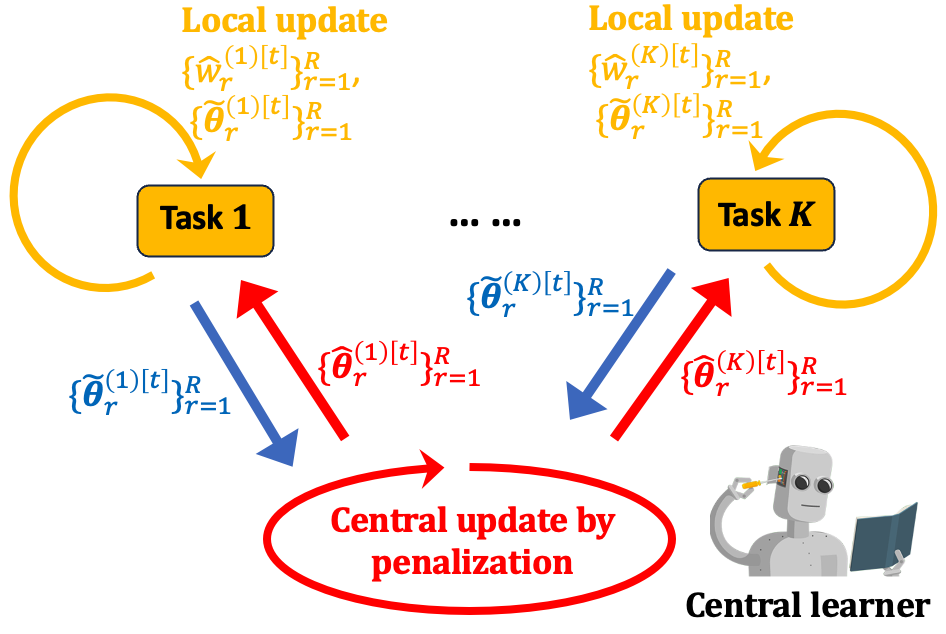}
	\caption{An illustration of Algorithm \ref{algo: FG-EM} (the iteration round $t$).}
	\label{fig: fed-alg-illustration}
\end{figure}

Now, we are ready to introduce our core algorithm \textit{FedGrEM} in Algorithm \ref{algo: FG-EM}. It executes the E-step and M-step locally on each task,  pooling the estimators of $\{\bthetaks{k}_r\}_{k =1}^K$ obtained in M-step by penalizing the $\ell_2$-distance between each estimator and a common center.  This approach mirrors the penalization strategy used in other MTL and TL literature, such as \citet{evgeniou2004regularized, li2007bayesian, lounici2011oracle, solnon2012multi, jalali2013dirty, kuzborskij2013stability, kuzborskij2017fast, denevi2018learning, t2020personalized, li2021transfer, tian2023transfer, he2024transfusion, lin2024hypothesis, lin2024smoothness}. After several iterations of local and central updates, FedGrEM yields the final estimators. Figure \ref{fig: fed-alg-illustration} provides an intuitive illustration of the workflow of FedGrEM.

While the core idea of FedGrEM is akin to FedEM in \citet{marfoq2021federated} and FedGMM in \citet{wu2023personalized}, two crucial distinctions set them apart. First, we employ gradient EM, while FedEM and FedGMM use the full EM. Second, the central update of FedGrEM can adapt to the \textit{heterogeneity} of $\bthetaks{k}_r$'s across tasks and remain \textit{robust} when a small proportion of tasks is contaminated, which are not present in FedEM or FedGMM. On the other hand, by setting $\lambdat{t} = +\infty$, FedGrEM can be viewed as a simplified gradient version of FedEM and FedGMM, therefore our non-asymptotic theory in Section \ref{sec: theory} can illustrate the empirical success achieved by FedEM and FedGMM. 

FedGrEM is \textit{computationally efficient} on local servers as it computes the gradient instead of explicitly solving the maximizer of $\hQk{k}$. It is also \textit{communicationally efficient} because it only necessitates the exchange of parameter estimators between local and central servers. Therefore, FedGrEM fulfills \textit{all five of the desired properties} we outlined in Section \ref{subsec: setting}.

\subsection{Our Contributions}\label{subsec: contributions}
Firstly, we introduced FedGrEM, a federated EM algorithm that exhibits robustness against a small number of adversarially contaminated tasks while maintaining computational and communication efficiency. FedGrEM supplements the existing federated EM algorithms in literature \citep{dieuleveut2021federated, marfoq2021federated, wu2023personalized} by considering the heterogeneity across tasks and adversarial contaminations.

Second, we provided an extensive non-asymptotic theory for FedGrEM on general mixture models. We characterized the estimation error of $\wks{k}_r$ and $\bthetaks{k}_r$ for non-outlier tasks by five main components:

\begin{itemize}
\itemsep0em
	\item Iterative error, which vanishes as the number of iterations goes to infinity;
	\item Aggregation rate, which depends on the combined sample sizes of non-outlier tasks;
	\item Cost of heterogeneous mixture proportions;
	\item Cost of task heterogeneity;
	\item Cost of outlier tasks.
\end{itemize}

This analysis revealed that when the tasks exhibit sufficient similarity and the proportion of outlier tasks is sufficiently small, the estimation error of FedGrEM surpasses the rate achieved by typical single-task algorithms such as the local single-task EM. Since the setting of existing federated EM papers can be seen as a special case (no model heterogeneity and contaminations) of the scenario we study, our theory helps illustrate the empirical success of existing federated EM algorithms \citep{dieuleveut2021federated, marfoq2021federated, wu2023personalized} and offers new theoretical insights for unsupervised federated learning.

Thirdly, we addressed the often overlooked issue of cluster label permutation in federated EMs. While label permutation is not a concern for single-task EM, most federated EM algorithms require all non-outlier tasks to share the same permutation as they take an average over the parameter estimators for the same cluster in the M-step. Failure to align the label permutations across non-outlier tasks can lead to the failure of federated EM algorithms. Due to space limit, we leave this part to Section \ref{sec: label permutation} of the appendix.

\section{Theory}\label{sec: theory}
In this section, we introduce a generic non-asymptotic upper bound for the estimation error of FedGrEM and apply this theory to two statistical examples: Gaussian Mixture Models (GMMs) and Mixture of Regressions (MoRs). Our theoretical findings offer a clear interpretation, shedding light on the conditions under which existing federated EM algorithms, including FedGrEM, can outperform local single-task learning. To streamline the presentation, we provide simplified versions of most theoretical results here, with formal details available in Section \ref{sec: formal theory appendix} of the appendix.

\subsection{Generic Analysis}
For simplicity, denote $\qk{k}(\btheta) = \Qk{k}(\btheta|\bwks{k}, \bthetaks{k})$. We first state a few assumptions that are necessary for our results on general mixture models.

\begin{assumption}[Concavity and smoothness, a simplified version of Assumption \ref{asmp: q appendix}]\label{asmp: q}
	For all $k \in S$, there exist non-negative constants $\{\muk{k}_r\}_{r=1}^R$ and $\{\Lk{k}_r\}_{r=1}^R$ such that for all $\btheta = \{\btheta_r\}_{r=1}^R$, $\btheta' = \{\btheta_r'\}_{r=1}^R$:
	\vspace{-0.1cm}
	\begin{enumerate}[(i)]
		\item (Strong concavity) $\qk{k}(\btheta') - \qk{k}(\btheta) - \frac{\partial}{\partial \btheta}\qk{k}(\btheta)^T(\btheta'-\btheta) \leq -\sum_{r=1}^R\frac{\muk{k}_r}{2}\twonorm{\btheta'_r-\btheta_r}^2$;
		\item (Smoothness) $\qk{k}(\btheta') - \qk{k}(\btheta) - \frac{\partial}{\partial \btheta} \qk{k}(\btheta)^T(\btheta'-\btheta) \geq -\sum_{r=1}^R\frac{\Lk{k}_r}{2}\twonorm{\btheta'_r-\btheta_r}^2$.
	\end{enumerate}
\end{assumption}

\begin{remark}
	The same conditions are imposed by \citet{balakrishnan2017statistical} in the single-task setting. The strong concavity is usually assumed to obtain the parametric convergence rate, and the smoothness is imposed for gradient descent to converge at a geometric rate.
\end{remark}

\begin{assumption}[Contraction and convergence, a simplified version of Assumptions \ref{asmp: w appendix} and \ref{asmp: theta appendix}]\label{asmp: w and theta}
	There exist a constant $\kappa \in (0, 1)$, and rate functions $\mathcal{R}_{w}(n)$, $\mathcal{R}_{\btheta}(n)$, such that for any $k \in S$, such that for all $\bw' = \{w_r'\}_{r=1}^R$ and $\btheta' = \{\btheta_r'\}_{r=1}^R$ close to  $\{\wks{k}_r\}_{r=1}^R$ and $\{\wks{k}_r\}_{r=1}^R$:
	\vspace{-0.3cm} 
	\begin{enumerate}[(i)]
		\item (Contraction) 
		\begin{enumerate}
			\item $\big|\tE\big[\tP(\zk{k}=r|\bxk{k};\bw', \btheta')\big]-\wks{k}_r\big| \leq \kappa\cdot \sum_{r=1}^R \big(\norm{w_r' - \wks{k}_r} + \twonorm{\btheta_r' - \bthetaks{k}_r}\big)$;
			\item $\big\|\frac{\partial}{\partial \btheta_r}\qk{k}(\btheta)|_{\btheta = \btheta'} - \frac{\partial}{\partial \btheta_r}\Qk{k}(\btheta|\bw', \btheta')|_{\btheta = \btheta'}\big\|_2 \allowbreak \leq \kappa\cdot \sum_{r=1}^R \big(\norm{w_r' - \wks{k}_r} + \twonorm{\btheta_r' - \bthetaks{k}_r}\big)$
		\end{enumerate} 
		\item (Uniform convergence)  w.p. $1-\smallo(1)$, 
		\begin{enumerate}
			\item $\big|\frac{1}{n}\sum_{i=1}^n\tP(\zk{k}=r|\bxk{k}_i;\bw', \btheta') - \tE\big[\tP(\zk{k}=r|\bxk{k};\bw', \btheta')\big]\big| \leq \mathcal{R}_{w}(n)$;
			\item $\big\|\frac{\partial}{\partial \btheta_r}\hQk{k}(\btheta|\bw', \btheta')|_{\btheta = \btheta'} - \frac{\partial}{\partial \btheta_r}\Qk{k}(\btheta|\bw', \btheta')\allowbreak|_{\btheta = \btheta'}\big\|_2  \leq \mathcal{R}_{\btheta}(n)$;
			\item $\big\|\frac{1}{|S|}\sum_{k \in S}\big[\frac{\partial}{\partial \btheta_r}\hQk{k}(\btheta|\bw', \btheta')|_{\btheta = \btheta'} - \frac{\partial}{\partial \btheta_r}\Qk{k}\allowbreak(\btheta|\bw', \btheta')|_{\btheta = \btheta'}\big]\big\|_2  \leq \mathcal{R}_{\btheta}(nK)$.
		\end{enumerate}
	\end{enumerate}	
\end{assumption}

\vspace*{0cm}

Note that $\mathcal{R}_{w}(n)$ and $\mathcal{R}_{\btheta}(n)$ also depend on other parameters such as $d$ and $R$, which we suppress in the notation for the ease of presentation.


\begin{remark}
	Note that by definition $\wks{k}_r = \tE[\tP(\zk{k}=r|\bxk{k};\bwks{k}, \bthetaks{k})]$. Therefore, condition (\rom{1}) describes the behavior of $\tE\big[\tP(\zk{k}=r|\bxk{k};\bw', \btheta')\big]$ and $\frac{\partial}{\partial \btheta_r}\Qk{k}(\btheta|\bw', \btheta')$, and condition (\rom{2}) is a uniform convergence assumption on the same quantities, when $\bw'$ and $\btheta'$ are close to the true values $\bwks{k}$ and $\bthetaks{k}$. Condition (\rom{1}) has been used by \citet{balakrishnan2017statistical} in the single-task setting. Condition (\rom{2}).(b) and condition (\rom{2}).(c) are uniform convergence assumptions on the gradient around the true parameter values, which are often needed when analyzing the EM without data splitting \citep{yan2017convergence, cai2019chime}. Condition (\rom{2}).(c) is a generalization of (\rom{2}).(b) when aggregating the data from multiple tasks. As we will see in later examples, we usually have $R_w(n) = \widetilde{\mathcal{O}}(R^2\sqrt{1/n})$ and $R_{\btheta}(n) = \widetilde{\mathcal{O}}(R^2\sqrt{d/n})$, and $\mathcal{R}_{\btheta}(n)$ is typically the estimation error of local single-task methods.
\end{remark}

\begin{assumption}[Good initialization and step size, a simplified version of Assumption \ref{asmp: thm generic appendix}]\label{asmp: thm generic}
	$\norm{\hw^{(k)[0]}_r - \wks{k}_r} \vee \twonorm{\htheta^{(k)[0]}_r - \bthetaks{k}_r} \leq C$, $\etak{k}_r \leq 1/L_r^{(k)}$, for all $k \in S$ and $r \in [R]$, where $C > 0$ is a constant whose explicit form can be found in the appendix. 
\end{assumption}

We set the penalty parameters in Algorithm \ref{algo: FG-EM} by induction as 
\vspace{-0.2cm}
\begin{align}
	\lambdat{0} &= C_1\sqrt{n},\\
	\lambdat{t} &= \kappa' \cdot \lambdat{t-1} + C_2\sqrt{n}[\mathcal{R}_{w}(n) + \mathcal{R}_{\btheta}(n)],
\end{align}
\vspace{-0.8cm}

where $t \geq 1$ and the explicit forms of $\kappa'\in (0, 1)$, $C_1$, $C_2$ can be found in the appendix.

Next, we present our primary result for the estimation error of FedGrEM in Theorem \ref{thm: generic}. 

\vspace{-1cm}

\begin{strip}
\begin{theorem}[Main result, a simplified version of Theorem \ref{thm: generic appendix}]\label{thm: generic}
	Suppose Assumptions \ref{asmp: q}, \ref{asmp: w and theta}, and \ref{asmp: thm generic} hold. Then for any contaminated set $S^c$ with $\epsilon=|S^c|/K < 1/3$ and any contamination distribution $\mathbb{Q}_{S^c}$, w.p. $1-\smallo(1)$, for all $T \geq 1$, FedGrEM satisfies
	\begin{align}
		\max_{k \in S, r \in [R]}\big(\norm{\hwkt{k}{T}_r-\wks{k}_r} \vee \twonorm{\hthetakt{k}{T}_r-\bthetaks{k}_r}\big) 
		&\lesssim \underbrace{\kappa_0^T}_{\textup{iterative error}}+ \underbrace{\mathcal{R}_{\btheta}(nK)}_{\textup{aggregation rate}} + \underbrace{\mathcal{R}_w(n)}_{\textup{cost of heterogeneous mixture proportions}} \\
		&\quad + \underbrace{\min\big\{h, \mathcal{R}_w(n) + \mathcal{R}_{\btheta}(n)\big\}}_{\textup{cost of task heterogeneity}} + \underbrace{\epsilon \big[\mathcal{R}_w(n) + \mathcal{R}_{\btheta}(n)\big]}_{\textup{cost of outlier tasks}}, \label{eq: main bound}
	\end{align}
	\vskip-0.4cm
	where $\kappa_0 \in (0, 1)$.
\end{theorem}
\end{strip}

The upper bound of convergence rate in Theorem \ref{thm: generic} comprises multiple terms, each with a clear interpretation. The first term corresponds to the geometric iterative error which vanishes as $T \rightarrow +\infty$, and the second term accounts for the aggregation error which arises from combining all the data. The third to fifth terms represent the cost of heterogeneous mixing proportions, task heterogeneity, and outlier tasks, respectively. 

As we will observe in later specific examples, $\mathcal{R}_{\btheta}(nK)$ scales as $R^2\sqrt{d/(nK)}$ which depends on the total sample size $nK$ of all $K$ tasks, $\mathcal{R}_{\btheta}(n) = \widetilde{\mathcal{O}}(R^2\sqrt{d/n})$, and $\mathcal{R}_{w}(n)  = \widetilde{\mathcal{O}}(R^2\sqrt{1/n})$. Note that $\mathcal{R}_{\btheta}(n)$ typically represents the estimation error of local single-task algorithms. Comparing \eqref{eq: main bound} with $\mathcal{R}_{\btheta}(n) =  \widetilde{\mathcal{O}}(R^2\sqrt{d/n})$, we can see that when both $h$ and $\epsilon$ are small --- indicating sufficient similarity shared across tasks and few contaminated tasks --- FedGrEM can achieve a better estimation error than the local single-task methods. When $h = \epsilon = 0$, implying that all tasks share the same parameters, we revert to the setting of \citet{dieuleveut2021federated}, \citet{marfoq2021federated}, and \citet{wu2023personalized}. In this context, our finite-sample upper bound demonstrates that federated EM can indeed outperform the local single-task methods, aligning with the empirical success of federated EM algorithms observed in these works.
 
In subsequent sections, we will substitute specific rate expressions for each term in concrete examples, by showing that $\mathcal{R}_{w} = \widetilde{\mathcal{O}}(R^2\sqrt{1/n})$ and $\mathcal{R}_{\btheta}(n) = \widetilde{\mathcal{O}}(R^2\sqrt{d/n})$. 

\subsection{Proof Sketch of Theorem \ref{thm: generic}}\label{subsec: proof sketch}

We briefly describe the proof of Theorem \ref{thm: generic} here. The proof follows an iterative fashion, where we show a connection between the estimation error rates in two consecutive iteration rounds and then iterate the analysis to obtain the final result. More specifically, by utilizing the contraction and uniform convergence conditions assumed in Assumption \ref{asmp: w and theta}, if we define the estimation error of round $t$ as $\textup{Er}(t) = \max_{k \in S, r \in [R]}\big(\norm{\hwkt{k}{t}_r-\wks{k}_r} \vee \twonorm{\hthetakt{k}{t}_r-\bthetaks{k}_r}\big)$, we can prove that with high probability,
\begin{align}
	\textup{Er}(t) \leq \kappa_0 \textup{Er}(t-1) + \textup{other terms},
\end{align}
where other terms include the sum of the last four terms on the RHS of \eqref{eq: main bound}, and $\kappa_0 \in (0,1)$. 

We want to highlight that the proof is much harder and more complicated than the proofs in standard EM theory. The reason is that the mixture proportions $\{\wks{k}_r\}_{r=1}^R$ can be heterogenous across tasks, where we can still benefit from federated learning because the similarity between $d$-dimensional parameters $\{\bthetaks{k}_r\}_{r=1}^R$ is more important than the heterogeneity of 1-dimensional scalers $\{\wks{k}_r\}_{r=1}^R$. However, in standard EM theory \citep{balakrishnan2017statistical, yan2017convergence, cai2019chime}, the estimation errors of $\{\wks{k}_r\}_{r=1}^R$ and $\{\bthetaks{k}_r\}_{r=1}^R$ are entangled and it is challenging to separate them by the existing theory. We creatively used a \textit{localization} technique to address the issue by adaptively shrinking the radius of the ball within which uniform convergence in Assumption \ref{asmp: w and theta} must hold. This adaptive radius shrinking trick during iterations finally leads to a ``fast rate", effectively replacing $\mathcal{R}_{\btheta}(n)$ (``the slow rate") with a much smaller $\mathcal{R}_w(n)$ for the term ``cost of heterogeneous mixing proportions" in \eqref{eq: main bound}. The intuition is visually interpreted in Figure \ref{fig: radius}, and more details can be found in the full proof of Theorem \ref{thm: generic} in the appendix.

\begin{figure}[!h]
	\centering
	\includegraphics[width=0.49\textwidth]{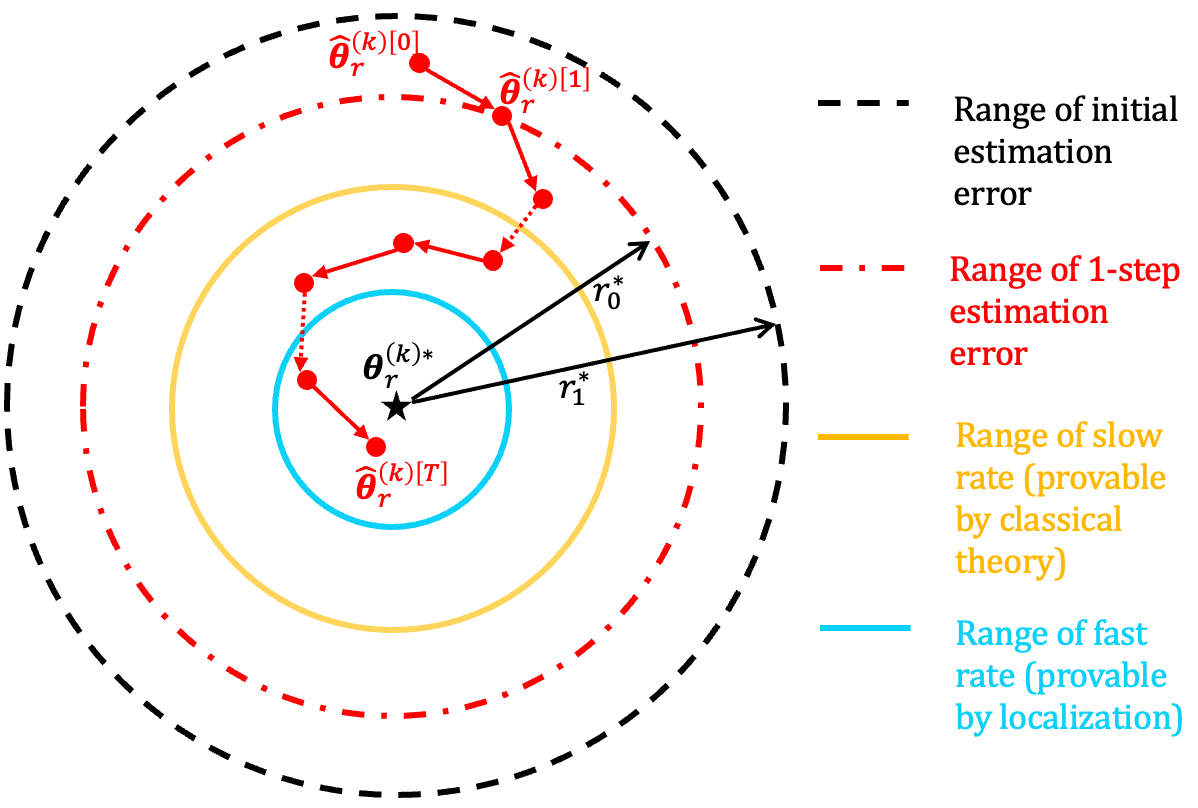}
	\caption{Schematic of the geometric convergence and the localization trick, where we shrink the radius of uniform convergence ball from $r_1^*$ to $r_0^*$ after the first iteration.}
	\label{fig: radius}
\end{figure}

\subsection{Example 1: Gaussian Mixture Models (GMMs)}\label{subsec: gmm}
In this section, we examine Gaussian Mixture Models (GMMs) as an example of \eqref{eq: generic model}. Each observation is from a mixture of $R$ Gaussian distributions ($R \geq 2$):
\vspace{-0.05cm}
\begin{equation}\label{eq: gmm}
	\bxk{k}_i \overset{i.i.d.}{\sim} \sum_{r=1}^R \wks{k}_r\cdot N(\bthetaks{k}_r, \bm{I}_{d \times d}).
\end{equation}
\vspace{-0.27cm}

Hence in \eqref{eq: generic model}, $\pk{k}_r(\,\bm{\cdot}\,\,;\bthetaks{k}_r)$ represents the Lebesgue density of Gaussian distribution $N(\bthetaks{k}_r, \bm{I}_{d \times d})$. We define $\Delta \coloneqq \min_{k \in S}\min_{r \neq r'}\twonorm{\bthetaks{k}_r - \bthetaks{k}_{r'}}$, which characterizes the signal-to-noise ratio of GMMs among $S$. We impose the following assumption.
\begin{assumption}\label{asmp: gmm}
	Suppose the following conditions hold:
	\begin{enumerate}[(i)]
		\item (Bounded parameters) $\wks{k}_r \gtrsim 1/R$, $\twonorm{\bthetaks{k}_r}\leq C$ for all $k \in S$ and $r \in [R]$, where $C$ is a constant;
		\item (Good initialization) $\max_{k \in S, r \in [R]}\norm{\hwkt{k}{0}_r-\wks{k}_r} \lesssim \frac{1}{R}$, $\max_{k \in S, r \in [R]}\twonorm{\hthetakt{k}{0}_r-\bthetaks{k}_r} \lesssim \Delta$;
		\item (Large signal strength) $\Delta \gtrsim \log(R)$;
		\item (Sample size) $n \gtrsim R^4[d+\log (RK)] \Delta^{-2}$;
		\item (Step size) $1-\etak{k}_r\wks{k}_r < c$, and $0 < \etak{k}_r \leq 1/\wks{k}_r$ for all $k \in S$ and $r \in [R]$, where $c > 0$ is a small constant.
	\end{enumerate}
\end{assumption}

\begin{proposition}\label{prop: gmm}
	Under Assumption \ref{asmp: gmm}, GMMs defined in \eqref{eq: gmm} satisfies Assumptions \ref{asmp: q}, \ref{asmp: w and theta}, and \ref{asmp: thm generic} with 
	\begin{align}
		\muk{k}_r &= \Lk{k}_r = \wks{k}_r, \quad \kappa \asymp R^2\exp\{-C\Delta^2\}, \\
		\mathcal{R}_w(n) &= \widetilde{\mathcal{O}}\bigg(R^2\sqrt{\frac{1}{n}}\bigg),	\quad \mathcal{R}_{\btheta}(n) =  \widetilde{\mathcal{O}}\bigg(R^2\sqrt{\frac{d}{n}}\bigg),
	\end{align}
	where $C > 0$ is some constant.
\end{proposition}

By plugging the rates in Propositions \ref{prop: gmm} into Theorem \ref{thm: generic}, we obtain the following estimation error for GMMs.

\begin{corollary}\label{cor: gmm}
	Under Assumption \ref{asmp: gmm}, for the GMMs defined in \eqref{eq: gmm}, for any contaminated set $S^c$ with $\epsilon=|S^c|/K \leq 1/3$ and contaminated distribution $\mathbb{Q}_{S^c}$, w.p. $1-\smallo(1)$, for all $T \geq 1$, FedGrEM satisfies
	\vspace{-0.03cm}
	\begin{align}
		&\max_{k \in S, r \in [R]}\big(\norm{\hwkt{k}{T}_r-\wks{k}_r} \vee \twonorm{\hthetakt{k}{T}_r-\bthetaks{k}_r}\big) \\
		&= \widetilde{\mathcal{O}}\Bigg(\kappa_0^T + R^2\sqrt{\frac{d}{nK}} + R^2\sqrt{\frac{1}{n}} + \min\bigg\{h, R^2\sqrt{\frac{d}{n}}\bigg\}  \\
		&\quad+ \epsilon R^2\sqrt{\frac{d}{n}}\Bigg).
	\end{align}
	\vspace{-0.4cm}
	
	where $\kappa_0 \in (0, 1)$ is a constant.
\end{corollary}

Our theoretical analysis is also applicable to local single-task EM and gradient EM, enabling us to establish an upper bound of estimation error $\widetilde{\mathcal{O}}_{\tP}\Big(R^2\sqrt{\frac{d}{n}}\Big)$ on $S$. Consequently, when $d \rightarrow \infty$ (diverging dimension), $K \rightarrow \infty$ (many similar tasks), $h \ll R^2\sqrt{\frac{d}{n}}$ (sufficient similarity), and $\epsilon \rightarrow 0$ (small proportion of outlier tasks), FedGrEM exhibits a better estimation error rate than single-task EM and gradient EM (up to logarithmic factors). Notably, FedGrEM always achieves an error rate at least as good as the single-task rate $\widetilde{\mathcal{O}}_{\tP}\Big(R^2\sqrt{\frac{d}{n}}\Big)$.

\subsection{Example 2: Mixture of Regressions (MoRs)}
As a second example, we consider a mixture of linear regressions (MoRs), where each observation comes from a mixture of $R$ linear regression models ($R \geq 2$):
\begin{align}
	\zk{k}_i &\overset{i.i.d.}{\sim} \sum_{r=1}^R \wks{k}_r\cdot \delta_r,\\
	\text{Given }\zk{k}_i = r:&\,\,\,\,\,\, \yk{k}_i= (\tbxk{k}_i)^T\bthetaks{k}_r + \epsilonk{k}_i, \label{eq: mor}\\
	\epsilonk{k}_i \overset{i.i.d.}{\sim} N(0, 1),\quad & \tbxk{k}_i \overset{i.i.d.}{\sim} N(\bm{0}_d, \bm{I}_{d \times d}) , \quad \epsilonk{k}_i \ind \tbxk{k}_i.
\end{align}
\vspace{-0.6cm}

Hence in \eqref{eq: generic model} and \eqref{eq: generic model 2}, $\bxk{k}_i$ is the pair $(\tbxk{k}_i, \yk{k}_i)$ and $\pk{k}_r(\,\bm{\cdot}\,\,;\bthetaks{k}_r)$ is the Lebesgue density of joint distribution of $(\tbxk{k}_i, \yk{k}_i)$. We impose the following assumption set.
\begin{assumption}\label{asmp: mor}
	Suppose the same conditions in Assumption \ref{asmp: gmm} hold by replacing (\rom{3}) with:
	\vspace{-0.1cm}
	
	\begin{enumerate}[(\rom{3})]
		\item (Large signal strength) $\Delta \gtrsim R^3 + R^2(\log\Delta)^{3/2}$.
	\end{enumerate}
\end{assumption}

\begin{proposition}\label{prop: mor}
	Under Assumption \ref{asmp: mor}, the MoRs defined in \eqref{eq: mor} satisfies Assumptions \ref{asmp: q}, \ref{asmp: w and theta}, and \ref{asmp: thm generic} with 
	\vspace{-0.7cm}
	
	\begin{align}
		&\muk{k}_r = \wks{k}_r - CR\frac{\sqrt{\log \Delta}}{\Delta}, \Lk{k}_r = \wks{k}_r + CR\frac{\sqrt{\log \Delta}}{\Delta},\\[-12pt]
		&\resizebox{1.03\columnwidth}{!}{$\kappa = \widetilde{\mathcal{O}}\bigg(\frac{R^2}{\Delta}\bigg), \mathcal{R}_w(n) = \widetilde{\mathcal{O}}\bigg(R^2\sqrt{\frac{1}{n}}\bigg),	\mathcal{R}_{\btheta}(n) =  \widetilde{\mathcal{O}}\bigg(R^2\sqrt{\frac{d}{n}}\bigg)$,}
	\end{align}
	\vspace{-0.9cm}
	
	where $C > 0$ is some constant.
\end{proposition}

\vspace{-0.1cm}
By plugging the rates in Propositions \ref{prop: mor} into Theorem \ref{thm: generic}, we have the following estimation error for MoRs.

\begin{corollary}\label{cor: mor}
	Under Assumption \ref{asmp: mor}, for the MoRs defined in \eqref{eq: mor}, for any contaminated set $S^c$ with $\epsilon=|S^c|/K \leq 1/3$ and contaminated distribution $\mathbb{Q}_{S^c}$, with probability $1-\smallo(1)$, for all $T \geq 1$, FedGrEM satisfies
	\vspace{-0.125cm}
	\begin{align}
		&\max_{k \in S, r \in [R]}\big(\norm{\hwkt{k}{T}_r-\wks{k}_r} \vee \twonorm{\hthetakt{k}{T}_r-\bthetaks{k}_r}\big) = \widetilde{\mathcal{O}}\Bigg(\\
		&\resizebox{1.03\columnwidth}{!}{$\kappa_0^T + R^2\sqrt{\frac{d}{nK}} + R^2\sqrt{\frac{1}{n}} + \min\bigg\{h, R^2\sqrt{\frac{d}{n}}\bigg\}+ \epsilon R^2\sqrt{\frac{d}{n}}\Bigg)$}.
	\end{align}
	\vspace{-0.6cm}
	
	where $\kappa_0 \in (0, 1)$ is a constant.
\end{corollary}

We can similarly discuss when the rate of FedGrEM is better than the estimation error of the local single-task algorithms for GMMs, which we do not repeat here.

\section{Numerical Results}\label{sec: numerical}

\subsection{Simulations}

\begin{figure*}[!h]
	\includegraphics[width=\textwidth]{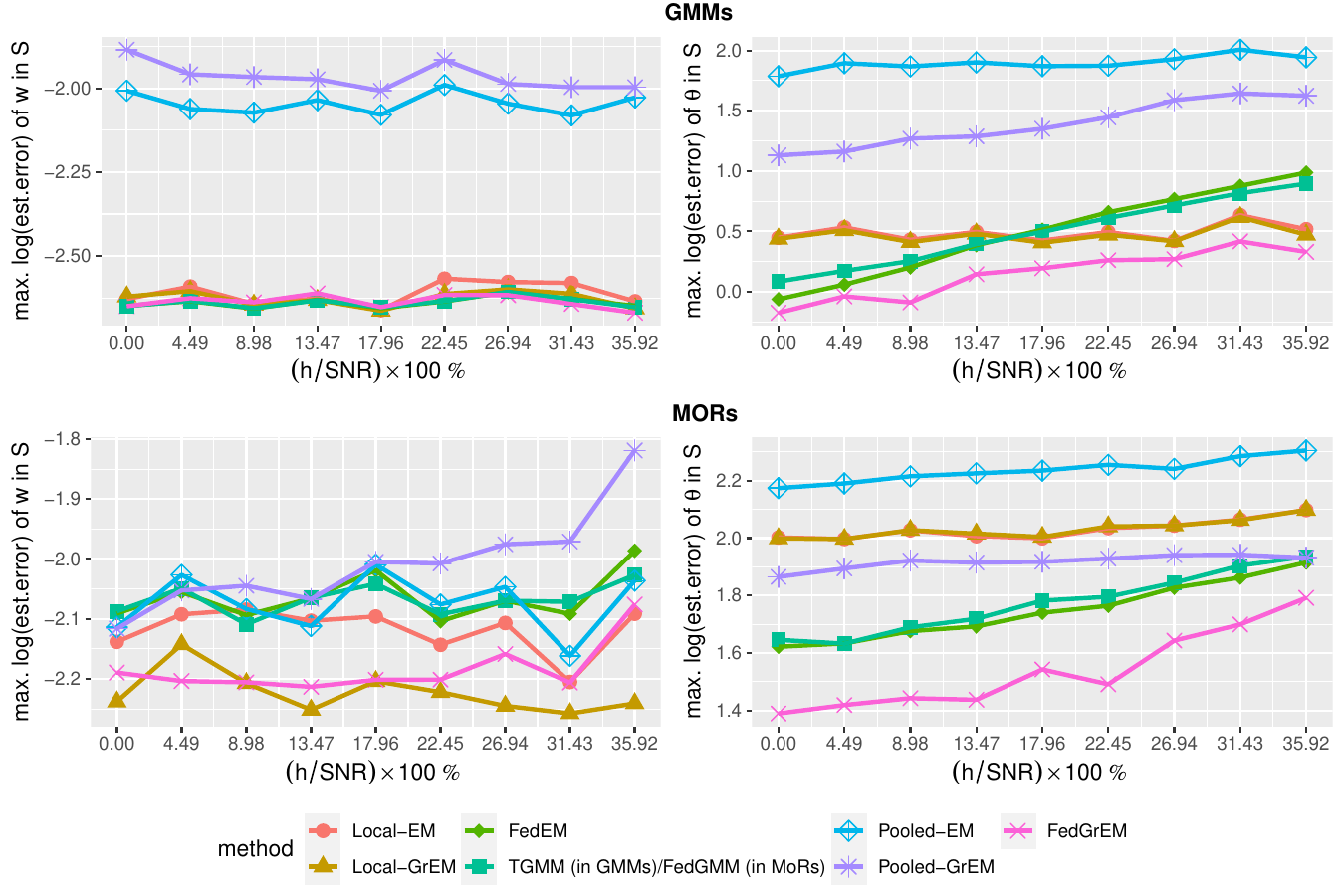}
	\caption{The average estimation errors of different methods in 100 replications of the GMM and MoR simulations (in $\log_e$ scale). The left two figures show the estimation error $\max_{k \in S}\max_{r \in [R]}\log(\norm{\hwkt{k}{T}_r - \wks{k}_r})$ and the right two figures show the estimation error $\max_{k \in S}\max_{r \in [R]}\log(\twonorm{\hthetakt{k}{T}_r - \bthetaks{k}_r})$. $x$-axis represents the ratio between model heterogeneity $h$ and SNR (signal-to-noise ratio), where the definition of SNR is in Section \ref{subsec: additional simulations} of the appendix.}
	\label{fig: simulation}
\end{figure*}

In this subsection, we present simulation results to empirically validate our theoretical insights. We consider two examples in the last section: Gaussian Mixture Models (GMMs) and Mixture of Regressions (MoRs). For both examples, we set the number of tasks $K = 10$, the number of clusters $R = 5$, the sample size of each task $n = 150$ (per-cluster sample size is around $30$), the dimension $p = 10$, and introduce one outlier task $(\epsilon = 0.1)$.  The simulations are conducted over 100 repetitions, and we generate $\wks{k}$ in dependently from $\text{Dirichlet}(5,5,5,5,5)$. Details regarding the values of $\bthetaks{k}$'s and the generation mechanism of the outlier task can be found in Section \ref{sec: additional numerical results appendix} of the appendix. We vary the value of $h$ from 0 to 2 with an increment of 0.25, calculating the average mean estimation error of $\wks{k}$'s and $\bthetaks{k}$'s in $S = 1:9$ for different approaches. 

The considered methods include several unsupervised federated learning or multi-task learning benchmarks: Local-EM (single-task EM), Local-GrEM (single-task gradient EM), FedEM  \citep{marfoq2021federated}, TGMM \citep{wang2021general}, FedGMM \citep{wu2023personalized}, Pooled-EM (EM on pooled data), Pooled-GrEM (gradient EM on pooled data), and FedGrEM (ours).

The results are presented in Figure \ref{fig: simulation}. In the GMM simulation, the estimation errors of $\wks{k}$'s for different methods are similar, except for EM methods on the pooled data. FedGrEM, FedEM, TGMM, and FedGMM outperform the others in estimating $\bthetaks{k}$'s with FedGrEM surpassing the other three due to its robustness to outlier tasks. As $h$ increases, the performance of FedEM and TGMM degrades, becoming inferior to EM and GrEM. Conversely, FedGrEM's performance becomes comparable to EM and GrEM as $h/\textup{SNR}$ approaches $35.92\%$, highlighting its adaptability to unknown $h$. Similar trends are observed in the MoR simulation. These numerical results align with our theoretical analyses and confirm FedGrEM's advantage in handling unknown similarity levels $h$ and a few outlier tasks.

\subsection{Real-data Studies}

\begin{table*}[!ht]
\centering
\begin{tabular}{@{} c|ccccccc @{}} 
 \toprule 
$\epsilon$/Method  & Local-EM  & Local-GrEM & FedEM & FedGrEM & TGMM & Pooled-EM & Pooled-GrEM\\
 \hline 
0\%  & 13.30 (1.17)     & 13.20 (1.77)     & 15.75 (2.09) & \textbf{9.62 (1.03)}     & 25.14 (2.72)     & 26.84 (5.54) & 34.88 (11.70)\\
6.8\% & 12.88 (1.11)     & 12.96 (1.70)     & 15.87 (2.20) & \textbf{9.31 (1.10)}     & 25.00 (3.16)     & 26.60 (5.80) & 34.61 (11.92)\\
13.6\%  & 12.62 (1.26)     & 12.99 (1.99)     & 17.56 (2.44) & \textbf{9.04 (1.06)}    & 25.68 (2.85)     & 26.79 (5.53) & 35.61 (12.52)\\
20.5\% & 12.94 (1.37)     & 13.31 (1.93)     & 18.52 (2.38) & \textbf{9.39 (1.25)}     & 26.14 (3.08)     & 27.05 (5.46) & 37.94 (13.65)\\
 \bottomrule 
\end{tabular}
\caption{Average mis-clustering error rates (standard deviations) in percentages for Pen-Based Recognition of Handwritten Digits dataset}
\label{table: pen}
\end{table*}

\begin{table*}[!ht]
\centering
\begin{tabular}{@{} c|ccccccc @{}} 
 \toprule 
$\epsilon$/Method  & Local-EM & Local-GrEM & FedEM  & FedGrEM & TGMM & Pooled-EM & Pooled-GrEM\\
 \hline 
0\%  & 12.47 (1.19)     & 12.06 (0.95)     & \textbf{9.97 (2.09)} & 10.63 (2.72)     & 12.18 (4.08)     & 12.84 (2.92) & 10.67 (2.41)\\
8\% & 12.30 (1.10)     & 11.98 (0.88)     & 11.97 (2.80) & \textbf{10.64 (1.98)}     & 13.66 (4.13)     & 15.07 (5.25) & 14.16 (4.23)\\
16\%  & 12.47 (1.19)     & 12.06 (0.93)     & 14.46 (2.95) & \textbf{10.96 (1.96)}     & 14.47 (4.24)     & 16.16 (5.46) & 14.97 (4.36)\\
24\% & 12.43 (1.11)     & 12.10 (0.89)     & 21.67 (4.33) & \textbf{10.97 (1.59)}    & 15.70 (5.01)     & 15.50 (5.09) & 14.35 (3.96)\\
 \bottomrule 
\end{tabular}
\caption{Average mis-clustering error rates (standard deviations) in percentages for MNIST dataset}
\label{table: mnist}
\end{table*}

\begin{table*}[!ht]
\centering
\begin{tabular}{@{} c|ccccccc @{}} 
 \toprule 
$\epsilon$/Method  & Local-EM & Local-GrEM & FedEM  & FedGrEM & TGMM & Pooled-EM & Pooled-GrEM\\
 \hline 
0\%  & 36.40 (0.67)     & 35.66 (0.64)     & 36.06 (1.42) & \textbf{34.44 (1.98)}     & 36.14 (2.89)     & 36.31 (1.25) & 35.72 (1.07)\\
8\% & 36.40 (0.72)     & 35.69 (0.65)     & 36.82 (1.15) & \textbf{33.27 (1.96)}     & 36.71 (3.32)     & 39.75 (1.41) & 39.71 (1.54)\\
16\%  & 36.34 (0.66)     & 35.65 (0.70)     & 36.80 (1.28) & \textbf{33.29 (1.86)}     & 37.50 (3.16)     & 39.65 (1.50) & 39.34 (1.91)\\
24\% & 36.31 (0.69)     & 35.61 (0.73)     & 37.93 (1.81) & \textbf{33.35 (1.69)}     & 38.20 (3.28)     & 39.51 (1.24) & 39.44 (1.57)\\
 \bottomrule 
\end{tabular}
\caption{Average mis-clustering error rates (standard deviations) in percentages for Fashion-MNIST dataset}
\label{table: fashionmnist}
\end{table*}

We also conduct experiments on three real datasets:
\begin{itemize}
	\item Pen-Based Recognition of Handwritten Digits \footnote{\url{https://archive.ics.uci.edu/dataset/81/pen+based+recognition+of+handwritten+digits}}: This dataset collects 0-9 digits written by 44 writers on the tablet with 16 features related to each digit such as the pressure level at certain coordinates. The ID of the writer for each handwritten digit is provided. Hence, it is a federated multi-task learning dataset in nature by viewing each writer as a client.
	\item MNIST \footnote{\url{https://www.kaggle.com/datasets/hojjatk/mnist-dataset}}: 70000 grayscale images of $28 \times 28$ pixels for the handwritten digits 0-9 from different writers. There is no information about the writers, hence we manually created a federated learning dataset by randomly assigning each image to one of 100 clients.
	\item Fashion-MNIST \footnote{\url{https://www.kaggle.com/datasets/zalando-research/fashionmnist}}: 70000 of Zalando's article grayscale images in $28 \times 28$ pixels, each associated with a label from 10 classes. We manually created a federated learning dataset by randomly assigning each image to one of 100 clients.
\end{itemize}

In each replication, 80\% data for each task is used as training data, and the remaining 20\% is used as test data to calculate the mis-clustering error. We also contaminate different proportions ($\epsilon$) of tasks to showcase the robustness of FedGrEM against adversarial attacks. We run different benchmark methods with GMMs (without the knowledge of the true labels) for 200 replications and compare their performances in terms of mis-clustering error rates, as shown in Tables \ref{table: pen}-\ref{table: fashionmnist}, where FedGrEM performs the best in most settings. More details about the datasets, pre-processing steps, and results can be found in Section \ref{subsec: real studies} of the appendix.

\section{Discussions}
In this work, we introduced a federated gradient EM algorithm (FedGrEM) to enhance the existing federated EM methods, by considering the task heterogeneity and adversarial attacks. We studied the non-asymptotic theory on general mixture models, and applied the theory to GMMs and MoRs to obtain the explicit estimation error of the model parameters and mixture proportions. Our theory helps illustrate the empirical success of existing federated EM algorithms in literature and offers new theoretical insights on unsupervised federated learning. The proposed FedGrEM was shown to be adaptive to unknown task similarity, robust against the adversarial attack on a small proportion of tasks, protective for the local data, computationally and communicationally efficient. It serves as a valuable supplement to existing federated EM algorithms.

Some additional discussions on the limitations and future extensions are available in Section \ref{sec: more discussions appendix} of the appendix.

\section*{Acknowledgements}
All the experiments were conducted on Ginsburg HPC cluster of Columbia University. Haolei Weng was partially supported by NSF-DMS 2210505. Yang Feng was partially supported by NSF Grant DMS-2324489 and NIH Grant 1R21AG074205-01.

\section*{Impact Statement}
This paper aims to study the non-asymptotic theory underlying the federated EM algorithms, accounting for task heterogeneity and adversarial contaminations. It seeks to illustrate when and how federated learning can improve local performance and offers novel insights into unsupervised federated learning, a field with potential societal impacts, particularly in data privacy and security. While this study primarily advances Machine Learning, we acknowledge its indirect implications on ethical considerations in data handling and algorithm deployment. We believe our findings will contribute to the development of more secure and ethically-aware federated learning algorithms.


\bibliography{reference.bib}
\bibliographystyle{icml2024}

\newpage
\appendix
\onecolumn

\section*{Appendix}

This appendix collects the additional theoretical and numerical details as well as all the technical proofs of the theory. We present the formal theoretical results in Section \ref{sec: formal theory appendix} which correspond to the simplified versions in Section \ref{sec: theory} of the main text. Section \ref{sec: additional numerical results appendix} contains more details of the numerical studies presented in Section \ref{sec: numerical} of the main text. Section \ref{sec: label permutation} discusses the label permutation issue we mentioned in Section \ref{subsec: contributions}. Section \ref{sec: more discussions appendix} includes additional discussions on limitations and potential extensions of the current work in the future. All the technical proofs are summarized in Section \ref{sec: proofs appendix}.

We recall our mathematical notations here. $\tP$ and $\tE$ denote the probability and expectation, respectively. $\tP$ and $\tE$ denote the probability and expectation, respectively. For two positive sequences $\{a_n\}$ and $\{b_n\}$, $a_n \ll b_n$ means $a_n/b_n \rightarrow 0$, $a_n \lesssim b_n$ or $a_n = \mathcal{O}(b_n)$ means $a_n/b_n \leq C < \infty$, and $a_n \asymp b_n$ means $a_n/b_n, b_n/a_n \leq C< \infty$. For a random variable $x_n$ and a positive sequence $a_n$, $x_n = \mathcal{O}_p(a_n)$ means that for any $\epsilon > 0$, there exists $M > 0$ such that $\sup_{n}\tP(|x_n/a_n|>M) \leq \epsilon$. $\widetilde{\mathcal{O}}_p(a_n)$ has a similar meaning up to logarithmic factors in $a_n$. For a vector $\bx \in \mathbb{R}^d$, $\twonorm{x}$ represents its Euclidean norm. For two numbers $a$ and $b$, $a \vee b = \max\{a,b\}$ and $a \wedge b = \min\{a,b\}$. For any positive integer $K$, $1:K$ and $[K]$ stand for the set $\{1, 2, \ldots, K\}$. Denote $\mathcal{B}_{\xi}(\bm{\btheta})$ as an Euclidean ball centered at $\btheta$ with radius $\xi > 0$. And for any set $S$, $|S|$ denotes its cardinality and $S^c$ denotes its complement. ``w.p." stands for ``with probability". ``WLOG" stands for ``Without loss of generality". The absolute constants $c$ and $C$ may vary from line to line.

\section{Formal Theoretical Results}\label{sec: formal theory appendix}
Denote $\bar{\eta} = \max_{k\in S, r \in [R]}\etak{k}_r$ as the maximum step size used in the local M-step on tasks in $S$. For simplicity, denote $\qk{k}(\btheta) = \Qk{k}(\btheta|\bwks{k}, \bthetaks{k})$.  We first state a few assumptions which are necessary for our result on general mixture models.

\begin{assumption}\label{asmp: q appendix}
	For any $k \in S$, there exist non-negative sets $\{\muk{k}_r\}_{r\in [R]}$, $\{\Lk{k}_r\}_{r\in [R]}$, and a positive constant $r^*_1$ such that for all $\btheta = \{\btheta_r\}_{r=1}^R$, $\btheta' = \{\btheta_r'\}_{r=1}^R$ with $ \btheta_r, \btheta_r' \in \mathcal{B}_{r_1^*}(\bthetaks{k}_r)$:
	\begin{enumerate}[(i)]
		\item (Strong concavity) $\qk{k}(\btheta') - \qk{k}(\btheta) - \frac{\partial}{\partial \btheta}\qk{k}(\btheta)^T(\btheta'-\btheta) \leq -\sum_{r=1}^R\frac{\mu_r^{(k)}}{2}\twonorm{\btheta'_r-\btheta_r}^2$;
		\item (Smoothness) $\qk{k}(\btheta') - \qk{k}(\btheta) - \frac{\partial}{\partial \btheta} \qk{k}(\btheta)^T(\btheta'-\btheta) \geq -\sum_{r=1}^R\frac{L_r^{(k)}}{2}\twonorm{\btheta'_r-\btheta_r}^2$.
	\end{enumerate}
\end{assumption}

\begin{remark}
	The same conditions are imposed by \citet{balakrishnan2017statistical} in the single-task setting. The strong concavity is usually assumed to obtain the parametric convergence rate, and the smoothness is imposed for gradient descent to converge at a geometric rate.
\end{remark}

\begin{assumption}\label{asmp: w appendix}
	There exist positive constants $r_w^*$, $r^*_2$, and $\kappa \in (0, 1)$, and a function $\mathcal{W}$, such that for any $k \in S$:
	\begin{enumerate}[(i)]
		\item For all $\bw' = \{w_r'\}_{r=1}^R$ and $\btheta' = \{\btheta_r'\}_{r=1}^R$ with $w_r' \in \mathcal{B}_{r_w^*}(\wks{k}_r)$ and $\btheta_r' \in \mathcal{B}_{r_2^*}(\bthetaks{k}_r)$, we have $\big|\tE\big[\tP(\zk{k}=r|\bxk{k};\bw', \btheta')\big]-\wks{k}_r\big| \leq \kappa\cdot \sum_{r=1}^R \big(\norm{w_r' - \wks{k}_r} + \twonorm{\btheta_r' - \bthetaks{k}_r}\big)$;
		\item w.p. at least $1-\delta$, for all $\bw' = \{w_r'\}_{r=1}^R$, $\xi > 0$, and $\btheta' = \{\btheta_r'\}_{r=1}^R$ with $w_r' \in \mathcal{B}_{r_w^*}(\wks{k}_r)$ and $\btheta_r' \in \mathcal{B}_{\xi}(\bthetaks{k}_r)$, we have $\big|\frac{1}{n}\sum_{i=1}^n\tP(\zk{k}=r|\bxk{k}_i;\bw', \btheta') - \tE\big[\tP(\zk{k}=r|\bxk{k};\bw', \btheta')\big]\big| \leq \mathcal{W}(n, \delta, \xi)$.
	\end{enumerate}	
\end{assumption}

\begin{remark}
	Note that by definition $\wks{k}_r = \tE[\tP(\zk{k}=r|\bxk{k};\bwks{k}, \bthetaks{k})]$. Therefore, condition (\rom{1}) describes the behavior of $\tE\big[\tP(\zk{k}=r|\bxk{k};\bw', \btheta')\big]$, and condition (\rom{2}) is a uniform convergence assumption on $\tP(\zk{k}=r|\bxk{k}_i;\bw', \btheta')$, when $\bw'$ and $\btheta'$ are close to the true values $\bwks{k}$ and $\bthetaks{k}$.
\end{remark}

\begin{assumption}\label{asmp: theta appendix}
	With the same constants $r_w^*$ and $r^*_2$ in Assumption \ref{asmp: w appendix}, there exists a constant $\gamma \in (0,1)$ and functions $\mathcal{E}_1$, $\mathcal{E}_2$ such that for any $k \in S$:
	\begin{enumerate}[(i)]
		\item For all $\bw' = \{w_r'\}_{r=1}^R$ and $\btheta' = \{\btheta_r'\}_{r=1}^R$ with $w_r' \in \mathcal{B}_{r_w^*}(\wks{k}_r)$ and $\btheta_r' \in \mathcal{B}_{r_2^*}(\bthetaks{k}_r)$, we have $\big\|\frac{\partial}{\partial \btheta_r}\qk{k}(\btheta)|_{\btheta = \btheta'} - \frac{\partial}{\partial \btheta_r}\Qk{k}(\btheta|\bw', \btheta')|_{\btheta = \btheta'}\big\| \leq \gamma\cdot \sum_{r=1}^R \big(\norm{w_r' - \wks{k}_r} + \twonorm{\btheta_r' - \bthetaks{k}_r}\big)$;
		\item w.p. at least $1-\delta$, for all $\bw' = \{w_r'\}_{r=1}^R$ and $\btheta' = \{\btheta_r'\}_{r=1}^R$ with $w_r' \in \mathcal{B}_{r_w^*}(\wks{k}_r)$ and $\btheta_r' \in \mathcal{B}_{r_2^*}(\bthetaks{k}_r)$, we have $\big\|\frac{\partial}{\partial \btheta_r}\hQk{k}(\btheta|\bw', \btheta')|_{\btheta = \btheta'} - \frac{\partial}{\partial \btheta_r}\Qk{k}(\btheta|\bw', \btheta')|_{\btheta = \btheta'}\big\|_2 \leq \mathcal{E}_1(n, \delta)$;
		\item w.p. at least $1-\delta$, for $\xi > 0$ and all $\bw^{(k)\prime} = \{w^{(k)\prime}_r\}_{r=1}^R$,  $\btheta' = \{\btheta_r'\}_{r=1}^R$, and $\{\etak{k}_r\}_{k \in S, r \in [R]}$ with $w^{(k)\prime}_r \in \mathcal{B}_{r_w^*}(\wks{k}_r)$ and $\btheta_r' \in \mathcal{B}_{r_2^*}(\bthetaks{k}_r)$, we have $\big\|\frac{1}{|S|}\sum_{k \in S}\etak{k}_r\cdot\big[\frac{\partial}{\partial \btheta_r}\hQk{k}(\btheta|\bw', \btheta')|_{\btheta = \btheta'} - \frac{\partial}{\partial \btheta_r}\Qk{k}(\btheta|\bw', \btheta')|_{\btheta = \btheta'}\big]\big\|_2 \leq \bar{\eta}\mathcal{E}_2(n, |S|, \delta)$.
	\end{enumerate}	
\end{assumption}

\begin{remark}
	Conditions (\rom{1}) and (\rom{2}) have been used by \citet{balakrishnan2017statistical} in the single-task setting, while condition (\rom{3}) is a generalization of (\rom{2}) when aggregating the data from multiple tasks. Similar to condition (\rom{2}) in Assumption \ref{asmp: w appendix}, condition (\rom{2}) and condition (\rom{3}) are uniform convergence assumptions on the gradient around the true parameter values, which are often needed when analyzing the EM without data splitting \citep{yan2017convergence, cai2019chime}.
\end{remark}

\begin{assumption}\label{asmp: thm generic appendix}
	Denote $r_{\btheta}^* = r_1^* \wedge r_2^*$ and $\wtkappa_0 = 119\Big(\sqrt{1-\min_{r, k \in S}(\muk{k}_r\etak{k}_r)}\allowbreak  + \bar{\eta}\gamma R + \kappa R\Big)$. Suppose
	\begin{equation}
		\max\limits_{k \in S, r \in [R]}\norm{\hw^{(k)[0]}_r - \wks{k}_r} \leq r_w^*, \max\limits_{k \in S, r \in [R]}\twonorm{\htheta^{(k)[0]}_r - \bthetaks{k}_r} \leq r_{\btheta}^*.
	\end{equation}
	In addition, $\wtkappa_0$, $\kappa$, $r_w^*$, $r_{\btheta}^*$, and functions $\mathcal{W}$, $\mathcal{E}_1$, and $\mathcal{E}_2$ defined in Assumptions \ref{asmp: q appendix}-\ref{asmp: theta appendix} satisfy
	\begin{enumerate}[(i)]
		\item $\wtkappa_0 \leq \frac{r_{\btheta}^*}{18(r_w^* + r_{\btheta}^*)}$, $\kappa R \leq \frac{9}{1199}\cdot \frac{r_w^*}{r_w^* + r_{\btheta}^*}$;
		\item $\bar{\eta}\cdot  \mathcal{E}_1\Big(n, \frac{\delta}{3RK}\Big) \leq \Big[\frac{(1-\tilde{\kappa}_0/119)(1-\tilde{\kappa}_0)}{4320}r_{\btheta}^*\Big]\wedge \Big(\frac{1}{3}r_w^*\Big)$;
		\item $\mathcal{W}\Big(n, \frac{\delta}{3RK}, r_{\btheta}^*\Big) \leq \frac{(1-\tilde{\kappa}_0/119)(1-\tilde{\kappa}_0)}{2160}r_{\btheta}^*$;
		\item $\bar{\eta}\cdot  \mathcal{E}_2\Big(n, |S|, \frac{\delta}{3R}\Big) \leq \frac{1-\tilde{\kappa}_0/119}{20}r_{\btheta}^*$;
		\item $\etak{k}_r \leq 1/L_r^{(k)}$ for all $k \in S$ and $r \in [R]$.
	\end{enumerate}
\end{assumption}

We set the penalty parameters in Algorithm \ref{algo: FG-EM} by induction as 
\begin{align}
	\lambdat{0} &= \frac{15}{119}\sqrt{n}(r_w^* + r_{\btheta}^*),\\
	\lambdat{t} &= \wtkappa_0\lambdat{t-1} + 15\sqrt{n}\Big[ \mathcal{W}\Big(n, \frac{\delta}{3RK}, r_{\btheta}^*\Big) + 2\bar{\eta}\mathcal{E}_1\Big(n, \frac{\delta}{3RK}\Big)\Big],
\end{align}

\begin{theorem}\label{thm: generic appendix}
	Suppose Assumptions \ref{asmp: q appendix}-\ref{asmp: thm generic appendix} hold. Then for any contaminated set $S^c$ with $\epsilon=|S^c|/K < 1/3$ and any contaminated distribution $\mathbb{Q}_{S^c}$, with probability at least $1-\delta$, for all $T \geq 1$, FedGrEM satisfies
	\begin{align}
		&\max_{k \in S, r \in [R]}\big(\norm{\hwkt{k}{T}_r-\wks{k}_r} \vee \twonorm{\hthetakt{k}{T}_r-\bthetaks{k}_r}\big) 
		\leq \underbrace{20T\tilde{\kappa}_0^{T-1}\times (r_w^* \vee r_{\btheta}^*)}_{\textup{iterative error}}+ \underbrace{\frac{1}{1-\tilde{\kappa}_0/119}\bar{\eta}\mathcal{E}_2\Big(n, |S|, \frac{\delta}{3R}\Big)}_{\textup{aggregation rate}} \\
		&+ \underbrace{\frac{1}{1-\tilde{\kappa}_0/119}\mathcal{W}\Big(n, \frac{\delta}{3RK}, r_{\btheta, T}^*\Big)}_{\textup{cost of heterogeneous mixing proportions}} + \underbrace{\frac{18}{1-\tilde{\kappa}_0/119}\min\bigg\{3h, \frac{6}{1-\tilde{\kappa}_0}\Big[\mathcal{W}\Big(n, \frac{\delta}{3RK}, r_{\btheta}^*\Big) + 2\bar{\eta}\mathcal{E}_1\Big(n, \frac{\delta}{3RK}\Big)\Big]\bigg\}}_{\textup{cost of task heterogeneity}}\\
		&+ \underbrace{\frac{30}{(1-\tilde{\kappa}_0)(1-\tilde{\kappa}_0/119)}\epsilon \Big[\mathcal{W}\Big(n, \frac{\delta}{3RK}, r_{\btheta}^*\Big) + 2\bar{\eta}\mathcal{E}_1\Big(n, \frac{\delta}{3RK}\Big)\Big]}_{\textup{cost of outlier tasks}},
	\end{align}
	where $r_{\btheta, T}^*$ is defined in an iterative fashion by
	\begin{align}
		A_t &= \bigg[9\tilde{\kappa}_0\bigg(\frac{\tilde{\kappa}_0}{119}\bigg)^{t-1} + \frac{118}{119}(t-1)\tilde{\kappa}_0^{t-1}\bigg](r_w^* +r_{\btheta}^*) + \frac{1}{1-\tilde{\kappa}_0/119}\bar{\eta}\mathcal{E}_2\bigg(n, |S|, \frac{\delta}{3R}\bigg) \\
		&\quad + \frac{18}{1-\tilde{\kappa}_0/119} \min\bigg\{3h, \frac{6}{1-\tilde{\kappa}_0}\bigg[\mathcal{W}\bigg(n, \frac{\delta}{3RK}, r_{\btheta}^*\bigg) + 2\bar{\eta}\mathcal{E}_1\bigg(n, \frac{\delta}{3R}\bigg)\bigg]\bigg\} \\
		&\quad + \frac{30}{(1-\tilde{\kappa}_0)(1-\tilde{\kappa}_0/119)}\epsilon\bigg[\mathcal{W}\bigg(n, \frac{\delta}{3RK}, r_{\btheta}^*\bigg) + 2\bar{\eta}\mathcal{E}_1\bigg(n, \frac{\delta}{3R}\bigg)\bigg];\\
		&A_t + \frac{18}{1-\tilde{\kappa}_0/119}\mathcal{W}\bigg(n, \frac{\delta}{3RK}, r_{\btheta, t}^*\bigg) = r_{\btheta, t+1}^*, 
	\end{align}
	for $t \geq 1$ with $r_{\btheta, 1}^* \coloneqq r_{\btheta}^*$.
\end{theorem}

Before jumping into two specific examples, we want to point out that the term $r_{\btheta, T}^*$ is introduced to address a specific challenge in the analysis of gradient EM mentioned in Section \ref{subsec: proof sketch}. In the classical EM theory, the estimates of similar $\{\bthetaks{k}_r\}_{k \in S}$ and heterogeneous $\{\wks{k}_r\}_{k \in S}$ are entangled in the iterations, which will make the heterogeneous scalars $\{\wks{k}_r\}_{k \in S}$ contribute a large dimension-dependent error $\mathcal{W}\Big(n, \frac{\delta}{3RK}, r_{\btheta}^*\Big)$ to the estimation and finally lead to a ``slow rate" of convergence for $\{\bthetaks{k}_r\}_{k \in S, r \in [R]}$. To mitigate this issue, we reduced the radius of the ball within which uniform convergence must hold during iterations. This localization trick finally leads to a ``fast rate", effectively replacing $\mathcal{W}\Big(n, \frac{\delta}{3RK}, r_{\btheta}^*\Big)$ in the slow rate with the current $\mathcal{W}\Big(n, \frac{\delta}{3RK}, r_{\btheta, T}^*\Big)$, where $r_{\btheta, T}^* \ll r_{\btheta}^*$. The intuition has been visually interpreted in Figure \ref{fig: radius}, and more details can be found in the proof of Theorem \ref{thm: generic}.

\subsection{Example 1: Gaussian Mixture Models (GMMs)}\label{subsec: gmm appendix}

\begin{assumption}\label{asmp: gmm appendix}
	Suppose the following conditions hold:
	\begin{enumerate}[(i)]
		\item (Bounded parameters) $\wks{k}_r \geq c_w/R$, $\twonorm{\bthetaks{k}_r}\leq M$ with some $c_w \in (0, 1]$ for all $k \in S$ and $r \in [R]$ and $M \geq C > 0$, where $C$ is a constant;
		\item (Good initialization) $\max_{k \in S, r \in [R]}\norm{\hwkt{k}{0}_r-\wks{k}_r} \leq C_b\frac{c_w}{R}$, $\max_{k \in S, r \in [R]}\twonorm{\hthetakt{k}{0}_r-\bthetaks{k}_r} \leq C_b\Delta$, with $C_b$ a small constant;
		\item (Large signal strength) $\Delta \gtrsim \log(MRc_w^{-1})$;
		\item (Sample size) $n \gtrsim [R^2M^6d+R^2\log^2(Rc_w^{-1})M^2+M^2\log(RK/\delta)]C_b^{-2}\Delta^{-2}\bar{\eta}^2$;
		\item (Step size) $1-\min\limits_{k \in S, r\in [R]}(\etak{k}_r\wks{k}_r) < $ a small constant $c$, and $0 < \etak{k}_r \leq 1/\wks{k}_r$ for all $k \in S$ and $r \in [R]$.
	\end{enumerate}
\end{assumption}

\begin{remark}\label{rmk: const change with n appendix}
	Note that we allow $c_w$, $M$, $C_b$, $\Delta$, $T$, $R$, $K$, and $d$ to change with sample size $n$.
\end{remark}

\begin{proposition}\label{prop: gmm appendix}
	Under Assumption \ref{asmp: gmm appendix}, GMMs defined in \eqref{eq: gmm} satisfies Assumptions \ref{asmp: q appendix}-\ref{asmp: thm generic appendix} with 
	\begin{align}
		\muk{k}_r &= \Lk{k}_r = \wks{k}_r,\\
		r^*_1 &= +\infty, r^*_w = C_b\frac{c_w}{R}, r^*_2 = C_b\Delta,\\
		\kappa &\asymp c_w^{-2}R^2\exp\{-C\Delta^2\}, \, \gamma \asymp M^2c_w^{-2}R^2\exp\{-C\Delta^2\},\\[8pt]	
		\mathcal{W}(n, \delta, \xi) &\smash{\asymp RM\xi\sqrt{\frac{d}{n}} + [RM^2 + R\log(Rc_w^{-1})]\sqrt{\frac{1}{n}}  + \sqrt{\frac{\log(1/\delta)}{n}}},\\[7pt]	
		\mathcal{E}_1(n, \delta) &\asymp  RM^3\sqrt{\frac{d}{n}} + RM\log(Rc_w^{-1})\sqrt{\frac{1}{n}}  + M\sqrt{\frac{\log(1/\delta)}{n}}, \\[8pt]	
		\mathcal{E}_2(n, |S|, \delta) &\asymp  \smash{RM^3\sqrt{\frac{d}{n|S|}} + [RM^3+RM\log(Rc_w^{-1})]\sqrt{\frac{1}{n}}} \\[8pt]	
		&\quad + M\sqrt{\frac{\log(1/\delta)}{n|S|}},
	\end{align}
	where $C > 0$ is some constant.
\end{proposition}

By plugging the rates in Propositions \ref{prop: gmm} into Theorem \ref{thm: generic}, we obtain the following result for GMMs.

\begin{corollary}\label{cor: gmm appendix}
	Set $\etak{k}_r = (1+C_b)^{-1}(\hwkt{k}{0}_r)^{-1}$. Under Assumption \ref{asmp: gmm}, for the GMMs defined in \eqref{eq: gmm}, for any contaminated set $S^c$ with $\epsilon=|S^c|/K \leq 1/3$ and contaminated distribution $\mathbb{Q}_{S^c}$, with probability at least $1-\delta$, for all $T \geq 1$, FedGrEM satisfies
	\begin{align}
		&\max_{k \in S, r \in [R]}\big(\norm{\hwkt{k}{T}_r-\wks{k}_r} \vee \twonorm{\hthetakt{k}{T}_r-\bthetaks{k}_r}\big) \\
		&\lesssim T^2\kappa_0^{T-1}M + R^2c_w^{-1}M^3\sqrt{\frac{d}{n|S|}} \\
		&\quad + Rc_w^{-1}M\sqrt{\frac{\log(RK/\delta)}{n}}\\
		&\quad + R^2c_w^{-1}M[M^2+\log(Rc_w^{-1})]\sqrt{\frac{1}{n}} \\
		&\quad + \min\bigg\{h, R^2c_w^{-1}M^3\sqrt{\frac{d}{n}}\bigg\}  + \epsilon R^2c_w^{-1}M^3\sqrt{\frac{d}{n}},
	\end{align}
	where $\kappa_0 = 119\sqrt{\frac{2C_b}{1+C_b}} + CM^2c_w^{-2}R^3\exp\{-C'\Delta^2\} + Cc_w^{-2}R^3\exp\{-C'\Delta^2\} +  \tilde{\kappa}_0' \in (0,1)$, and $\tilde{\kappa}_0'$ satisfies $1> \tilde{\kappa}_0' > CMR\sqrt{\frac{d}{n}}$ for some $C > 0$..
\end{corollary}

\begin{remark}\label{rmk: gmm simplified rate appendix}
	If $M$ and $c_w$ are bounded, when $T \gtrsim \log n$, we will have
	\begin{align}
		&\max_{k \in S, r \in [R]}\big(\norm{\hwkt{k}{T}_r-\wks{k}_r} \vee \twonorm{\hthetakt{k}{T}_r-\bthetaks{k}_r}\big) \\
		&= \widetilde{\mathcal{O}}_{\tP}\bigg(R^2\sqrt{\frac{d}{n|S|}} + R^2\sqrt{\frac{1}{n}} + \min\bigg\{h, R^2\sqrt{\frac{d}{n}}\bigg\} + \epsilon R^2\sqrt{\frac{d}{n}}\bigg).
	\end{align}
\end{remark}

Next, we want to illustrate the choice of step size $\etak{k}_r$. In Corollary 
\ref{cor: gmm appendix}, we set $\etak{k}_r = (1+C_b)^{-1}(\hwkt{k}{0}_r)^{-1}$, then under Assumption \ref{asmp: gmm appendix}, it can be shown that Assumptions \ref{asmp: thm generic appendix}.(\rom{1}) and \ref{asmp: thm generic appendix}.(\rom{5}) hold, $\sqrt{1-\min_{k \in S, r\in [R]}(\etak{k}_r\muk{k}_r)} \leq \sqrt{\frac{2C_b}{1+C_b}}$, and we can replace $\bar{\eta}$ with $CRc_w^{-1}$ for some constant $C$ in Assumptions \ref{asmp: thm generic appendix}.(\rom{2}) and \ref{asmp: thm generic appendix}.(\rom{5}).

Second, we have the following upper bound for $r_{\btheta, T}^*$, which can be plugged into Theorem \ref{thm: generic appendix} with the rates in Proposition \ref{prop: gmm appendix} to obtain the upper bound of estimation error in Corollary \ref{cor: gmm appendix}.

\begin{proposition}\label{prop: gmm rj appendix}
	Under Assumption \ref{asmp: gmm appendix}, for the GMMs defined in \eqref{eq: gmm}, we have
	\begin{align}
		r_{\btheta, T}^* &\lesssim T^2\kappa_0^{T-1}M + \bar{\eta}RM^3\sqrt{\frac{d}{n|S|}}  + [(\bar{\eta}M)\vee 1][RM^2 + R\log(Rc_w^{-1})]\sqrt{\frac{1}{n}}  + [(\bar{\eta}M)\vee 1]\sqrt{\frac{\log(RK/\delta)}{n}} \\
		&\quad + \min\bigg\{h, \bar{\eta}RM^3\sqrt{\frac{d}{n}}\bigg\} +\epsilon\bar{\eta}RM^2[(\bar{\eta}M)\vee 1]\sqrt{\frac{d}{n}},
	\end{align}
	where $\kappa_0 = 119\sqrt{\frac{2C_b}{1+C_b}} + CM^2c_w^{-2}R^3\exp\{-C'\Delta^2\} + Cc_w^{-2}R^3\exp\{-C'\Delta^2\} +  \tilde{\kappa}_0'$, and $\tilde{\kappa}_0'$ satisfies $1> \tilde{\kappa}_0' > CMR\sqrt{\frac{d}{n}}$ for some $C > 0$.
\end{proposition}

\subsection{Example 2: Mixture of Regressions (MoRs)}
\begin{assumption}\label{asmp: mor appendix}
	Suppose the same conditions in Assumption \ref{asmp: gmm appendix} hold by replacing (\rom{3}) with:
	\begin{enumerate}[(\rom{3})]
		\item (Strong signal strength) $\Delta \gtrsim R^3c_w^{-1} + R^2c_w^{-1}(\log\Delta)^{3/2}$;
	\end{enumerate}
\end{assumption}

Similar to our previous comments in Remark \ref{rmk: const change with n appendix} for GMMs, we also allow $c_w$, $M$, $C_b$, $\Delta$, $T$, $R$, $K$, and $d$ to change with sample size $n$ in MoRs.

\begin{proposition}\label{prop: mor appendix}
	Under Assumption \ref{asmp: mor appendix}, the MoRs defined in \eqref{eq: mor} satisfies Assumptions \ref{asmp: q appendix}-\ref{asmp: thm generic appendix} with $r^*_1$, $r_2^*$, $\mathcal{W}(n, \delta, \xi)$, $\mathcal{E}_1(n, \delta)$, $\mathcal{E}_2(n, |S|, \delta)$ the same as in Proposition \ref{prop: gmm appendix}, and
	\begin{align}
		\muk{k}_r &= \wks{k}_r - CR\frac{\sqrt{\log \Delta}}{\Delta},\\
		\Lk{k}_r &= \wks{k}_r + CR\frac{\sqrt{\log \Delta}}{\Delta},\\
		\kappa &\asymp c_w^{-1}R\frac{\sqrt{\log \Delta}}{\Delta} + c_w^{-1}RC_b + R^2c_w^{-1}\frac{1}{\Delta},\\
		\gamma &\asymp c_w^{-1}R\frac{(\log \Delta)^{3/2}}{\Delta} + c_w^{-1}RC_b + R^2c_w^{-1}\frac{1}{\Delta},
	\end{align}
	where $C > 0$ is some constant.
\end{proposition}

\begin{corollary}\label{cor: mor appendix}
	Set $\etak{k}_r = (1+C_b)^{-1}(\hwkt{k}{0}_r)^{-1}$. Under Assumption \ref{asmp: mor appendix}, for the MoRs defined in \eqref{eq: mor}, for any contaminated set $S^c$ with $\epsilon=|S^c|/K \leq 1/3$ and contaminated distribution $\mathbb{Q}_{S^c}$, with probability at least $1-\delta$, for all $T \geq 1$, FedGrEM satisfies
	\begin{align}
		&\max_{k \in S, r \in [R]}\big(\norm{\hwkt{k}{T}_r-\wks{k}_r} \vee \twonorm{\hthetakt{k}{T}_r-\bthetaks{k}_r}\big) \\
		&\lesssim T^2\kappa_0^{T-1}M + R^2c_w^{-1}M^3\sqrt{\frac{d}{n|S|}} \\
		&\quad + Rc_w^{-1}M\sqrt{\frac{\log(RK/\delta)}{n}}\\
		&\quad + R^2c_w^{-1}M[M^2+\log(Rc_w^{-1})]\sqrt{\frac{1}{n}} \\
		&\quad + \min\bigg\{h, R^2c_w^{-1}M^3\sqrt{\frac{d}{n}}\bigg\}  + \epsilon R^2c_w^{-1}M^3\sqrt{\frac{d}{n}},
	\end{align}
	where $\kappa_0 = 119\sqrt{\frac{3C_b}{1+2C_b}+ CR^3c_w^{-2}\frac{(\log \Delta)^{3/2}}{\Delta}} + CR^3c_w^{-2}C_b + CR^4c_w^{-2}\frac{1}{\Delta} +  \tilde{\kappa}_0' \in (0,1)$, and $\tilde{\kappa}_0'$ satisfies $1> \tilde{\kappa}_0' > CMR\sqrt{\frac{d}{n}}$ for some $C > 0$.
\end{corollary}

\begin{remark}
	If $M$ and $c_w$ are bounded, when $T \gtrsim \log n$, we will have the same rate for MoRs as in Remark \ref{rmk: gmm simplified rate appendix}. We can also compare the rate of FedGrEM and the local single-task rates as in GMMs, which we do not repeat here.
\end{remark}

In Corollary \ref{cor: mor appendix}, we set $\etak{k}_r = (1+2C_b)^{-1}(\hwkt{k}{0}_r)^{-1}$ and let $C_b \gtrsim Rc_w^{-1}\frac{\sqrt{\log \Delta}}{\Delta}$, then under Assumption \ref{asmp: mor appendix}, it can be shown that Assumptions \ref{asmp: thm generic appendix}.(\rom{1}) and (\rom{5}) hold, $\sqrt{1-\min_{k \in S, r\in [R]}(\etak{k}_r\muk{k}_r)} \leq \sqrt{\frac{3C_b}{1+2C_b}+ C\frac{\sqrt{\log \Delta}}{\Delta}}$, and we can replace $\bar{\eta}$ with $CRc_w^{-1}$ for some constant $C$ in Assumptions \ref{asmp: thm generic appendix}.(\rom{2}) and \ref{asmp: thm generic appendix}.(\rom{4}).

In addition, we have the following upper bound for $r_{\btheta, T}^*$, which can be plugged into Theorem \ref{thm: generic appendix} with the rates in Proposition \ref{prop: mor appendix} to obtain the upper bound of estimation error in Corollary \ref{cor: mor appendix}.

\begin{proposition}\label{prop: mor rj appendix}
	Under Assumption \ref{asmp: mor appendix}, for the MoRs defined in \eqref{eq: mor}, we have
	\begin{align}
		r_{\btheta, T}^* &\lesssim T^2\kappa_0^{T-1}M + \bar{\eta}RM^3\sqrt{\frac{d}{n|S|}}  + [(\bar{\eta}M)\vee 1][RM^2 + R\log(Rc_w^{-1})]\sqrt{\frac{1}{n}}  + [(\bar{\eta}M)\vee 1]\sqrt{\frac{\log(RK/\delta)}{n}} \\
		&\quad + \min\bigg\{h, \bar{\eta}RM^3\sqrt{\frac{d}{n}}\bigg\} +\epsilon\bar{\eta}RM^2[(\bar{\eta}M)\vee 1]\sqrt{\frac{d}{n}},
	\end{align}
	where $\kappa_0 = 119\sqrt{\frac{3C_b}{1+2C_b}+ CR^3c_w^{-2}\frac{(\log \Delta)^{3/2}}{\Delta}} + CR^3c_w^{-2}C_b + CR^4c_w^{-2}\frac{1}{\Delta} +  \tilde{\kappa}_0'$, and $\tilde{\kappa}_0'$ satisfies $1> \tilde{\kappa}_0' > CMR\sqrt{\frac{d}{n}}$ for some $C > 0$.
\end{proposition}

\section{Additional Details of Numerical Results}\label{sec: additional numerical results appendix}
In this section, we provide more details of the numerical studies in Section \ref{sec: numerical}. All experiments are implemented in R, where EM is executed using the \texttt{mclust} package for GMMs and the \texttt{mixreg} package for MoRs. GrEM, FedEM \citep{marfoq2021federated}, TGMM \citep{wang2021general}, FedGMM \citep{wu2023personalized}, and FedGrEM are initialized with the estimates from local EM. As the empirical results in \citet{wang2021general} suggested, we set the tuning parameter $\lambda = 0.4$ in TGMM, which controls how much information to borrow from the other tasks. TGMM was originally designed for transfer learning in \citet{wang2021general} and we ran it on each task with all the other tasks as sources. For FedGMM, we set the number of mixtures $M_1 = 3$.

\subsection{Additional Details of Simulations}\label{subsec: additional simulations}
In both examples, we generate centers $\bar{\btheta}_r$ of parameters $\{\bthetaks{k}_r\}_{k \in S}$ as
\begin{align}
	\bar{\btheta}_1 &= (1,0,3,-1,1,-1,0,1,1,-1)^T, \\
	\bar{\btheta}_2 &= (0,1,-1,-3,2,-1,2,-1,1,-)^T, \\
	\bar{\btheta}_3 &= (-3,-1,2,-1,2,-1,1,-3,-1,-2)^T, \\
    \bar{\btheta}_4 &= (1,-2,0,-1,-2,2,1,3,1,-1)^T, \\
    \bar{\btheta}_5 &= (3,1,2,-1,-2,1,2,-1,-1,2)^T.
\end{align}
And we generate the parameter $\bthetaks{k}_r$ by
\begin{equation}
	\bthetaks{k}_r = \bar{\btheta}_r + h\times \frac{\bm{z}^{(k)}_r}{\twonorm{\bm{z}^{(k)}_r}}, \quad r \in [R],
\end{equation}
where $\bm{z}_r \sim N(\bm{0}_p, \bm{I}_{p \times p})$ and they are independent for different $k$ and $r$. In the GMM simulation, the observations of the outlier task are generated i.i.d. from $N(2\cdot \bm{1}_p, 3\bm{I}_{p \times p})$. In the MoR simulation, all $\bxk{k}_i$'s are generated from $N(\bm{0}_p, \bm{I}_{p \times p})$, and the responses $\{\yk{k}_i\}_{i=1}^n$ of the outlier task are generated i.i.d. from the regression model $\yk{k}_i = (\bxk{k}_i)^T\bthetaks{k}_r + \epsilonk{k}_i$ with $\bthetaks{k}_r = 3\cdot \bm{1}_p$ and $\epsilonk{k}_i \sim N(0, 1)$. In Figure \ref{fig: simulation}, the SNR is defined to be $\min_{k \in S}\max_{r\neq r' \in [R]}\twonorm{\bthetaks{k}_r - \bthetaks{k}_{r'}}$.

In FedGrEM, we set the number of iterations $T = 1000$, and the penalty parameters $\lambdat{t}$ are updated in iterations as follows:
\begin{align}
	\lambdat{0} = 1, \quad \lambdat{t} = \kappa_0\lambdat{t-1} + C\sqrt{p+\log K},
\end{align}
where $\kappa = 0.1$ and $C = 2$. 

The optimization in the central update is solved by an alternating optimization procedure. We first reparameterize the problem by $\bm{\nu}^{(k)} = \bar{\bm{\nu}} + \bm{\Delta}^{(k)}$, and we solve 
\begin{equation}\label{eq: repara}
	\argmin\limits_{\{\bm{\Delta}^{(k)}\}_{k=1}^K \subseteq \mathbb{R}^d, \overline{\bm{\nu}} \in \mathbb{R}^d}\bigg\{\sum_{k=1}^K\Big(\frac{n}{2}\twonorm{\bar{\bm{\nu}} + \bm{\Delta}^{(k)} - \tthetakt{k}{t}_r}^2+\sqrt{n}\lambdat{t}\cdot \twonorm{\bm{\Delta}^{(k)}}\Big)\bigg\},
\end{equation}
and assign $\hthetakt{k}{t}_r = \bar{\bm{\nu}} + \bm{\Delta}^{(k)}$. We solve \eqref{eq: repara} by the alternating optimization as follows:
\begin{enumerate}[(i)]
	\item Set $\bm{\Delta}^{(k)} = \bm{0}$ for all $k \in [K]$;
	\item Fix $\{\bm{\Delta}^{(k)}\}_{k=1}^K$, and update $\bar{\bm{\nu}} = \frac{1}{K}\sum_{k=1}^K (\tthetakt{k}{t}_r - \bm{\Delta}^{(k)})$;
	\item Fix $\bar{\bm{\nu}}$, and update $\bm{\Delta}^{(k)}$ as
		\begin{align}
			\bm{\Delta}^{(k)} &= \argmin_{\bm{\Delta}} \Big\{\frac{n}{2}\twonorm{\bar{\bm{\nu}} + \bm{\Delta} - \tthetakt{k}{t}_r}^2+\sqrt{n}\lambdat{t}\cdot \twonorm{\bm{\Delta}}\Big\}\\
			&=\begin{cases}
				\bigg(1-\frac{\lambdat{t}/\sqrt{n}}{\twonorm{\tthetakt{k}{t}_r- \bar{\bm{\nu}}}}\bigg)(\tthetakt{k}{t}_r - \bar{\bm{\nu}}), \quad &\text{if }\twonorm{\tthetakt{k}{t}_r- \bar{\bm{\nu}}} \geq \frac{\lambdat{t}}{\sqrt{n}},\\
				\bm{0}, \quad &\text{else}.
			\end{cases}
		\end{align}
\end{enumerate}
We iterate $(\rom{2})$ and $(\rom{3})$ for a few times until convergence.

We also run our simulation example with different $C_{\eta}$ values ranging from 0.55 to 1.35. The results are summarized in the following Tables \ref{table: w C_eta tuning} and \ref{table: theta C_eta tuning}. It can be seen that the estimation error of mixture proportions $w^{(k)*}_r$'s is quite stable with the choice of $C_{\eta}$, and the estimation error of parameters $\btheta^{(k)*}_r$ first decreases then becomes stable when $C_{\eta}$ changes from 0.55 to 1.35. In fact, according to our observation in the experiments, if we continue increasing the $C_{\eta}$, the algorithm would fail to converge and output useless estimates.  Overall the performance is robust to the choice of $C_{\eta}$ within this range $[0.55, 1.35]$, and we can further tune it in practice, although the theoretically-guided choice already leads to a decent performance.

\begin{table*}[h!]
\centering
\begin{tabular}{c*{9}{c}}
\toprule
$C_{\eta}$  / $h$ & $h = 0$ & $h = 0.25$ & $h = 0.5$ & $h = 0.75$ & $h = 1$ & $h = 1.25$ & $h = 1.5$ \\
\midrule
0.55 & 0.072 (0.017) & 0.071 (0.015) & 0.068 (0.015) & 0.071 (0.016) & 0.072 (0.017) & 0.071 (0.017) & 0.071 (0.016) \\
0.65 & 0.072 (0.017) & 0.071 (0.015) & 0.068 (0.015) & 0.071 (0.016) & 0.072 (0.017) & 0.071 (0.017) & 0.071 (0.016)\\
0.75 & 0.072 (0.017) & 0.071 (0.018) & 0.068 (0.015) & 0.071 (0.016) & 0.072 (0.017) & 0.071 (0.019) & 0.071 (0.015) \\
0.85 & 0.072 (0.017) & 0.071 (0.015) & 0.068 (0.015) & 0.071 (0.016) & 0.072 (0.019) & 0.071 (0.018) & 0.071 (0.015)  \\
0.95 & 0.072 (0.017) & 0.071 (0.016) & 0.068 (0.015) & 0.071 (0.016) & 0.072 (0.02) & 0.071 (0.018) & 0.071 (0.015)  \\
1.05 & 0.072 (0.017) & 0.071 (0.016) & 0.068 (0.015) & 0.071 (0.016) & 0.071 (0.017) & 0.071 (0.017) & 0.071 (0.015)  \\
1.15 & 0.072 (0.017) & 0.071 (0.016) & 0.069 (0.017) & 0.071 (0.016) & 0.071 (0.017) & 0.071 (0.017) & 0.071 (0.016)  \\
1.25 & 0.072 (0.017) & 0.071 (0.016) & 0.068 (0.015) & 0.071 (0.017) & 0.071 (0.017) & 0.072 (0.019) & 0.071 (0.017) \\
1.35 & 0.073 (0.021) & 0.071 (0.016) & 0.069 (0.017) & 0.071 (0.016) & 0.071 (0.017) & 0.072 (0.018) & 0.071 (0.016)  \\
\bottomrule
\end{tabular}
\caption{Average of the maximum estimation error of $\{w^{(k)*}_r\}_{k \in S, r}$ (standard deviations) in $\log_e$ scale in the GMM simulation, with different constants $C_{\eta}$ in learning rate $\eta^{(k)}_r = C_{\eta}/\hat{w}^{(k)[0]}_r$ and heterogeneity parameter $h$.}
\label{table: w C_eta tuning}
\end{table*}

\begin{table*}[!ht]
\centering
\begin{tabular}{@{} c|cccccccc @{}} 
 \toprule 
 $C_{\eta}$/ $h$ &    $h = 0$    &  $h = 0.25$   &   $h = 0.5$   &  $h = 0.75$   &    $h = 1$    &  $h = 1.25$   &   $h = 1.5$   & $h = 1.75$   \\
 \midrule 
     0.55        & 1.36 (0.62) & 1.47 (0.63)  & 1.62 (0.63) & 1.71 (0.56) & 1.89 (0.70) & 2.07 (0.72) & 2.23 (0.84) & 2.44 (0.99)  \\
    0.65        & 1.28 (0.59) & 1.38 (0.61) & 1.52 (0.62)  & 1.62 (0.56)  & 1.79 (0.71) & 1.99 (0.75)  & 2.14 (0.90) & 2.33 (0.99)\\
  0.75        & 1.22 (0.55) & 1.34 (0.63) & 1.44 (0.59)  & 1.54 (0.55)  & 1.71 (0.69) & 1.93 (0.78) & 2.02 (0.81) &2.25 (1.01) \\
    0.85        & 1.17 (0.51)  & 1.30 (0.68) & 1.38 (0.57) & 1.49 (0.54)  & 1.66 (0.68) & 1.88 (0.77) & 1.95 (0.79) & 2.22 (1.05)\\
   0.95        & 1.13 (0.47) & 1.30 (0.80) & 1.34 (0.56) & 1.45 (0.52) &  1.60 (0.61)  & 1.82 (0.74) &  1.90 (0.77)  & 2.14 (1.05)  \\
   1.05        & 1.12 (0.44) & 1.27 (0.75) & 1.30 (0.51) & 1.42 (0.48) & 1.54 (0.52)  & 1.79 (0.73) & 1.85 (0.74) & 2.07 (0.97)\\
   1.15        &  1.10 (0.41)  & 1.25 (0.77) & 1.32 (0.66) & 1.39 (0.46)  &  1.52 (0.50)  & 1.79 (0.79) & 1.82 (0.72) & 2.04 (0.98)\\
    1.25        & 1.09 (0.38) & 1.23 (0.76) & 1.34 (0.71) & 1.40 (0.59) & 1.50 (0.48) & 1.78 (0.81)  & 1.81 (0.75) & 1.99 (0.96) \\
    1.35        & 1.21 (1.42) & 1.24 (0.81) & 1.31 (0.57) & 1.36 (0.41) & 1.49 (0.48) & 1.81 (0.91) & 1.79 (0.74) & 1.99 (1.00) \\ \bottomrule
\end{tabular}
\caption{Average of the maximum estimation error of $\{\btheta^{(k)*}_r\}_{k \in S, r}$ (standard deviations) in $\log_e$ scale in the GMM simulation, with different constants $C_{\eta}$ in learning rate $\eta^{(k)}_r = C_{\eta}/\hat{w}^{(k)[0]}_r$ and heterogeneity parameter $h$.}
\label{table: theta C_eta tuning}
\end{table*}

\subsection{Additional Details of Real Studies}\label{subsec: real studies}
Due to the high dimensionality of the MNIST and Fashion-MNIST datasets, we first applied tSNE \citep{hinton2002stochastic, van2008visualizing} to reduce the dimension to 10 then performed all methods on the transformed datasets. In each replication, 80\% data for each task is used as training data and the remaining 20\% is used as test data to calculate the mis-clustering error. We also contaminate different proportions of tasks to showcase the robustness of FedGrEM against adversarial attacks.  

We record the average computational time for each method and the results are summarized below. The experiments were conducted with Dual Intel Xeon Gold 6226R processors (2.9 GHz) and a single core. The results are summarized in Tables \ref{table: time pen}, \ref{table: time mnist}, and \ref{table: fmnist}. From the performance and computational time, we can see that FedGrEM can be adapted to a federated learning environment with hundreds of nodes, even without parallel computing. By parallelizing the local update steps in each iteration, we can further speed things up. 

The reason why gradient EM-based methods are slower than full EM-based methods is that the M-step of full EM has an explicit expression and does not require calculating the matrix inverse since all covariances are identities. Full EM-based methods do not apply to the problems with non-explicit M-steps and would be more time-consuming than gradient EM-based methods if the explicit M-step expression is very complicated. 

\begin{table*}[h!]
\centering
\begin{tabular}{c*{7}{c}}
\toprule
$\epsilon$ / Method & Local-EM & Local-GrEM & FedEM  & FedGrEM & TGMM  & Pooled-EM & Pooled-GrEM \\
\midrule
0\% & 26.01 (1.22) & 36.55 (1.58) & 29.86 (1.48) & 47.07 (2.38) & 86.79 (3.69) & 18.85 (2.10) & 30.24 (2.97) \\
6.8\% & 25.92 (1.04) & 36.19 (1.41) & 29.74 (1.27) & 46.76 (1.92) & 85.07 (3.57) & 18.28 (1.66) & 29.71 (2.23) \\
13.6\% & 25.94 (1.46) & 36.07 (1.46) & 29.78 (1.54) & 46.90 (2.31) & 84.85 (3.73) & 17.81 (1.47) & 29.39 (2.00) \\
20.5\% & 25.62 (1.05) & 35.80 (1.48) & 29.49 (1.21) & 46.49 (2.12) & 83.84 (3.65) & 17.36 (1.35) & 28.56 (2.14) \\
\bottomrule
\end{tabular}
\caption{Average computational time (standard deviations) in seconds for Pen-Based Recognition of Handwritten Digits dataset.}
\label{table: time pen}
\end{table*}

\begin{table*}[h!]
\centering
\resizebox{\columnwidth}{!}{\begin{tabular}{c*{7}{c}}
\toprule
$\epsilon$ / Method & Local-EM & Local-GrEM & FedEM & FedGrEM & TGMM  & Pooled-EM & Pooled-GrEM \\
\midrule
0\%  & 83.99 (4.07) & 113.81 (6.59) & 93.00 (4.85) & 134.69 (6.41) & 299.56 (19.18) & 94.87 (12.26) & 166.12 (27.84) \\
8\%  & 80.63 (3.92) & 110.90 (5.31) & 89.36 (4.57) & 132.49 (6.63) & 291.81 (17.09) & 93.51 (13.56) & 159.38 (28.76) \\
16\% & 80.06 (4.08) & 110.27 (5.51) & 88.52 (4.82) & 133.32 (7.12) & 290.71 (17.95) & 89.39 (14.91) & 151.79 (27.51) \\
24\% & 79.32 (3.82) & 109.31 (5.20) & 90.16 (4.55) & 133.74 (6.47) & 287.80 (15.12) & 86.54 (15.00) & 145.39 (25.01) \\
\bottomrule
\end{tabular}}
\caption{Average computational time (standard deviations) in seconds for MNIST dataset.}
\label{table: time mnist}
\end{table*}

\begin{table*}[h!]
\centering
\resizebox{\columnwidth}{!}{\begin{tabular}{c*{7}{c}}
\toprule
$\epsilon$ / Method & Local-EM & Local-GrEM & FedEM & FedGrEM & TGMM & Pooled-EM & Pooled-GrEM \\
\midrule
0\%  & 83.89 (3.58) & 113.21 (4.94) & 92.82 (4.10) & 134.92 (5.98) & 297.01 (13.73) & 96.46 (14.15) & 169.07 (32.25) \\
8\%  & 81.15 (3.54) & 111.78 (4.89) & 89.41 (4.08) & 133.29 (6.01) & 293.78 (14.59) & 94.52 (15.28) & 162.50 (30.77) \\
16\% & 80.70 (3.56) & 111.28 (4.78) & 89.80 (4.16) & 134.38 (6.17) & 293.31 (15.13) & 90.28 (15.88) & 154.20 (29.21) \\
24\% & 80.39 (3.46) & 110.75 (4.59) & 90.91 (4.17) & 135.52 (5.95) & 292.91 (13.93) & 87.20 (15.59) & 147.38 (27.98) \\
\bottomrule
\end{tabular}}
\caption{Average computational time (standard deviations) in seconds for Fashion-MNIST dataset.}
\label{table: fmnist}
\end{table*}

\section{Label Permutation}\label{sec: label permutation}
Denote the family of permutation functions on $[R]$ as $\mathcal{P}^R$.

\subsection{Label Permutation in Initialization}

One challenge that hinders the practical application of FedGrEM (and other federated EM algorithms) and our theoretical framework relates to the initialization condition outlined in Assumption \ref{asmp: thm generic} (more rigorously, Assumption \ref{asmp: thm generic appendix}). As is common in many unsupervised learning problems, the parameters are estimated up to a label permutation. Our results still hold if the initialization condition holds up to a permutation, i.e. there exists a permutation $\pi \in \mathcal{P}^R$ such that $\max\limits_{k \in S, r \in [R]}\norm{\hw^{(k)[0]}_{\pi(r)}- \wks{k}_r} \leq r_w^*, \max\limits_{k \in S, r \in [R]}\twonorm{\htheta^{(k)[0]}_{\pi(r)} - \bthetaks{k}_r} \leq r_{\btheta}^*$, however, it necessitates the presence of a \textit{shared} permutation $\pi \in \mathcal{P}^R$ among the tasks in $S$. However, in all clustering methods typically employed for initialization in individual tasks, the estimations are inherently invariant to permutations, and different tasks may yield distinct permutations. To the best of our knowledge, there has been limited discussion in the existing literature on unsupervised FDL regarding this issue, with the exception of \citet{tian2022unsupervised}. We generalized the solutions proposed in \citet{tian2022unsupervised} to address the permutation issue in initialization. By ensuring that different tasks in $S$ share the same permutation, we make FedGrEM and our accompanying theory applicable in practice.

\subsection{Alignment Algorithms}
We define the following score function of the permutations $\bm{\pi} = \{\pi_k\}_{k=1}^K \in (\mathcal{P}^R)^{\otimes K}$:

\begin{equation}
	\text{score}(\bm{\pi}, K) = \sum_{r=1}^R \sum_{k \neq k' \in [K]}\twonorm{\hthetakt{k}{0}_{\pi_k(r)} - \hthetakt{k}{0}_{\pi_{k'}(r)}}.
\end{equation}

Intuitively, the score is smaller if the permutations $\bm{\pi} = \{\pi_k\}_{k=1}^K \in (\mathcal{P}^R)^{\otimes K}$ are more aligned, serving as the basis for adjusting the permutations of each task. We also define the best permutation for task $k \in S$ as

\begin{equation}
	\pi_k^* = \argmin\limits_{\pi_k \in \mathcal{P}^R} \sum_{r=1}^R\twonorm{\hthetakt{k}{0}_{\pi_k(r)} - \bthetaks{k}_r}.
\end{equation}

Next, we introduce an exhaustive search algorithm for permutation alignment, which seeks the permutations minimizing the score.  

\underline{\textbf{Permutation Alignment Algorithm 1 (Exhaustive search)}:} Let $\widehat{\bm{\pi}} = \argmin\limits_{\bm{\pi} \in (\mathcal{P}^R)^{\otimes K}}\text{score}(\bm{\pi}, K)$.

Under the assumption detailed below, we demonstrate that the output permutations in tasks of $S$ from the exhaustive search algorithm are well-aligned.
\begin{assumption}\label{asmp: exhaustive}
	The following conditions hold:
	\begin{enumerate}[(i)]
		\item $\Delta > \frac{2+2\epsilon}{1-\epsilon}h+\frac{4+4\epsilon}{1-\epsilon}\max_{k\in S}\min_{\pi_k \in \mathcal{P}^R} \max_{r \in [R]}\allowbreak \twonorm{\hthetakt{k}{0}_{\pi_k(r)} - \bthetaks{k}_r}$;
		\item $\epsilon < 1/2$.
	\end{enumerate}
\end{assumption}

\begin{theorem}\label{thm: exhaustive search appendix}
	Under Assumption \ref{asmp: exhaustive}, for Alignment Algorithm 1 (Exhaustive search), there exists a permutation $\iota \in \mathcal{P}^R$ such that $\widehat{\pi}_k = \iota \circ \pi_k^*$ for all $k \in S$.
\end{theorem}

The computational cost of the exhaustive search algorithm is $\mathcal{O}((R!)^K\cdot K^2R)$ as it explores all the possible permutations on $[R]$ for all $K$ tasks and takes $\mathcal{O}(K^2R)$ time to calculate the score for each permutation. We introduce the following stepwise search algorithm which can reduce the computational cost to $\mathcal{O}(R! K\cdot K^2R)$.

\underline{\textbf{Permutation Alignment Algorithm 2 (Stepwise search)}:} For \allowbreak $k =1:K$, with $\{\widehat{\pi}_{k'}\}_{k'=1}^{k-1}$ fixed, set $\widehat{\pi}_k = \argmin\limits_{\pi_k \in \mathcal{P}^R} \allowbreak \text{score}(\{\widehat{\pi}_{k'}\}_{k'=1}^{k-1} \cup \pi_k, k)$. Finally, let $\widehat{\bm{\pi}} = \{\widehat{\pi}_k\}_{k=1}^K$.

Under the following assumption, we show that the output permutations in tasks of $S$ from the stepwise search algorithm are well-aligned.
\begin{assumption}\label{asmp: stepwise}
	Suppose there are no outlier tasks in the first $K_0$ tasks and
	\begin{enumerate}[(i)]
		\item $\Delta > 2\frac{K_0+K\epsilon}{K_0-K\epsilon}h+6\frac{K_0+K\epsilon}{K_0-K\epsilon}\max_{k\in S}\min_{\pi_k \in \mathcal{P}^R} \allowbreak\max_{r \in [R]} \twonorm{\hthetakt{k}{0}_{\pi_k(r)} - \bthetaks{k}_r}$;
		\item $K_0 > K\epsilon$;
		\item $\epsilon < 1/2$.
	\end{enumerate}
\end{assumption}

\begin{theorem}\label{thm: stepwise search appendix}
	Under Assumption \ref{asmp: stepwise}, there exists a permutation $\iota \in \mathcal{P}^R$ such that $\widehat{\pi}_k = \iota \circ \pi_k^*$ for all $k \in S$.
\end{theorem}

The limitation of the stepwise search algorithm is that it requires the first $K_0$ tasks to be non-outlier tasks. One potential solution is running the algorithm multiple times with a random shuffling of tasks and then picking the permutations based on recurring patterns observed across multiple experiment runs. We leave a full investigation of this random algorithm for future study.

\section{Additional Discussions}\label{sec: more discussions appendix}
We want to comment a bit more about our FedGrEM algorithm.
\begin{itemize}
	\item When mixture proportions $\{w^{(k)*}_r\}_{k=1}^K$ are also similar across tasks, we can apply the same aggregation by regularization in the central update to further improve the performance of FedGrEM. The same analysis tools can be applied to obtain stronger results.
	\item When there exist various computational capabilities across different tasks, we can replace the current local gradient descent with full local data by local stochastic gradient descent (SGD) with small batches of local data. This would decrease the computational cost for each task. Moreover, instead of including all users in each iteration round, we can randomly sample a few users in each round to update their local estimates and run the central update with only these active users. These two approaches could further decrease the computational cost. And the size of SGD batches and the proportion of active users in each round could depend on the computational and communicational budget for each user, which could differ from user to user.
	\item The current communicational cost for FedGrEM or FedEM \citep{marfoq2021federated} is already low because they only pass the gradients across tasks in each iteration. The sampling of active users could help further decrease the communication cost. 
	\item Our current analysis and the method FedGrEM can be extended to a fully decentralized version with the same idea used by Algorithm 4 in \cite{marfoq2021federated}. In each iteration, instead of sending all local estimates to the central server, each node can send their local estimates only to their neighbors (the nodes that are closest to them in the geometric sense or the communication cost sense) and perform the aggregation locally with the estimates received from the neighbors. We can use the current theoretical framework to analyze the estimation error of this fully decentralized algorithm and derive similar results.
	\item We can consider other types of attacks and contaminations. For example, instead of the corruption of the entire dataset from some users, we can assume partial observations are contaminated. In this case, we can create a robust version of local estimates by using truncated gradients, similar to the gradient clipping used in differential privacy \cite{varshney2022nearly}. Aggregating these robust local estimates in the central update can make the whole procedure robust to both observation-level and user-level attacks.
	\item We assume that the central update can be exactly solved and we use alternating optimization to solve it in practice. We did not directly analyze the alternating optimization itself because the EM procedure is already very complicated to study due to its iterative nature. We believe that this is an important question. One nice characteristic about our central update in FedGrEM (Algorithm \ref{algo: FG-EM}) is that it is a convex problem. There exist convergence results about alternating optimization, such as \cite{li2019alternating, guminov2021combination, tupitsa2021alternating}, which can be helpful. We can also consider other optimization methods such as proximal gradient descent \citep{polson2015proximal}. We will work on this problem in the future.
\end{itemize}

We also want to point out the possibility of generalizing our theory and method to high-dimensional or non-i.i.d. data.
\begin{itemize}
	\item For i.i.d. high-dimensional data, which is very common in healthcare and biomedical studies, we can add an additional regularization term (e.g., $\ell_1$-penalty) for the global estimator $\bar{\bm{\nu}}$ in the central update of FedGrEM (Algorithm \ref{algo: FG-EM}). Another solution is to truncate the current local estimator (one-step gradient descent) by a coordinate-wise soft-thresholding function and keep the central update as it is. The challenges of analyzing such a problem are mainly aligned with the challenges in other high-dimensional problems. For example, the strong concavity would fail for the empirical surrogate risk function $\widehat{Q}^{(k)}$. Instead, as one of the standard techniques used in the high-dimensional analysis, we need to first prove that the estimators belong to a small subset (usually a cone in $\mathbb{R}^d$), and within this subset, the so-called restricted strong convexity or concavity (RSC) holds. In the context of federated EM algorithms, the analysis would be more complicated due to the nature of the iterative procedure. Some analysis has been done for the single-task EM in \cite{cai2019chime}, and some techniques therein might be helpful.
	\item For non-i.i.d. data, for example, the data of social networks, we can first apply some embedding methods to transform the original data into a standard unsupervised learning problem. For instance, for adjacency or Laplacian matrices in social networks, we can compute its spectral embedding and use the embedding as the input for the federated EM algorithms. The challenge here is that the embedded data is not independent. But in many situations, the dependence within the embedded data can be shown to be somewhat weak, which is sufficient to derive some theoretical guarantees \citep{rohe2011spectral, tang2018limit, abbe2020entrywise}.
\end{itemize}

\section{Proofs}\label{sec: proofs appendix}

\subsection{Proof of Theorem \ref{thm: generic appendix}}
Let us first fix an $S \subseteq [K]$ and introduce the following key lemma. 
\begin{lemma}\citep{duan2022adaptive}\label{lem: duan}
	The following results hold:
	\begin{enumerate}[(i)]
		\item If $\lambdat{t} \geq \frac{5\sqrt{n}\max_{k \in S}\twonorm{\widetilde{\btheta}^{(k)[t]}_r - \bthetaks{k}_r}}{1-2\epsilon}$, then
			\begin{equation}
				\max_{k \in S}\twonorm{\hthetakt{k}{t}_r - \bthetaks{k}_r} \leq \frac{1}{|S|}\twonorma{\sum_{k \in S}(\widetilde{\btheta}^{(k)[t]}_r - \bthetaks{k}_r)} + \frac{6}{1-2\epsilon}\min\Big\{3h, \frac{2\lambdat{t}}{5\sqrt{n}}\Big\} + \frac{2\lambdat{t}}{\sqrt{n}}\epsilon.
			\end{equation}
		\item If we further have $\lambdat{t} \geq \frac{15\sqrt{n}}{1-2\epsilon}h$, then
			\begin{equation}
				\max_{k \in S}\twonorm{\hthetakt{k}{t}_r - \bthetaks{k}_r} \leq \frac{1}{|S|}\twonorma{\sum_{k \in S}(\widetilde{\btheta}^{(k)[t]}_r - \bthetaks{k}_r)} + 2h + \frac{2\lambdat{t}}{\sqrt{n}}\epsilon.
			\end{equation}
	\end{enumerate}
\end{lemma}

Define a random event $\mathcal{V}$ which is the intersection of the following three events:
\begin{itemize}
	\item The event in Assumption \ref{asmp: w appendix}.(\rom{2}) holds for all $k \in S$ with failure probability  $\frac{\delta}{3RK}$;
	\item The event in Assumption \ref{asmp: w appendix}.(\rom{2}) holds for all $k \in S$ with failure probability  $\frac{\delta}{3RK}$;
	\item The event in Assumption \ref{asmp: theta appendix}.(\rom{3}) holds with failure probability $\frac{\delta}{3}$.
\end{itemize} 
Then by the union bound, $\tP(\mathcal{V}) \geq 1-\delta$. In the following analysis, we condition on $\mathcal{V}$. Hence all arguments hold with probability at least $1-\delta$.

(\Rom{1}) Part 1: Iteration round $t = 1$.

By Lemma \ref{lem: duan}, when $\lambdat{t} \geq \frac{5\sqrt{n}}{1-2\epsilon}\max_{k \in S}\max_{r \in [R]}\twonorm{\widetilde{\btheta}^{(k)[1]}_r - \bthetaks{k}_r}$:
\begin{equation}
	\max_{k \in S}\twonorm{\hthetakt{k}{1}_r - \bthetaks{k}_r} \leq \frac{1}{|S|}\twonorma{\sum_{k \in S}(\widetilde{\btheta}^{(k)[1]}_r - \bthetaks{k}_r)} + \frac{6}{1-2\epsilon}\min\Big\{3h, \frac{2\lambdat{1}}{5\sqrt{n}}\Big\} + \frac{2\lambdat{1}}{\sqrt{n}}\epsilon.
\end{equation}
Note that
\begin{align}
	\frac{1}{|S|}\twonorma{\sum_{k \in S}(\widetilde{\btheta}^{(k)[1]}_r - \bthetaks{k}_r)} &\leq \underbrace{\frac{1}{|S|}\twonorma{\sum_{k \in S}\bigg[\hthetakt{k}{0}_r - \bthetaks{k}_r+ \etak{k}_r\frac{\partial}{\partial \btheta_r} \Qk{k}(\hthetakt{k}{0}|\hwkt{k}{0}, \hthetakt{k}{0})\bigg]}}_{[1]} \\
	&\quad + \underbrace{\frac{1}{|S|}\twonorma{\sum_{k \in S}\etak{k}_r\bigg[\frac{\partial}{\partial \btheta_r} \Qk{k}(\hthetakt{k}{0}|\hwkt{k}{0}, \hthetakt{k}{0})-\frac{\partial}{\partial \btheta_r} \hQk{k}(\hthetakt{k}{0}|\hwkt{k}{0}, \hthetakt{k}{0})\bigg]}}_{[2]}.
\end{align}
For [1], we have
\begin{align}
	[1] &\leq \frac{1}{|S|}\twonorma{\sum_{k \in S}\bigg[\hthetakt{k}{0}_r - \bthetaks{k}_r+ \etak{k}_r\frac{\partial}{\partial \btheta_r} \qk{k}(\hthetakt{k}{0})\bigg]} \\
	&\quad + \frac{1}{|S|}\twonorma{\sum_{k \in S}\etak{k}_r\bigg[\frac{\partial}{\partial \btheta_r} \Qk{k}(\hthetakt{k}{0}|\hwkt{k}{0}, \hthetakt{k}{0})-\frac{\partial}{\partial \btheta_r} \qk{k}(\hthetakt{k}{0})\bigg]} \\
	&\leq \frac{1}{|S|}\sum_{k \in S}\sqrt{1-\etak{k}_r\muk{k}_r}\twonorm{\hthetakt{k}{0}-\bthetaks{k}_r} + \frac{1}{|S|}\sum_{k \in S}\gamma\cdot \sum_{r=1}^R\etak{k}_r(\norm{\hwkt{k}{0}_r - \wks{k}_r} + \twonorm{\hthetakt{k}{0}_r - \bthetaks{k}_r}) \\
	&\leq \bigg(\sqrt{1-\min_{k \in S, r\in [R]}(\etak{k}_r\muk{k}_r)} + \gamma\bar{\eta}R\bigg)\max_{k \in S}\max_{r \in [R]}\twonorm{\hthetakt{k}{0}_r - \bthetaks{k}_r} + \gamma\bar{\eta}R\cdot \max_{k \in S}\max_{r \in [R]}\norm{\hwkt{k}{0}_r - \wks{k}_r},
\end{align}
where the first part of the second inequality comes from the classical result of gradient descent (e.g., see Theorem 3.4 in \citet{lan2020first}). 

Similarly, we can show that
\begin{equation}
	\max_{k \in S}\twonorm{\widetilde{\btheta}^{(k)[1]}_r - \bthetaks{k}_r} \leq \kappa_0 G^{[0]} + \bar{\eta}\mathcal{E}_1\Big(n, \frac{\delta}{3RK}\Big).
\end{equation}
Therefore $\lambdat{1} = \widetilde{\kappa}_0\lambdat{0} +15\sqrt{n}[\mathcal{W}(n, \frac{\delta}{3RK}, r_{\btheta}^*) + 2\bar{\eta}\mathcal{E}_1(n, \frac{\delta}{3RK})] = \frac{15}{119}\widetilde{\kappa}_0\sqrt{n}(r_w^* + r_{\btheta}^*) +15\sqrt{n}[\mathcal{W}(n, \frac{\delta}{3RK}, r_{\btheta}^*) + 2\bar{\eta}\mathcal{E}_1(n, \frac{\delta}{3RK})]\geq \frac{5\sqrt{n}}{1-2\epsilon}\max_{k \in S}\twonorm{\widetilde{\btheta}^{(k)[1]}_r - \bthetaks{k}_r}$ indeed holds, where $\widetilde{\kappa}_0 = 119\Big[\sqrt{1-\min_{k \in S, r\in [R]}(\etak{k}_r\muk{k}_r)} + \gamma\bar{\eta}R + \kappa R\Big]$.

For [2], we have
\begin{equation}
	[2] \leq \bar{\eta}\mathcal{E}_1\Big(n, \frac{\delta}{3RK}\Big).
\end{equation}
Combine the bounds of [1] and [2]:
\begin{align}
	\max_{k \in S}\max_{r \in [R]}\twonorm{\hthetakt{k}{1}_r - \bthetaks{k}_r} &\leq \bigg(\sqrt{1-\min_{k \in S, r\in [R]}(\etak{k}_r\muk{k}_r)} + \gamma\bar{\eta}R\bigg)\max_{k \in S}\max_{r \in [R]}(\twonorm{\hthetakt{k}{0}_r - \bthetaks{k}_r} + \norm{\hwkt{k}{0}_r - \wks{k}_r}) \\
	&\quad + \frac{6}{1-2\epsilon}\min\Big\{3h, \frac{2\lambdat{1}}{5\sqrt{n}}\Big\} + \frac{2\lambdat{1}}{\sqrt{n}}\epsilon + \bar{\eta}\mathcal{E}_1\Big(n, \frac{\delta}{3RK}\Big).
\end{align}
On the other hand,
\begin{align}
	\max_{k \in S}\max_{r \in [R]}\norm{\hwkt{k}{0}_r - \wks{k}_r} &\leq \max_{k \in S}\max_{r \in [R]}\norma{\frac{1}{n}\sum_{i=1}^n \tP(\zk{k} = r| \bxk{k}_i, \hwkt{k}{0}, \hthetakt{k}{0}) - \tE\big[\tP(\zk{k} = r| \bxk{k}_i, \hwkt{k}{0}, \hthetakt{k}{0})\big]} \\
	&\quad + \max_{k \in S}\max_{r \in [R]}\norma{\tE\big[\tP(\zk{k} = r| \bxk{k}_i, \hwkt{k}{0}, \hthetakt{k}{0})\big] - \wks{k}_r} \\
	&\leq \mathcal{W}\Big(n, \frac{\delta}{3RK}, r_{\btheta}^*\Big) + \kappa \max_{k \in S}\sum_{r=1}^R (\twonorm{\hthetakt{k}{0}_r - \bthetaks{k}_r} + \norm{\hwkt{k}{0}_r - \wks{k}_r}) \\
	&\leq \mathcal{W}\Big(n, \frac{\delta}{3RK}, r_{\btheta}^*\Big) + \kappa R \max_{k \in S}\max_{r \in [R]} (\twonorm{\hthetakt{k}{0}_r - \bthetaks{k}_r} + \norm{\hwkt{k}{0}_r - \wks{k}_r}). \label{eq: w, t = 1}
\end{align}
As a result,
\begin{align}
	&\max_{k \in S}\max_{r \in [R]}(\twonorm{\hthetakt{k}{1}_r - \bthetaks{k}_r} + \norm{\hwkt{k}{1}_r - \wks{k}_r})\\
    &\leq \bigg(\sqrt{1-\min_{k \in S, r\in [R]}(\etak{k}_r\muk{k}_r)} + \gamma\bar{\eta}R + \kappa R\bigg)\max_{k \in S}\max_{r \in [R]}(\twonorm{\hthetakt{k}{0}_r - \bthetaks{k}_r} + \norm{\hwkt{k}{0}_r - \wks{k}_r}) \\
    &\quad + \frac{6}{1-2\epsilon}\min\Big\{3h, \frac{2\lambdat{1}}{5\sqrt{n}}\Big\} + \frac{2\lambdat{1}}{\sqrt{n}}\epsilon + \bar{\eta}\mathcal{E}_1\Big(n, \frac{\delta}{3RK}\Big) + \mathcal{W}\Big(n, \frac{\delta}{3RK}, r_{\btheta}^*\Big).
\end{align}
Denote $G^{[t]} = \max_{k \in S}\max_{r \in [R]}(\twonorm{\hthetakt{k}{t}_r - \bthetaks{k}_r} + \norm{\hwkt{k}{t}_r - \wks{k}_r})$ and $\kappa_0 = \sqrt{1-\min_{k \in S, r\in [R]}(\etak{k}_r\muk{k}_r)} + \gamma\bar{\eta}R + \kappa R$. Then
\begin{align}
	G^{[1]} &\leq \kappa_0 G^{[0]} + \frac{6}{1-2\epsilon}\min\Big\{3h, \frac{2\lambdat{1}}{5\sqrt{n}}\Big\} + \frac{2\lambdat{1}}{\sqrt{n}}\epsilon + \bar{\eta}\mathcal{E}_1\Big(n, \frac{\delta}{3RK}\Big) + \mathcal{W}\Big(n, \frac{\delta}{3RK}, r_{\btheta}^*\Big) \\
	&\leq \kappa_0 G^{[0]} +\bigg[\frac{12}{5(1-2\epsilon)}+2\epsilon\bigg]\frac{\lambdat{1}}{\sqrt{n}} + \bar{\eta}\mathcal{E}_1\Big(n, \frac{\delta}{3RK}\Big) + \mathcal{W}\Big(n, \frac{\delta}{3RK}, r_{\btheta}^*\Big) \\
	&\leq \kappa_0 G^{[0]} +\frac{118}{15}\cdot \frac{\lambdat{1}}{\sqrt{n}} + \bar{\eta}\mathcal{E}_1\Big(n, \frac{\delta}{3RK}\Big) + \mathcal{W}\Big(n, \frac{\delta}{3RK}, r_{\btheta}^*\Big).
\end{align}

(\Rom{2}) Part 2: Iteration round $t \geq 2$.

Repeating the analysis in (\Rom{1}), we can see that when $\lambdat{t} \geq 15\sqrt{n} \kappa_0G^{[t-1]} + 15\sqrt{n}\bar{\eta}\mathcal{E}_1(n, \frac{\delta}{3RK})\geq 15\sqrt{n}\Big(\sqrt{1-\min_{k \in S, r\in [R]}(\etak{k}_r\muk{k}_r)} + \gamma\bar{\eta}R\Big)G^{[t-1]} + 15\sqrt{n}\bar{\eta}\mathcal{E}_1(n, \frac{\delta}{3RK})$,
\begin{align}
	G^{[t]} &\leq \kappa_0 G^{[t-1]} +\frac{118}{15}\cdot \frac{\lambdat{t}}{\sqrt{n}} + \bar{\eta}\mathcal{E}_1\Big(n, \frac{\delta}{3RK}\Big) + \mathcal{W}\Big(n, \frac{\delta}{3RK}, r_{\btheta}^*\Big) \\
	&\leq \frac{119}{15}\cdot \frac{\lambdat{t}}{\sqrt{n}} + \bar{\eta}\mathcal{E}_1\Big(n, \frac{\delta}{3RK}\Big) + \mathcal{W}\Big(n, \frac{\delta}{3RK}, r_{\btheta}^*\Big). \label{eq: eq Gt}
\end{align}
Recall our setting of $\{\lambda^{[t]}\}_{t=1}^T$:
\begin{align}
	\lambdat{0} &= \frac{15}{119}\sqrt{n}(r_w^* + r_{\btheta}^*),\\
	\lambdat{t} &= \wtkappa_0\lambdat{t-1} + 15\sqrt{n}\Big[ \mathcal{W}\Big(n, \frac{\delta}{3RK}, r_{\btheta}^*\Big) + 2\bar{\eta}\mathcal{E}_1\Big(n, \frac{\delta}{3RK}\Big)\Big].
\end{align}
Hence $\lambdat{t} \geq 15\sqrt{n} \kappa_0G^{[t-1]} + 15\sqrt{n}\bar{\eta}\mathcal{E}_1(n, \frac{\delta}{3RK})$ indeed holds and
\begin{equation}\label{eq: lambda t}
	\lambdat{t} = (\widetilde{\kappa}_0)^t \lambdat{0} + \frac{1-(\widetilde{\kappa}_0)^t}{1-\widetilde{\kappa}_0}15\sqrt{n}\Big[\mathcal{W}\Big(n, \frac{\delta}{3RK}, r_{\btheta}^*\Big) + 2\bar{\eta}\mathcal{E}_1\Big(n, \frac{\delta}{3RK}\Big)\Big],
\end{equation}
which together with \eqref{eq: eq Gt} implies
\begin{equation}\label{eq: Gt large h}
	G^{[t]} \leq (\widetilde{\kappa}_0)^t (r_w^* + r_{\btheta}^*) + \bigg(\frac{119}{1-\widetilde{\kappa}_0}+1\bigg)\Big[ \mathcal{W}\Big(n, \frac{\delta}{3RK}, r_{\btheta}^*\Big) + 2\bar{\eta}\mathcal{E}_1\Big(n, \frac{\delta}{3RK}\Big)\Big] \leq r_{\btheta}^*,
\end{equation}
when $t \geq 1$. The last inequality holds due to Assumption \ref{asmp: thm generic appendix}. Similar to \eqref{eq: w, t = 1}, we have
\begin{align}
	\max_{k \in S}\max_{r \in [R]}\norm{\hwkt{k}{t}_r - \wks{k}_r} &\leq \mathcal{W}\Big(n, \frac{\delta}{3RK}, r_{\btheta}^*\Big) + \kappa R \cdot G^{[t-1]} \\
	&\leq \mathcal{W}\Big(n, \frac{\delta}{3RK}, r_{\btheta}^*\Big) + \frac{119}{15}(\widetilde{\kappa}_0)^{t-1} \kappa R(r_w^* + r_{\btheta}^*) \\
	&\quad + \kappa R\bigg(\frac{119}{1-\widetilde{\kappa}_0}+1\bigg)\Big[\mathcal{W}\Big(n, \frac{\delta}{3RK}, r_{\btheta}^*\Big) + 2\bar{\eta}\mathcal{E}_1\Big(n, \frac{\delta}{3RK}\Big)\Big] \\
	&\leq r_w^*,
\end{align}
where the last inequality is due to Assumption \ref{asmp: thm generic appendix}.

(\Rom{3}) Part 3: The case when $h \leq \frac{1}{3}[\mathcal{W}\big(n, \frac{\delta}{3RK}, r_{\btheta}^*\big) + 2\bar{\eta}\mathcal{E}_1\big(n, \frac{\delta}{3RK}\big)\big]$.

In this case, we have $\lambdat{t} \geq 15\sqrt{n}\big[\mathcal{W}\big(n, \frac{\delta}{3RK}, r_{\btheta}^*\big) + 2\bar{\eta}\mathcal{E}_1\big(n, \frac{\delta}{3RK}\big)\big] \geq \frac{15\sqrt{n}}{1-2\epsilon}h$ for $t \geq 1$. Then by Lemma \ref{lem: duan}.(\rom{2}), $\hthetakt{k}{t}$'s are equal for $k \in S$ when $t \geq 1$. Thus
\begin{equation}
	G^{[1]} \leq \frac{119}{15}\wtkappa_0 \lambdat{0} + \bigg(\frac{119}{1-\widetilde{\kappa}_0}+1\bigg)\Big[\mathcal{W}\Big(n, \frac{\delta}{3RK}, r_{\btheta}^*\Big) + 2\bar{\eta}\mathcal{E}_1\Big(n, \frac{\delta}{3RK}\Big)\Big],
\end{equation}
\begin{align}
	\max_{k \in S}\max_{r \in [R]}\twonorm{\hthetakt{k}{t}_r - \bthetaks{k}_r} &\leq \bigg(\sqrt{1-\min_{k \in S, r\in [R]}(\etak{k}_r\muk{k}_r)} + \gamma\bar{\eta}R\bigg)G^{[t-1]} + \bar{\eta}\mathcal{E}_2\Big(n, |S|, \frac{\delta}{3R}\Big) \\
	&\quad + \frac{6}{1-2\epsilon}\min\Big\{3h, \frac{2\lambdat{t}}{5\sqrt{n}}\Big\} + \frac{2\lambdat{t}}{\sqrt{n}}\epsilon,\\
	\max_{k \in S}\max_{r \in [R]}\norm{\hwkt{k}{t}_r - \wks{k}_r} &\leq \mathcal{W}\Big(n, \frac{\delta}{3RK}, r_{\btheta}^*\Big) + \kappa R G^{[t-1]},
\end{align}
which implies that
\begin{equation}
	G^{[t]} \leq \kappa_0G^{[t-1]} +  \bar{\eta}\mathcal{E}_2\Big(n, |S|, \frac{\delta}{3R}\Big) + \frac{6}{1-2\epsilon}\min\Big\{3h, \frac{2\lambdat{t}}{5\sqrt{n}}\Big\} + \frac{2\lambdat{t}}{\sqrt{n}}\epsilon + \mathcal{W}\Big(n, \frac{\delta}{3RK}, r_{\btheta}^*\Big),
\end{equation}
when $t \geq 2$. By induction,
\begin{align}
	G^{[t]} &\leq \kappa_0^{t-1}G^{[1]} + \frac{1-\kappa_0^{t-1}}{1-\kappa_0}\bar{\eta}\mathcal{E}_2\Big(n, |S|, \frac{\delta}{3R}\Big) + \frac{6}{1-2\epsilon}\underbrace{\sum_{t'=2}^t \kappa_0^{t-t'}\cdot \min\Big\{3h, \frac{2\lambdat{t'}}{5\sqrt{n}}\Big\}}_{[3]} + \frac{2\epsilon}{\sqrt{n}}\underbrace{\sum_{t'=2}^t \kappa_0^{t-t'}\cdot\lambdat{t'}}_{[4]} \\
	&\quad + \mathcal{W}\Big(n, \frac{\delta}{3RK}, r_{\btheta}^*\Big).
\end{align}
By \eqref{eq: lambda t},
\begin{align}
	[3] &\leq \min\bigg\{3 \frac{1-\kappa_0^{t-1}}{1-\kappa_0}h, \frac{2}{5}(t-1)(\wtkappa_0)^t\cdot \frac{\lambdat{0}}{\sqrt{n}} + \frac{6}{1-\wtkappa_0}\cdot \frac{1-\kappa_0^{t-1}}{1-\kappa_0}\Big[\mathcal{W}\Big(n, \frac{\delta}{3RK}, r_{\btheta}^*\Big) + 2\bar{\eta}\mathcal{E}_1\Big(n, \frac{\delta}{3RK}\Big)\Big]\bigg\}, \\
	[4] &\leq (t-1)(\wtkappa_0)^t\lambdat{0} + \frac{15}{1-\wtkappa_0}\cdot \frac{1-\kappa_0^{t-1}}{1-\kappa_0}\Big[\mathcal{W}\Big(n, \frac{\delta}{3RK}, r_{\btheta}^*\Big) + 2\bar{\eta}\mathcal{E}_1\Big(n, \frac{\delta}{3RK}\Big)\Big]\sqrt{n}.
\end{align}
Therefore,
\begin{align}
	G^{[t]} &\leq (\wtkappa_0/119)^{t-1}G^{[1]} + \frac{1}{1-\kappa_0}\Big[\mathcal{W}\Big(n, \frac{\delta}{3RK}, r_{\btheta}^*\Big) + \bar{\eta}\mathcal{E}_2\Big(n, |S|, \frac{\delta}{3R}\Big)\Big] \\
	&\quad + \frac{18}{1-\wtkappa_0/119}\cdot\min\bigg\{3h, \frac{6}{1-\wtkappa_0}\Big[\mathcal{W}\Big(n, \frac{\delta}{3RK}, r_{\btheta}^*\Big) + 2\bar{\eta}\mathcal{E}_1\Big(n, \frac{\delta}{3RK}\Big)\Big]\bigg\} \\
	&\quad + \frac{30}{(1-\wtkappa_0)(1-\wtkappa_0/119)}\epsilon\cdot  \Big[\mathcal{W}\Big(n, \frac{\delta}{3RK}, r_{\btheta}^*\Big) + 2\bar{\eta}\mathcal{E}_1\Big(n, \frac{\delta}{3RK}\Big)\Big] + \Big(\frac{2}{3} + \frac{2}{5}\cdot 18\Big)\cdot (t-1)(\wtkappa_0)^t \frac{\lambdat{0}}{\sqrt{n}} \\
	&\leq \frac{119}{15}\wtkappa_0(\wtkappa_0/119)^{t-1}\bigg\{\frac{\lambdat{0}}{\sqrt{n}} + \Big(\frac{119}{1-\wtkappa_0}+1\Big)\Big[\mathcal{W}\Big(n, \frac{\delta}{3RK}, r_{\btheta}^*\Big) + 2\bar{\eta}\mathcal{E}_1\Big(n, \frac{\delta}{3RK}\Big)\Big]\bigg\} \\
	&\quad + \frac{1}{1-\kappa_0}\Big[\bar{\eta}\mathcal{E}_2\Big(n, |S|, \frac{\delta}{3R}\Big) + \mathcal{W}\Big(n, \frac{\delta}{3RK}, r_{\btheta}^*\Big)\Big] \\
	&\quad + \frac{18}{1-\wtkappa_0/119}\cdot\min\bigg\{3h, \frac{6}{1-\wtkappa_0}\Big[\mathcal{W}\Big(n, \frac{\delta}{3RK}, r_{\btheta}^*\Big) + 2\bar{\eta}\mathcal{E}_1\Big(n, \frac{\delta}{3RK}\Big)\Big]\bigg\} \\
	&\quad + \frac{30}{(1-\wtkappa_0)(1-\wtkappa_0/119)}\epsilon\cdot  \Big[\mathcal{W}\Big(n, \frac{\delta}{3RK}, r_{\btheta}^*\Big) + 2\bar{\eta}\mathcal{E}_1\Big(n, \frac{\delta}{3RK}\Big)\Big] + \frac{118}{15}\cdot (t-1)(\wtkappa_0)^t \frac{\lambdat{0}}{\sqrt{n}} \\
	&\leq \wtkappa_0(\wtkappa_0/119)^{t-1}(r_w^* + r_{\btheta}^*) + \bigg[\frac{119}{15}\wtkappa_0(\wtkappa_0/119)^{t-1} + \frac{118}{119}(t-1)(\wtkappa_0)^t\bigg](r_w^* + r_{\btheta}^*)\\
	&\quad + \frac{1}{1-\kappa_0}\Big[\bar{\eta}\mathcal{E}_2\Big(n, |S|, \frac{\delta}{3R}\Big) + \mathcal{W}\Big(n, \frac{\delta}{3RK}, r_{\btheta}^*\Big)\Big] \\
	&\quad + \frac{18}{1-\wtkappa_0/119}\cdot\min\bigg\{3h, \frac{6}{1-\wtkappa_0}\Big[\mathcal{W}\Big(n, \frac{\delta}{3RK}, r_{\btheta}^*\Big) + 2\bar{\eta}\mathcal{E}_1\Big(n, \frac{\delta}{3RK}\Big)\Big]\bigg\} \\
	&\quad + \frac{30}{(1-\wtkappa_0)(1-\wtkappa_0/119)}\epsilon\cdot  \Big[\mathcal{W}\Big(n, \frac{\delta}{3RK}, r_{\btheta}^*\Big) + 2\bar{\eta}\mathcal{E}_1\Big(n, \frac{\delta}{3RK}\Big)\Big].
\end{align}
Note that $\wtkappa_0(\wtkappa_0/119)^{t-1} + \frac{119}{15}\wtkappa_0(\wtkappa_0/119)^{t-1} + \frac{118}{119}(t-1)(\wtkappa_0)^t \leq 9\wtkappa_0(\wtkappa_0/119)^{t-1} + \frac{118}{119}(t-1)(\wtkappa_0)^t \leq 10t(\wtkappa_0)^t$, hence
\begin{align}
	G^{[t]} &\leq 20t(\wtkappa_0)^{t-1}(r_w^* \vee r_{\btheta}^*) + \bigg[\frac{119}{15}\wtkappa_0(\wtkappa_0/119)^{t-1} + \frac{118}{119}(t-1)(\wtkappa_0)^t\bigg](r_w^* + r_{\btheta}^*)\\
	&\quad + \frac{1}{1-\kappa_0}\Big[\bar{\eta}\mathcal{E}_2\Big(n, |S|, \frac{\delta}{3R}\Big) + \mathcal{W}\Big(n, \frac{\delta}{3RK}, r_{\btheta}^*\Big)\Big] \\
	&\quad + \frac{18}{1-\wtkappa_0/119}\cdot\min\bigg\{3h, \frac{6}{1-\wtkappa_0}\Big[\mathcal{W}\Big(n, \frac{\delta}{3RK}, r_{\btheta}^*\Big) + 2\bar{\eta}\mathcal{E}_1\Big(n, \frac{\delta}{3RK}\Big)\Big]\bigg\} \\
	&\quad + \frac{30}{(1-\wtkappa_0)(1-\wtkappa_0/119)}\epsilon\cdot  \Big[\mathcal{W}\Big(n, \frac{\delta}{3RK}, r_{\btheta}^*\Big) + 2\bar{\eta}\mathcal{E}_1\Big(n, \frac{\delta}{3RK}\Big)\Big]. \label{eq: Gt rate}
\end{align}
By Assumption 4, we obtain $\max_{k \in S}\max_{r \in [R]}\twonorm{\hthetakt{k}{t}_r - \bthetaks{k}_r} \leq G^{[t]} \leq r_{\btheta}^*$.

Next, let us shrink the contraction radius to obtain the desired rate. Recall that 
\begin{align}
	A_t &= \bigg[9\tilde{\kappa}_0\Big(\frac{\tilde{\kappa}_0}{119}\Big)^{t-1} + \frac{118}{119}(t-1)\tilde{\kappa}_0^{t-1}\bigg](r_w^* +r_{\btheta}^*) + \frac{1}{1-\tilde{\kappa}_0/119}\bar{\eta}\mathcal{E}_2\Big(n, |S|, \frac{\delta}{3R}\Big) \\
	&\quad+ \frac{18}{1-\tilde{\kappa}_0/119} \min\bigg\{3h, \frac{6}{1-\tilde{\kappa}_0}\Big[\mathcal{W}\Big(n, \frac{\delta}{3RK}, r_{\btheta}^*\Big) + 2\bar{\eta}\mathcal{E}_1\Big(n, \frac{\delta}{3R}\Big)\Big]\bigg\}\\
	&\quad + \frac{30}{(1-\tilde{\kappa}_0)(1-\tilde{\kappa}_0/119)}\epsilon\bigg[\mathcal{W}\Big(n, \frac{\delta}{3RK}, r_{\btheta}^*\Big) + 2\bar{\eta}\mathcal{E}_1\Big(n, \frac{\delta}{3R}\Big)\bigg],
\end{align}
and
\begin{equation}
	A_{t} + \frac{18}{1-\tilde{\kappa}_0/119}\mathcal{W}\Big(n, \frac{\delta}{3RK}, r_{\btheta, t}^*\Big) = r_{\btheta, t+1}^*,
\end{equation}
with $r_{\btheta, 1}^* \coloneqq r_{\btheta}^*$. Then we repeat the previous analysis in part (\Rom{3}) for $t = 1:T$, then we will get the same rate as in \eqref{eq: Gt rate} but replace $r_{\btheta}^*$ with $r_{\btheta, T}^*$ in the term $\frac{1}{1-\kappa_0}\big[\bar{\eta}\mathcal{E}_2(n, |S|, \frac{\delta}{3R}) + \mathcal{W}(n, \frac{\delta}{3RK}, r_{\btheta}^*)\big]$.

(\Rom{4}) Part 4: Combining this rate (which holds when $h \leq \frac{1}{3}[\mathcal{W}(n, \frac{\delta}{3RK}, r_{\btheta}^*) + 2\bar{\eta}\mathcal{E}_1(n, \frac{\delta}{3RK})]$) with \eqref{eq: Gt large h} (which holds for any $h \geq 0$ but is only used when $h > \frac{1}{3}[\mathcal{W}(n, \frac{\delta}{3RK}, r_{\btheta}^*) + 2\bar{\eta}\mathcal{E}_1(n, \frac{\delta}{3RK})]$) completes our proof.

\subsection{Proof of Proposition \ref{prop: gmm appendix}}
We first introduce two useful lemmas.
\begin{lemma}[Theorem 3 in \citet{maurer2021concentration}]\label{lem: maurer concentration}
	Let $f: \mathcal{X}^n \rightarrow \mathbb{R}$ and $X = (X_1, \ldots, X_n)$ be a vector of independent random variables with values in a space $\mathcal{X}$. Then for any $t > 0$ we have
	\begin{equation}
		\tP(f(X) - \tE f(X) > t) \leq \exp\left\{-\frac{t^2}{32e\infnorma{\sum_{i=1}^n \|f_i(X)\|_{\psi_2}^2}}\right\},
	\end{equation}
	where $f_i(X)$ as a random function of $x$ is defined to be $(f_i(X))(x) \coloneqq f(x_1, \ldots, x_{i-1}, X_{i}, x_{i+1}, \ldots, X_n) - \tE_{X_i}[f(x_1, \ldots, x_{i-1}, X_{i}, x_{i+1}, \ldots, X_n)]$, the sub-Gaussian norm $\|Z\|_{\psi_2} \coloneqq \sup_{d\geq 1}\{\|Z\|_d/\sqrt{d}\}$, and $\|Z\|_d = (\tE |Z|^d)^{1/d}$.
\end{lemma}

\begin{lemma}[Vectorized contraction of Rademacher complexity, Corollary 1 in \citet{maurer2016vector}]\label{lem: vec contraction}
	Suppose $\{\epsilon_{ir}\}_{i\in [n], r \in [R]}$ and $\{\epsilon_i\}_{i=1}^n$ are independent Rademacher variables. Let $\mathcal{F}$ be a class of functions $f: \mathbb{R}^d \rightarrow  \mathcal{S}\subseteq \mathbb{R}^R$ and $h: \mathcal{S} \rightarrow\mathbb{R}$ is $L$-Lipschitz under $\ell_2$-norm, i.e., $\norm{h(\bm{y}) - h(\bm{y}')} \leq L\twonorm{\bm{y} - \bm{y}'}$, where $\bm{y} = (y_1,\ldots,y_R)^T$, $\bm{y}' = (y_1',\ldots,y_R')^T \in \mathcal{S}$. Then
	\begin{equation}
		\tE \sup_{f \in \mathcal{F}} \sum_{i=1}^n \epsilon_i h(f(x_i)) \leq \sqrt{2}L\tE \sup_{f \in \mathcal{F}} \sum_{i=1}^n\sum_{r=1}^R \epsilon_{ir}f_r(x_i), 
	\end{equation}
	where $f_r(x_i)$ is the $r$-th component of $f(x_i) \in \mathcal{S}\subseteq \mathbb{R}^R$.
\end{lemma}

Define the posterior
\begin{align}
	&\gamma^{(r)}_{\bthetak{k}, \bwk{k}}(\bxk{k}) = \frac{\wk{k}_r\exp\{(\bxk{k})^T(\bthetak{k}_r - \bthetak{k}_1) - \frac{1}{2}(\twonorm{\bthetak{k}_r}^2 - \twonorm{\bthetak{k}_1}^2)\}}{\wk{k}_1 + \sum_{r=2}^R\wk{k}_r\exp\{(\bxk{k})^T(\bthetak{k}_r - \bthetak{k}_1) - \frac{1}{2}(\twonorm{\bthetak{k}_r}^2 - \twonorm{\bthetak{k}_1}^2)\}} \\
	&\quad = \tP(\zk{k} = r|\bxk{k};\bthetak{k}, \bwk{k}), \, r \in [R],
\end{align}
where $\bwk{k} = \{\wk{k}\}_{k=1}^K$ and $\bthetak{k} = \{\bthetak{k}_r\}_{k=1}^K$.

By definition, $\qk{k}(\btheta) = \Qk{k}(\btheta|\bthetaks{k}, \bwks{k}) = -\frac{1}{2}\tE\big[\sum_{r=1}^R \gamma^{(r)}_{\bthetaks{k}, \bwks{k}}(\bxk{k})\twonorm{\bxk{k}-\btheta}^2\big]$, hence $\muk{k}_r = \Lk{k}_r = \wks{k}_r$ with $r_1^* = +\infty$. And
\begin{align}
	\hQk{k}(\btheta|\btheta', \bw') &= -\frac{1}{2n_k}\sum_{i=1}^{n}\sum_{r=1}^R\gamma^{(r)}_{\btheta', \bw'}(\bxk{k}_i)\twonorm{\bxk{k}_i-\btheta}^2, \\
	\frac{\partial}{\partial \btheta_r}\Qk{k}(\btheta|\btheta', \bw') &= -\tE_{\bxk{k}}\big[\gamma^{(r)}_{\btheta', \bw'}(\bxk{k})(\btheta-\bxk{k})\big], \\
	\frac{\partial}{\partial \btheta_r}\hQk{k}(\btheta|\btheta', \bw') &= -\frac{1}{n_k}\sum_{i=1}^{n}\gamma^{(r)}_{\btheta', \bw'}(\bxk{k}_i)(\btheta-\bxk{k}_i).
\end{align}
From the proof of Theorem 1 in \citet{tian2022unsupervised}, we have $\kappa \asymp c_w^{-2}R^2\exp\{-C\Delta^2\}$, $\gamma \asymp M^2c_w^{-2}R^2\exp\{-C\Delta^2\}$ with $r_2 = C_b\Delta$. Consider $r_{\btheta}^* = r_1^* \wedge r_2^* = C_b\Delta \leq M$. In the remaining proof of Proposition 1, we will derive the expressions of $\mathcal{W}, \mathcal{E}_1$ and $\mathcal{E}_2$. Let
\begin{align}
	V &= \sup_{\substack{|w_r - \wks{k}_r| \leq \frac{c_w}{2R} \\ \twonorm{\btheta_r - \bthetaks{k}_r} \leq \xi}}\bigg|\frac{1}{n}\sum_{i=1}^n\tP(\zk{k}=r|\bxk{k}_i;\bw, \btheta) - \tE_{\bxk{k}}\big[\tP(\zk{k}=r|\bxk{k};\bw, \btheta)\big]\bigg| \\
	&= \sup_{\substack{|w_r - \wks{k}_r| \leq \frac{c_w}{2R} \\ \twonorm{\btheta_r - \bthetaks{k}_r} \leq \xi}}\bigg|\frac{1}{n}\sum_{i=1}^n\gamma^{(r)}_{\btheta, \bw}(\bxk{k}_i) - \tE\big[\gamma^{(r)}_{\btheta, \bw}(\bxk{k})\big]\bigg|.
\end{align}
By bounded difference inequality (Corollary 2.21 in \citet{wainwright2019high}), w.p. at least $1-\delta$,
\begin{equation}
	V \leq \tE V + \sqrt{\frac{\log(1/\delta)}{n}}.
\end{equation}
And by classical symmetrization arguments (e.g., see Proposition 4.11 in \citet{wainwright2019high}),
\begin{equation}
	\tE V \leq \frac{2}{n}\tE_{\bxk{k}}\tE_{\bm{\epsilon}}\sup_{\substack{|w_r - \wks{k}_r| \leq \frac{c_w}{2R} \\ \twonorm{\btheta_r - \bthetaks{k}_r} \leq \xi}} \bigg|\sum_{i=1}^n \epsilonk{k}_i\gamma^{(k)}_{\btheta, \bw}(\bxk{k}_i)\bigg|.
\end{equation}
Let $g_{ir}^{(k)} = (\btheta_r - \btheta_1)^T\bxk{k}_i - \frac{1}{2}(\twonorm{\btheta_r}^2 - \twonorm{\btheta_1}^2) + \log w_r - \log w_1$, $\varphi(\bx) = \frac{\exp\{x_r\}}{1+\sum_{r=2}^R \exp\{x_r\}}$, where $\varphi$ is $1$-Lipschitz (w.r.t. $\ell_2$-norm) and $\gamma^{(r)}_{\btheta, \bw}(\bx) = \varphi(\{g^{(k)}_{ir}\}_{r=2}^R)$. Then by Lemma \ref{lem: vec contraction},
\begin{align}
	&\frac{2}{n}\tE_{\bxk{k}}\tE_{\bm{\epsilon}}\sup_{\substack{|w_r - \wks{k}_r| \leq \frac{c_w}{2R} \\ \twonorm{\btheta_r - \bthetaks{k}_r} \leq \xi}} \bigg|\sum_{i=1}^n \epsilonk{k}_i\gamma^{(k)}_{\btheta, \bw}(\bxk{k}_i)\bigg| \\
	&\lesssim \frac{1}{n}\tE_{\bxk{k}}\tE_{\bm{\epsilon}}\sup_{\substack{|w_r - \wks{k}_r| \leq \frac{c_w}{2R} \\ \twonorm{\btheta_r - \bthetaks{k}_r} \leq \xi}} \bigg|\sum_{i=1}^n \sum_{r=2}^R \epsilonk{k}_{ir}g^{(k)}_{ir}\bigg| \\
	&\lesssim \frac{1}{n}\sum_{r=2}^R \tE_{\bxk{k}}\tE_{\bm{\epsilon}}\sup_{\substack{|w_r - \wks{k}_r| \leq \frac{c_w}{2R} \\ \twonorm{\btheta_r - \bthetaks{k}_r} \leq \xi}} \bigg|\sum_{i=1}^n \epsilonk{k}_{ir}g^{(k)}_{ir}\bigg| \\
	&\lesssim \sum_{r=2}^R\Bigg\{\frac{1}{n}\tE_{\bxk{k}}\tE_{\bm{\epsilon}} \sup_{\substack{|w_r - \wks{k}_r| \leq \frac{c_w}{2R} \\ \twonorm{\btheta_r - \bthetaks{k}_r} \leq \xi}} \bigg|\sum_{i=1}^n \epsilonk{k}_{ir}(\btheta_r - \btheta_1)^T\bxk{k}_i\bigg| + \frac{1}{n}\tE_{\bxk{k}}\tE_{\bm{\epsilon}} \sup_{\substack{|w_r - \wks{k}_r| \leq \frac{c_w}{2R} \\ \twonorm{\btheta_r - \bthetaks{k}_r} \leq \xi}} \bigg|\sum_{i=1}^n \epsilonk{k}_{ir}(\twonorm{\btheta_r}^2 - \twonorm{\btheta_1}^2)\bigg| \\
	&\quad\quad\quad +  \frac{1}{n}\tE_{\bxk{k}}\tE_{\bm{\epsilon}} \sup_{\substack{|w_r - \wks{k}_r| \leq \frac{c_w}{2R} \\ \twonorm{\btheta_r - \bthetaks{k}_r} \leq \xi}} \bigg|\sum_{i=1}^n \epsilonk{k}_{ir}(\log w_r - \log w_1)\bigg|\Bigg\}\\
	&\lesssim \sum_{r=2}^R\Bigg\{\frac{1}{n}\tE_{\bxk{k}}\tE_{\bm{\epsilon}} \sup_{\substack{|w_r - \wks{k}_r| \leq \frac{c_w}{2R} \\ \twonorm{\btheta_r - \bthetaks{k}_r} \leq \xi}} \bigg|\sum_{i=1}^n \epsilonk{k}_{ir}(\btheta_r - \bthetaks{k}_r)^T\bxk{k}_i\bigg| + \frac{1}{n}\tE_{\bxk{k}}\tE_{\bm{\epsilon}} \sup_{\substack{|w_r - \wks{k}_r| \leq \frac{c_w}{2R} \\ \twonorm{\btheta_r - \bthetaks{k}_r} \leq \xi}} \bigg|\sum_{i=1}^n \epsilonk{k}_{ir}(\btheta_1 - \bthetaks{k}_1)^T\bxk{k}_i\bigg| \\
	&\quad\quad\quad + \frac{1}{n}\tE_{\bxk{k}}\tE_{\bm{\epsilon}} \bigg|\sum_{i=1}^n \epsilonk{k}_{ir}(\bthetaks{k}_r - \bthetaks{k}_1)^T\bxk{k}_i\bigg|  + \frac{1}{n}\tE_{\bxk{k}}\tE_{\bm{\epsilon}} \sup_{\substack{|w_r - \wks{k}_r| \leq \frac{c_w}{2R} \\ \twonorm{\btheta_r - \bthetaks{k}_r} \leq \xi}} \bigg|\sum_{i=1}^n \epsilonk{k}_{ir}(\btheta_r + \btheta_1)^T(\btheta_r - \btheta_1)\bigg|\\
	&\quad\quad\quad  +  \frac{1}{n}\tE_{\bxk{k}}\tE_{\bm{\epsilon}} \sup_{\substack{|w_r - \wks{k}_r| \leq \frac{c_w}{2R} \\ \twonorm{\btheta_r - \bthetaks{k}_r} \leq \xi}} \bigg|\sum_{i=1}^n \epsilonk{k}_{ir}(\log w_r - \log w_1)\bigg|\Bigg\}\\
	&\lesssim RM\xi\sqrt{\frac{d}{n}} + [RM^2+R\log(Rc_w^{-1})]\sqrt{\frac{1}{n}}, 
\end{align}
which implies 
\begin{equation}
	V \lesssim RM\xi\sqrt{\frac{d}{n}} + [RM^2+R\log(Rc_w^{-1})]\sqrt{\frac{1}{n}} + \sqrt{\frac{\log(1/\delta)}{n}} \asymp \mathcal{W}(n, \delta, \xi).
\end{equation}
w.p. at least $1-\delta$.

Next, let 
\begin{align}
	U &= \sup_{\substack{|w_r - \wks{k}_r| \leq \frac{c_w}{2R} \\ \twonorm{\btheta_r - \bthetaks{k}_r} \leq r_{\btheta}^*}}\bigg\|\frac{1}{n}\sum_{i=1}^n\gamma^{(r)}_{\btheta, \bw}(\bxk{k}_i)\bxk{k}_i - \tE\big[\gamma^{(r)}_{\btheta, \bw}(\bxk{k})\bxk{k}\big]\bigg\|_2 \\
	&= \sup_{\twonorm{\bm{u}}\leq 1} \sup_{\substack{|w_r - \wks{k}_r| \leq \frac{c_w}{2R} \\ \twonorm{\btheta_r - \bthetaks{k}_r} \leq r_{\btheta}^*}}\bigg|\frac{1}{n}\sum_{i=1}^n\gamma^{(r)}_{\btheta, \bw}(\bxk{k}_i)(\bxk{k}_i)^T\bm{u} - \tE\big[\gamma^{(r)}_{\btheta, \bw}(\bxk{k})(\bxk{k})^T\bm{u}\big]\bigg| \\
	&\leq 2\max_{j=1:N}\underbrace{\sup_{\substack{|w_r - \wks{k}_r| \leq \frac{c_w}{2R} \\ \twonorm{\btheta_r - \bthetaks{k}_r} \leq r_{\btheta}^*}}\bigg|\frac{1}{n}\sum_{i=1}^n\gamma^{(r)}_{\btheta, \bw}(\bxk{k}_i)(\bxk{k}_i)^T\bm{u}_j - \tE\big[\gamma^{(r)}_{\btheta, \bw}(\bxk{k})(\bxk{k})^T\bm{u}_j\big]\bigg|}_{U_j},
\end{align}
where $\{\bm{u}_j\}_{j=1}^N$ is a $1/2$-cover of the unit ball $\mathcal{B}(\bm{0}, 1)$ in $\mathbb{R}^d$ w.r.t. $\ell_2$-norm, with $N \leq 5^d$ (by Example 5.8 in \cite{wainwright2019high}). We first bound $U_j - \tE U_j$ as follows. Fix $\bxk{k}_1, \ldots, \bxk{k}_{i-1}, \bxk{k}_{i+1}, \ldots, \bxk{k}_n$ and define $s_{ir}^{(k)}(\bxk{k}_i) = V_j - \tE[V_j|\bxk{k}_1, \ldots, \bxk{k}_{i-1}, \bxk{k}_{i+1}, \ldots, \bxk{k}_n]$. Then
\begin{equation}
	|s_{ir}^{(k)}(\bxk{k}_i)| \leq \underbrace{\frac{1}{n}\sup_{\substack{|w_r - \wks{k}_r| \leq \frac{c_w}{2R} \\ \twonorm{\btheta_r - \bthetaks{k}_r} \leq r_{\btheta}^*}} \Big|\gamma^{(r)}_{\btheta, \bw}(\bxk{k}_i)(\bxk{k}_i)^T\bm{u}_j\Big|}_{W_1} + \underbrace{\frac{2}{n}\tE\Bigg|\sup_{\substack{|w_r - \wks{k}_r| \leq \frac{c_w}{2R} \\ \twonorm{\btheta_r - \bthetaks{k}_r} \leq r_{\btheta}^*}} \gamma^{(r)}_{\btheta, \bw}(\bxk{k}_i)(\bxk{k}_i)^T\bm{u}_j\Bigg|}_{W_2},
\end{equation}
where $[\tE(W_1 + W_2)^d]^{1/d} \leq (\tE W_1^d)^{1/d} + (\tE W_2^d)^{1/d}$, and $(\tE W_1^d)^{1/d}, (\tE W_2^d)^{1/d} \leq CM\sqrt{d}/n$ with some constant$C > 0$. Then by Lemma \ref{lem: maurer concentration},
\begin{equation}
	\tP(U_j - \tE U_j \geq t) \lesssim \exp\bigg\{-\frac{Cnt^2}{M^2}\bigg\}.
\end{equation}
By a similar procedure used in deriving $\mathcal{W}(n, \delta, \xi)$, we can show that
\begin{equation}
	\tE U_j \lesssim RM^2r_{\btheta}^*\sqrt{\frac{d}{n}} + [RM^3+RM\log(Rc_w^{-1})]\sqrt{\frac{1}{n}}.
\end{equation}
As a consequence,
\begin{equation}
	\tP\bigg(U_j \geq CRM^2r_{\btheta}^*\sqrt{\frac{d}{n}} + C[RM^3+RM\log(Rc_w^{-1})]\sqrt{\frac{1}{n}} + t\bigg) \lesssim \exp\bigg\{-\frac{Cnt^2}{M^2}\bigg\}.
\end{equation}
Therefore
\begin{equation}
	\tP\bigg(\max_{j=1:N}U_j \geq CRM^2r_{\btheta}^*\sqrt{\frac{d}{n}} + C[RM^3+RM\log(Rc_w^{-1})]\sqrt{\frac{1}{n}} + t\bigg) \lesssim N\exp\bigg\{-\frac{Cnt^2}{M^2}\bigg\},
\end{equation}
which implies that
\begin{equation}
	U \lesssim (RM^2r_{\btheta}^*+M)\sqrt{\frac{d}{n}} + [RM^3+RM\log(Rc_w^{-1})]\sqrt{\frac{1}{n}} + M\sqrt{\frac{\log(1/\delta)}{n}},
\end{equation}
w.p. at least $1-\delta$. On the other hand, by $\mathcal{W}(n, \delta, r_{\btheta}^*)$, we have
\begin{equation}
	\sup_{\substack{|w_r - \wks{k}_r| \leq \frac{c_w}{2R} \\ \twonorm{\btheta_r - \bthetaks{k}_r} \leq r_{\btheta}^*}}\bigg\|\frac{1}{n}\sum_{i=1}^n\gamma^{(r)}_{\btheta, \bw}(\bxk{k}_i)\btheta - \tE\big[\gamma^{(r)}_{\btheta, \bw}(\bxk{k})\btheta\big]\bigg\|_2 \lesssim \bigg[RM\xi\sqrt{\frac{d}{n}} + [RM^2+R\log(Rc_w^{-1})]\sqrt{\frac{1}{n}} + \sqrt{\frac{\log(1/\delta)}{n}}\bigg]\cdot M,
\end{equation}
hence
\begin{align}
	\mathcal{E}_1(n, \delta) &\asymp (RM^2r_{\btheta}^*+M)\sqrt{\frac{d}{n}} + [RM^3+RM\log(Rc_w^{-1})]\sqrt{\frac{1}{n}} + M\sqrt{\frac{\log(1/\delta)}{n}} \\
	&\asymp RM^3\sqrt{\frac{d}{n}} + [RM^3+RM\log(Rc_w^{-1})]\sqrt{\frac{1}{n}} + M\sqrt{\frac{\log(1/\delta)}{n}},
\end{align}
where the last inequality is due to $r_{\btheta}^* = r_1^* \wedge r_2^* = C_b\Delta \leq M$.

Let 
\begin{align}
	Z &= \sup_{\substack{\norm{w_r - \wks{k}_r} \leq \frac{c_w}{2R} \\ \twonorm{\btheta_r - \bthetaks{k}_r} \leq r_{\btheta}^* \\ 0 < \leq \etak{k}_r \leq \bar{\eta}}} \frac{1}{n|S|}\Bigg\|\sum_{k \in S}\etak{k}_r\cdot \sum_{i=1}^n\big(\gamma^{(k)}_{\btheta, \bw}(\bxk{k}_i)\bxk{k}_i - \tE[\gamma^{(k)}_{\btheta, \bw}(\bxk{k})\bxk{k}]\big)\Bigg\|_2 \\
	&= \sup_{\twonorm{\bm{u}}\leq 1}\sup_{\substack{\norm{w_r - \wks{k}_r} \leq \frac{c_w}{2R} \\ \twonorm{\btheta_r - \bthetaks{k}_r} \leq r_{\btheta}^* \\ 0 < \leq \etak{k}_r \leq \bar{\eta}}} \frac{1}{n|S|}\Bigg|\sum_{k \in S}\etak{k}_r\cdot \sum_{i=1}^n\big(\gamma^{(k)}_{\btheta, \bw}(\bxk{k}_i)(\bxk{k}_i)^T\bm{u} - \tE[\gamma^{(k)}_{\btheta, \bw}(\bxk{k})(\bxk{k})^T\bm{u}]\big)\Bigg| \\
	&\leq \sup_{j'_1,\ldots, j_k'=1:N'}\sup_{j=1:N}\underbrace{\frac{2}{n|S|}\sup_{\substack{\norm{w_r - \wks{k}_r} \leq \frac{c_w}{2R} \\ \twonorm{\btheta_r - \bthetaks{k}_r} \leq r_{\btheta}^*}}\Bigg|\sum_{k \in S}\eta_{j'_k}\cdot \sum_{i=1}^n\big(\gamma^{(k)}_{\btheta, \bw}(\bxk{k}_i)(\bxk{k}_i)^T\bm{u}_j - \tE[\gamma^{(k)}_{\btheta, \bw}(\bxk{k})(\bxk{k})^T\bm{u}_j]\big)\Bigg|}_{Z(j,j'_1,\ldots, j_k')},
\end{align}
where $\{\bm{u}_j\}_{j=1}^N$ is a $1/2$-cover of the unit ball $\mathcal{B}(\bm{0}, 1)$ in $\mathbb{R}^d$ w.r.t. $\ell_2$-norm with $N \leq 5^d$ and $\{\eta_{j'}\}_{j'=1}^{N'}$ is a $1/2$-cover of $[0,1]$ with $N' \leq 2$. We first bound $Z(j,j'_1,\ldots, j_k') - \tE Z(j,j'_1,\ldots, j_k')$ as follows. Fix $\bxk{k}_1, \ldots, \bxk{k}_{i-1}, \bxk{k}_{i+1}, \ldots, \bxk{k}_n$ and define $v_{ir}^{(k)}(\bxk{k}_i) = Z(j,j'_1,\ldots, j_k') - \tE[Z(j,j'_1,\ldots, j_k')|\{\bxk{k}_i\}_{k \in S, i \in [n]}\backslash \{\bxk{k}_i\}]$. Then
\begin{equation}
	|v_{ir}^{(k)}(\bxk{k}_i)| \leq \underbrace{\frac{\eta_{j_k}}{n|S|}\sup_{\substack{|w_r - \wks{k}_r| \leq \frac{c_w}{2R} \\ \twonorm{\btheta_r - \bthetaks{k}_r} \leq r_{\btheta}^*}} \Big|\gamma^{(r)}_{\btheta, \bw}(\bxk{k}_i)(\bxk{k}_i)^T\bm{u}_j\Big|}_{W_1} + \underbrace{\frac{2\eta_{j_k}}{n|S|}\tE\Bigg|\sup_{\substack{|w_r - \wks{k}_r| \leq \frac{c_w}{2R} \\ \twonorm{\btheta_r - \bthetaks{k}_r} \leq r_{\btheta}^*}} \gamma^{(r)}_{\btheta, \bw}(\bxk{k}_i)(\bxk{k}_i)^T\bm{u}_j\Bigg|}_{W_2}.
\end{equation}
Via the same procedure used to bound $U_j$, it can be shown that
\begin{align}
	\tP(Z(j,j'_1,\ldots, j_k') - \tE Z(j,j'_1,\ldots, j_k') \geq t) \lesssim \exp\bigg\{-\frac{Cn|S|t^2}{M^2\bar{\eta}^2}\bigg\}, \\
	\tE Z(j,j'_1,\ldots, j_k') \lesssim \bar{\eta}RM^2r_{\btheta}^*\sqrt{\frac{d}{n|S|}} + \bar{\eta}[RM^3+RM\log(Rc_w^{-1})]\sqrt{\frac{1}{n}},
\end{align}
leading to
\begin{equation}
	\tP\bigg(Z(j,j'_1,\ldots, j_k') \geq \bar{\eta}RM^2r_{\btheta}^*\sqrt{\frac{d}{n|S|}} + \bar{\eta}[RM^3+RM\log(Rc_w^{-1})]\sqrt{\frac{1}{n}} + t\bigg) \lesssim \exp\bigg\{-\frac{Cn|S|t^2}{M^2\bar{\eta}^2}\bigg\}.
\end{equation}
Therefore
\begin{align}
	&\tP\bigg(\max_{j'_1,\ldots, j_k'=1:N'}\max_{j=1:N}Z(j,j'_1,\ldots, j_k') \geq C\bar{\eta}RM^2r_{\btheta}^*\sqrt{\frac{d}{n|S|}} + C\bar{\eta}[RM^3+RM\log(Rc_w^{-1})]\sqrt{\frac{1}{n}} + t\bigg) \\
	&\lesssim N(N')^K\exp\bigg\{-\frac{Cn|S|t^2}{M^2\bar{\eta}^2}\bigg\},
\end{align}
which implies that
\begin{equation}
	Z\leq \max_{j'_1,\ldots, j_k'=1:N'}\max_{j=1:N}Z(j,j'_1,\ldots, j_k') \lesssim \bar{\eta}(RM^2r_{\btheta}^*+M)\sqrt{\frac{d}{n|S|}} + \bar{\eta}[RM^3+RM\log(Rc_w^{-1})]\sqrt{\frac{1}{n}} + \bar{\eta}M\sqrt{\frac{\log(1/\delta)}{n|S|}},
\end{equation}
w.p. at least $1-\delta$. Similarly,
\begin{align}
	&\sup_{\substack{\norm{w_r - \wks{k}_r} \leq \frac{c_w}{2R} \\ \twonorm{\btheta_r - \bthetaks{k}_r} \leq r_{\btheta}^* \\ 0 < \leq \etak{k}_r \leq \bar{\eta}}} \frac{1}{n|S|}\Bigg\|\sum_{k \in S}\etak{k}_r\cdot \sum_{i=1}^n\big(\gamma^{(k)}_{\btheta, \bw}(\bxk{k}_i)\btheta_r - \tE[\gamma^{(k)}_{\btheta, \bw}(\bxk{k})\btheta_r]\big)\Bigg\|_2 \\
	&\lesssim \bar{\eta}(RM^2r_{\btheta}^*+M)\sqrt{\frac{d}{n|S|}} + \bar{\eta}[RM^3+RM\log(Rc_w^{-1})]\sqrt{\frac{1}{n}} + \bar{\eta}M\sqrt{\frac{\log(1/\delta)}{n|S|}},
\end{align}
w.p. at least $1-\delta$. Considering that $r_{\btheta}^* = r_1^* \wedge r_2^* = C_b\Delta \leq M$, we have
\begin{equation}
	\mathcal{E}_2(n, |S|, \delta) \asymp  RM^3\sqrt{\frac{d}{n|S|}} + [RM^3+RM\log(Rc_w^{-1})]\sqrt{\frac{1}{n}} + M\sqrt{\frac{\log(1/\delta)}{n|S|}}.
\end{equation}

\subsection{Proof of Proposition \ref{prop: gmm rj appendix}}
Recall that 
\begin{align}
	A_t &= \bigg[9\tilde{\kappa}_0\Big(\frac{\tilde{\kappa}_0}{119}\Big)^{t-1} + \frac{118}{119}(t-1)\tilde{\kappa}_0^{t-1}\bigg](r_w^* +r_{\btheta}^*) + \frac{1}{1-\tilde{\kappa}_0/119}\bar{\eta}\mathcal{E}_2\Big(n, |S|, \frac{\delta}{3R}\Big) \\
	&\quad+ \frac{18}{1-\tilde{\kappa}_0/119} \min\bigg\{3h, \frac{6}{1-\tilde{\kappa}_0}\Big[\mathcal{W}\Big(n, \frac{\delta}{3RK}, r_{\btheta}^*\Big) + 2\bar{\eta}\mathcal{E}_1\Big(n, \frac{\delta}{3R}\Big)\Big]\bigg\}\\
	&\quad + \frac{30}{(1-\tilde{\kappa}_0)(1-\tilde{\kappa}_0/119)}\epsilon\bigg[\mathcal{W}\Big(n, \frac{\delta}{3RK}, r_{\btheta}^*\Big) + 2\bar{\eta}\mathcal{E}_1\Big(n, \frac{\delta}{3R}\Big)\bigg],
\end{align}
and
\begin{equation}
	A_{t} + \frac{18}{1-\tilde{\kappa}_0/119}\mathcal{W}\Big(n, \frac{\delta}{3RK}, r_{\btheta, t}^*\Big) = r_{\btheta, t+1}^*,
\end{equation}
for $t \geq 1$ with $r_{\btheta, 1}^* \coloneqq r_{\btheta}^*$.

By Assumption \ref{asmp: gmm appendix}.(\rom{4}), there exists $\wtkappa_0' \in (0,1)$ such that $CRM\sqrt{\frac{p}{n}} \leq \wtkappa_0'$ with a large $C$. Hence by plugging in the explicit rates obtained in Proposition \ref{prop: gmm appendix},
\begin{align}
	r_{\btheta, t+1}^* &\leq \wtkappa_0'r_{\btheta, t}^* + Ct(\wtkappa_0)^{t-1}(r_w^* \vee r_{\btheta}^*) + C\bar{\eta}RM^3\sqrt{\frac{d}{n|S|}} + C[(\bar{\eta}M)\vee 1][RM^2 + R\log(Rc_w^{-1})]\sqrt{\frac{1}{n}} \\
	&\quad + C[(\bar{\eta}M)\vee 1]\sqrt{\frac{\log(RK/\delta)}{n}} + C\min\bigg\{h, \bar{\eta}RM^3\sqrt{\frac{d}{n}}\bigg\} + \epsilon\bar{\eta}RM^2[(\bar{\eta}M)\vee 1]\sqrt{\frac{d}{n}},
\end{align}
implying that
\begin{align}
	r_{\btheta, T}^* &\lesssim (\wtkappa_0')^{T-1}r_{\btheta}^* + T^2(\wtkappa_0')^{T-1}(r_w^* \vee r_{\btheta}^*) + \bar{\eta}RM^3\sqrt{\frac{d}{n|S|}} + [(\bar{\eta}M)\vee 1][RM^2 + R\log(Rc_w^{-1})]\sqrt{\frac{1}{n}} \\
	&\quad + [(\bar{\eta}M)\vee 1]\sqrt{\frac{\log(RK/\delta)}{n}} + \min\bigg\{h, \bar{\eta}RM^3\sqrt{\frac{d}{n}}\bigg\} + \epsilon\bar{\eta}RM^2[(\bar{\eta}M)\vee 1]\sqrt{\frac{d}{n}} \\
	&\lesssim T^2(\wtkappa_0 \vee \wtkappa_0')^{T-1}(r_w^* \vee r_{\btheta}^*) + \bar{\eta}RM^3\sqrt{\frac{d}{n|S|}} + [(\bar{\eta}M)\vee 1][RM^2 + R\log(Rc_w^{-1})]\sqrt{\frac{1}{n}} \\
	&\quad + [(\bar{\eta}M)\vee 1]\sqrt{\frac{\log(RK/\delta)}{n}} + \min\bigg\{h, \bar{\eta}RM^3\sqrt{\frac{d}{n}}\bigg\} + \epsilon\bar{\eta}RM^2[(\bar{\eta}M)\vee 1]\sqrt{\frac{d}{n}} \\
	&\lesssim T^2(\wtkappa_0 \vee \wtkappa_0')^{T-1}(r_w^* \vee r_{\btheta}^*) + R^2M^3c_w^{-1}\sqrt{\frac{d}{n|S|}} + R^2Mc_w^{-1}[M^2 + \log(Rc_w^{-1})]\sqrt{\frac{1}{n}} \\
	&\quad + MRc_w^{-1}\sqrt{\frac{\log(RK/\delta)}{n}} + \min\bigg\{h, R^2M^3c_w^{-1}\sqrt{\frac{d}{n}}\bigg\} + \epsilon RM^3c_w^{-1}\sqrt{\frac{d}{n}},
\end{align}
where $\kappa_0 = 119\sqrt{\frac{2C_b}{1+C_b}} + CM^2c_w^{-2}R^3\exp\{-C'\Delta^2\} + Cc_w^{-2}R^3\exp\{-C'\Delta^2\} +  \tilde{\kappa}_0'$, and $\tilde{\kappa}_0'$ satisfies $1> \tilde{\kappa}_0' > CMR\sqrt{\frac{d}{n}}$ for some $C > 0$.

\subsection{Proof of Corollary \ref{cor: gmm appendix}}
By the rate of $\mathcal{W}(n, \frac{\delta}{3RK}, r_{\btheta, T}^*)$ in Proposition \ref{prop: gmm appendix} and the upper bound of $r_{\btheta, T}^*$ in Proposition \ref{prop: gmm rj appendix},
\begin{align}
	\mathcal{W}\Big(n, \frac{\delta}{3RK}, r_{\btheta, T}^* \Big) &\asymp RM r_{\btheta, T}^*\sqrt{\frac{d}{n}} + [RM^2+R\log(Rc_w^{-1})]\sqrt{\frac{1}{n}} + \sqrt{\frac{\log(RK/\delta)}{n}} \\
	&\lesssim T^2(\wtkappa_0 \vee \wtkappa_0')^{T-1}(r_w^* \vee r_{\btheta}^*) + R^2M^3c_w^{-1}\sqrt{\frac{d}{n|S|}} + R^2Mc_w^{-1}[M^2 + \log(Rc_w^{-1})]\sqrt{\frac{1}{n}} \\
	&\quad + MRc_w^{-1}\sqrt{\frac{\log(RK/\delta)}{n}} + \min\bigg\{h, R^2M^3c_w^{-1}\sqrt{\frac{d}{n}}\bigg\} + \epsilon RM^3c_w^{-1}\sqrt{\frac{d}{n}}.
\end{align}
Applying Theorem \ref{thm: generic appendix}, we have
\begin{align}
	&\max_{k \in S}\max_{r \in [R]}(\twonorm{\hthetakt{k}{T}_r - \bthetaks{k}_r} + \norm{\hwkt{k}{T}_r - \wks{k}_r}) \\
	&\leq 20T(\wtkappa_0)^{T-1}(r_w^* \vee r_{\btheta}^*) + \bigg[\frac{119}{15}\wtkappa_0(\wtkappa_0/119)^{T-1} + \frac{118}{119}(T-1)(\wtkappa_0)^T\bigg](r_w^* + r_{\btheta}^*)\\
	&\quad + \frac{1}{1-\kappa_0}\Big[\bar{\eta}\mathcal{E}_2\Big(n, |S|, \frac{\delta}{3R}\Big) + \mathcal{W}\Big(n, \frac{\delta}{3RK}, r_{\btheta, J}^*\Big)\Big] \\
	&\quad + \frac{18}{1-\wtkappa_0/119}\cdot\min\bigg\{3h, \frac{6}{1-\wtkappa_0}\Big[\mathcal{W}\Big(n, \frac{\delta}{3RK}, r_{\btheta}^*\Big) + 2\bar{\eta}\mathcal{E}_1\Big(n, \frac{\delta}{3RK}\Big)\Big]\bigg\} \\
	&\quad + \frac{30}{(1-\wtkappa_0)(1-\wtkappa_0/119)}\epsilon\cdot  \Big[\mathcal{W}\Big(n, \frac{\delta}{3RK}, r_{\btheta}^*\Big) + 2\bar{\eta}\mathcal{E}_1\Big(n, \frac{\delta}{3RK}\Big)\Big] \\
	&\leq T^2(\wtkappa_0 \vee \wtkappa_0')^{T-1}(r_w^* \vee r_{\btheta}^*) + R^2M^3c_w^{-1}\sqrt{\frac{d}{n|S|}} + R^2Mc_w^{-1}[M^2 + \log(Rc_w^{-1})]\sqrt{\frac{1}{n}} \\
	&\quad + MRc_w^{-1}\sqrt{\frac{\log(RK/\delta)}{n}} + \min\bigg\{h, R^2M^3c_w^{-1}\sqrt{\frac{d}{n}}\bigg\} + \epsilon RM^3c_w^{-1}\sqrt{\frac{d}{n}}. \label{eq: GMM cor 1}
\end{align}
Note that conditioned on the event $\mathcal{V}$ defined in the proof of Theorem \ref{thm: generic appendix}, 
\begin{equation}
	\etak{k}_r = (1+C_b)^{-1}(\hwkt{k}{0}_r)^{-1} \lesssim Rc_w^{-1},
\end{equation}
for all $k \in S$ and $r \in [R]$. Plugging it in equation \eqref{eq: GMM cor 1} implies the desired upper bound in Corollary \ref{cor: gmm appendix}.

\subsection{Proof of Proposition \ref{prop: mor appendix}}
Since this proof is very long, we divide it into several parts.

\underline{(\Rom{1}) Part 1:} Deriving the expressions of $\muk{k}_r$ and $\Lk{k}_r$.

First, note that 
\begin{equation}
	\qk{k}(\btheta) = \Qk{k}(\btheta|\bthetaks{k}, \bwks{k}) 
	= -\frac{1}{2}\tE\Bigg[\sum_{r=1}^R \gamma^{(r)}_{\bthetaks{k}, \bwks{k}}(\bxk{k}, \yk{k})(\yk{k} - (\bxk{k})^T\btheta_r)^2\Bigg].
\end{equation}
and
\begin{align}
	\hQk{k}(\btheta|\btheta', \bw') &= -\frac{1}{2n}\sum_{i=1}^{n}\sum_{r=1}^R\gamma^{(r)}_{\btheta', \bw'}(\bxk{k}_i)(\yk{k}_i - (\bxk{k}_i)^T\btheta_r)^2, \\
	\frac{\partial}{\partial \btheta_r}\Qk{k}(\btheta|\btheta', \bw') &= -\tE_{\bxk{k}}\big[\gamma^{(r)}_{\btheta', \bw'}(\bxk{k}, \yk{k})\bxk{k}((\bxk{k})^T\btheta_r-\yk{k})\big], \\
	\frac{\partial}{\partial \btheta_r}\hQk{k}(\btheta|\btheta', \bw') &= -\frac{1}{n}\sum_{i=1}^{n}\gamma^{(r)}_{\btheta', \bw'}(\bxk{k}_i, \yk{k}_i)\bxk{k}_i((\bxk{k}_i)^T\btheta_r-\yk{k}_i),\\
	\nabla_{\btheta}^2\qk{k}(\btheta) &= \textup{diag}\Big(\Big\{\tE\Big[\gamma^{(r)}_{\bthetaks{k}_r, \bwks{k}_r}(\bxk{k}, \yk{k})\bxk{k}(\bxk{k})^T\Big]\Big\}_{r=1}^R\Big). 
\end{align}
We have the following lemma.

\begin{lemma}\label{lem: mor eigenvalues}
	Under Assumption \ref{asmp: mor appendix}:
	\begin{enumerate}[(i)]
		\item $\lambdamax\big(\tE\big[\gamma^{(r)}_{\bthetaks{k}_r, \bwks{k}_r}(\bxk{k}, \yk{k})\bxk{k}(\bxk{k})^T\big]\big) \leq \wks{k}_r + C\frac{\sqrt{\log \Delta}}{\Delta} \coloneqq \Lk{k}_r$;
		\item $\lambdamin\big(\tE\big[\gamma^{(r)}_{\bthetaks{k}_r, \bwks{k}_r}(\bxk{k}, \yk{k})\bxk{k}(\bxk{k})^T\big]\big) \geq \wks{k}_r - CR\frac{\sqrt{\log \Delta}}{\Delta} \coloneqq \muk{k}_r$.
	\end{enumerate}
\end{lemma}
Now let us prove the lemma.

(\rom{1}) Note that
\begin{equation}
	\gamma^{(r)}_{\bwks{k}, \bthetaks{k}}(\bxk{k}, \yk{k}) = \frac{\wks{k}_r\exp\{\yk{k}(\bxk{k})^T(\bthetaks{k}_r - \bthetaks{k}_1) - \frac{1}{2}[((\bxk{k})^T\bthetaks{k}_r)^2 - ((\bxk{k})^T\bthetaks{k}_1)^2]\}}{\wks{k}_1 + \sum_{r=2}^R\wks{k}_r\exp\{\yk{k}(\bxk{k})^T(\bthetaks{k}_r - \bthetaks{k}_1) - \frac{1}{2}[((\bxk{k})^T\bthetaks{k}_r)^2 - ((\bxk{k})^T\bthetaks{k}_1)^2]\}},
\end{equation}
where
\begin{align}
	&\yk{k}(\bxk{k})^T(\bthetaks{k}_r - \bthetaks{k}_1) - \frac{1}{2}[((\bxk{k})^T\bthetaks{k}_r)^2 - ((\bxk{k})^T\bthetaks{k}_1)^2] \\
	&=  [\yk{k}-(\bxk{k})^T\bthetaks{k}_1]\cdot (\bxk{k})^T(\bthetaks{k}_r - \bthetaks{k}_1) + [(\bxk{k})^T(\bthetaks{k}_r - \bthetaks{k}_1)]\bigg[(\bxk{k})^T\bthetaks{k}_1 - \frac{1}{2}((\bxk{k})^T\bthetaks{k}_r \\
	&\quad\quad + (\bxk{k})^T\bthetaks{k}_1)\bigg] \\
	&= [\yk{k}-(\bxk{k})^T\bthetaks{k}_1]\cdot (\bxk{k})^T(\bthetaks{k}_r - \bthetaks{k}_1) - \frac{1}{2}[(\bxk{k})^T(\bthetaks{k}_r - \bthetaks{k}_1)]^2.
\end{align}
Conditioned on the event $\{\zk{k} = 1\}$, we have $[\yk{k}-(\bxk{k})^T\bthetaks{k}_1]\cdot (\bxk{k})^T(\bthetaks{k}_r - \bthetaks{k}_1) \sim N(\bm{0}, [(\bxk{k})^T(\bthetaks{k}_r - \bthetaks{k}_1)]^2)$. Define events
\begin{align}
	\mathcal{V}_1 &= \big\{|[\yk{k}-(\bxk{k})^T\bthetaks{k}_1]\cdot (\bxk{k})^T(\bthetaks{k}_r - \bthetaks{k}_1)| \leq \tau_1|(\bxk{k})^T(\bthetaks{k}_r - \bthetaks{k}_1)|\big\},\\
	\mathcal{V}_2 &= \big\{|(\bxk{k})^T(\bthetaks{k}_r - \bthetaks{k}_1)| \geq \tau_2\twonorm{\bthetaks{k}_r - \bthetaks{k}_1}\big\},
\end{align}
then by the tail bounds for Gaussian variables and the boundedness of Gaussian density, we have $\tP(\mathcal{V}_1^c) \lesssim \exp\{-C\tau_1^2\}$, $\tP(\mathcal{V}_2^c) \lesssim \tau_2$. Therefore,
\begin{align}
	&\lambdamax\Big(\tE\Big[\gamma^{(r)}_{\bthetaks{k}_r, \bwks{k}_r}(\bxk{k}, \yk{k})\bxk{k}(\bxk{k})^T\Big|\zk{k}=1\Big]\Big) \\
	&\lesssim \underbrace{\sup_{\twonorm{\bm{u}} = 1}\tE\Big[\gamma^{(r)}_{\bthetaks{k}_r, \bwks{k}_r}(\bxk{k}, \yk{k})((\bxk{k})^T\bm{u})^2\Big|\zk{k}=1\Big]\tP(\mathcal{V}_1 \cap \mathcal{V}_2)}_{[1]} + \underbrace{\sup_{\twonorm{\bm{u}} = 1}\sqrt{\tE[((\bxk{k})^T\bm{u})^4|\zk{k}=1]}\sqrt{\tP(\mathcal{V}_1^c)}}_{[2]} \\
	&\quad  + \underbrace{\sup_{\twonorm{\bm{u}} = 1}\tE\Big[\gamma^{(r)}_{\bthetaks{k}_r, \bwks{k}_r}(\bxk{k}, \yk{k})((\bxk{k})^T\bm{u})^2 \mathds{1}(\mathcal{V}_2^c)\Big|\zk{k}=1\Big]}_{[3]},
\end{align}
where
\begin{align}
	[1] &\lesssim \frac{\wks{k}_r}{\wks{k}_1}\cdot \tE\bigg[\exp\Big\{\tau_1 |(\bxk{k})^T(\bthetaks{k}_r - \bthetaks{k}_1)| - \frac{1}{2}|(\bxk{k})^T(\bthetaks{k}_r - \bthetaks{k}_1)|^2 \Big| \zk{k}=1, \mathcal{V}_1 \cap \mathcal{V}_2\Big\}\bigg] \\
	&\lesssim \frac{\wks{k}_r}{\wks{k}_1}\cdot\exp\Big\{\tau_1\tau_2 \twonorm{\bthetaks{k}_r - \bthetaks{k}_1} - \frac{1}{2}\tau_2^2\twonorm{\bthetaks{k}_r - \bthetaks{k}_1}^2\Big\},\\
	[2] &\lesssim \exp\{-C\tau_1^2\},\\
	[3] &\lesssim \sup_{\twonorm{\bm{u}} = 1}\tE[((\bxk{k})^T\bm{u})^2|\mathcal{V}_2]\cdot \tP(\mathcal{V}_2) \lesssim \tP(\mathcal{V}_2) \lesssim \tau_2, \label{eq: proof prop 3 eq condition}.
\end{align}
The second inequality in \eqref{eq: proof prop 3 eq condition} holds due to Lemma A.1 in \citet{kwon2020converges}. Let $\tau_1 = c\sqrt{\log \twonorm{\bthetaks{k}_r - \bthetaks{k}_1}}$, $\tau_2 = 3C\frac{\sqrt{\log \twonorm{\bthetaks{k}_r - \bthetaks{k}_1}}}{\twonorm{\bthetaks{k}_r - \bthetaks{k}_1}}$ with some constant $c > 0$. Note that $\twonorm{\bthetaks{k}_r - \bthetaks{k}_1} \leq \tau_1/\tau_2 = \twonorm{\bthetaks{k}_r - \bthetaks{k}_1}/3$. Then
\begin{equation}
	\lambdamax\Big(\tE\Big[\gamma^{(r)}_{\bthetaks{k}_r, \bwks{k}_r}(\bxk{k}, \yk{k})\bxk{k}(\bxk{k})^T\Big|\zk{k}=1\Big]\Big) \leq [1] + [2] + [3] \lesssim \frac{\wks{k}_r}{\wks{k}_1}\frac{1}{\Delta} + \frac{\sqrt{\log \Delta}}{\Delta}.
\end{equation}
Similarly, the same bound holds for $\lambdamax(\tE[\gamma^{(r)}_{\bthetaks{k}_r, \bwks{k}_r}(\bxk{k}, \yk{k})\bxk{k}(\bxk{k})^T|\zk{k}=r'])$ with all $r' \neq r$. In addition, we can rewrite $\gamma^{(r)}_{\bwks{k}, \bthetaks{k}}(\bxk{k}, \yk{k})$ as
\begin{align}
	\gamma^{(r)}_{\bwks{k}, \bthetaks{k}}(\bxk{k}, \yk{k}) &= \frac{\wks{k}_r}{\wks{k}_r + \sum_{r'\neq r}\wks{k}_{r'}\exp\{\yk{k}(\bxk{k})^T(\bthetaks{k}_{r'} - \bthetaks{k}_1) - \frac{1}{2}[((\bxk{k})^T\bthetaks{k}_{r'})^2 - ((\bxk{k})^T\bthetaks{k}_1)^2]\}} \\
	&\leq 1,
\end{align}
which implies that
\begin{equation}
	\lambdamax\Big(\tE\Big[\gamma^{(r)}_{\bthetaks{k}_r, \bwks{k}_r}(\bxk{k}, \yk{k})\bxk{k}(\bxk{k})^T\Big|\zk{k}=r\Big]\Big) \leq 1.
\end{equation}
Hence by the convexity of maximum eigenvalues,
\begin{align}
	\lambdamax\Big(\tE\Big[\gamma^{(r)}_{\bthetaks{k}_r, \bwks{k}_r}(\bxk{k}, \yk{k})\bxk{k}(\bxk{k})^T\Big]\Big) &\leq \sum_{r'=1}^R \wks{k}_{r'}\cdot \lambdamax\Big(\tE\Big[\gamma^{(r)}_{\bthetaks{k}_r, \bwks{k}_r}(\bxk{k}, \yk{k})\bxk{k}(\bxk{k})^T\Big|\zk{k}=r'\Big]\Big) \\
	&\leq  \wks{k}_r\Big(1+\frac{C}{\Delta}\Big) + C\frac{\sqrt{\log \Delta}}{\Delta} \\
	&\leq \wks{k}_r+ C'\frac{\sqrt{\log \Delta}}{\Delta}.
\end{align}

(\rom{2}) We have
\begin{align}
	\lambdamin\Big(\tE\Big[\gamma^{(r)}_{\bthetaks{k}_r, \bwks{k}_r}(\bxk{k}, \yk{k})\bxk{k}(\bxk{k})^T\Big]\Big) &\geq \wks{k}_r \lambdamin\Big(\tE\Big[\gamma^{(r)}_{\bthetaks{k}_r, \bwks{k}_r}(\bxk{k}, \yk{k})\bxk{k}(\bxk{k})^T\Big| \zk{k} = r\Big]\Big) \\
	&\geq \wks{k}_r \lambdamin\big(\tE\big[\bxk{k}(\bxk{k})^T\big| \zk{k} = r\big]\big) \\
	&\quad - \wks{k}_r \lambdamin\Big(\tE\Big[(1-\gamma^{(r)}_{\bthetaks{k}_r, \bwks{k}_r}(\bxk{k}, \yk{k}))\bxk{k}(\bxk{k})^T\Big| \zk{k} = r\Big]\Big).
\end{align}
Similar to (\rom{1}), it is straightforward to show that
\begin{align}
	&\lambdamax\Bigg(\tE\Bigg[\frac{\wks{k}_{r'}\exp\{\yk{k}(\bxk{k})^T(\bthetaks{k}_{r'} - \bthetaks{k}_1) - \frac{1}{2}[((\bxk{k})^T\bthetaks{k}_{r'})^2 - ((\bxk{k})^T\bthetaks{k}_1)^2]\}}{\wks{k}_r + \sum_{r'\neq r}\wks{k}_{r'}\exp\{\yk{k}(\bxk{k})^T(\bthetaks{k}_{r'} - \bthetaks{k}_1) - \frac{1}{2}[((\bxk{k})^T\bthetaks{k}_{r'})^2 - ((\bxk{k})^T\bthetaks{k}_1)^2]\}}\bxk{k}(\bxk{k})^T \\
	&\quad\quad\quad \Bigg| \zk{k} = r\Bigg]\Bigg) \lesssim \frac{\wks{k}_{r'}}{\wks{k}_r}\frac{1}{\Delta} + \frac{\sqrt{\log \Delta}}{\Delta},
\end{align}
for any $r' \neq r$. Hence
\begin{equation}
	\lambdamin\Big(\tE\Big[\gamma^{(r)}_{\bthetaks{k}_r, \bwks{k}_r}(\bxk{k}, \yk{k})\bxk{k}(\bxk{k})^T\Big]\Big) \geq \wks{k}_r - C\sum_{r' \neq r}\bigg(\wks{k}_{r'}\frac{1}{\Delta} + \wks{k}_r\frac{\sqrt{\log \Delta}}{\Delta}\bigg) \geq \wks{k}_r - CR\frac{\sqrt{\log \Delta}}{\Delta},
\end{equation}
which completes the proof of Lemma \ref{lem: mor eigenvalues}.

\underline{(\Rom{2}) Part 2:} Deriving the rate of $\kappa$ in Assumption 2.(\rom{1}). 

Since Assumption \ref{asmp: w appendix} is assumed to hold for all $k \in [K]$, in this part, for notation simplicity, we drop the task index $k$ in the superscript and write $\bwk{k} = \{\wk{k}_r\}_{r=1}^R$, $\bthetak{k} = \{\bthetak{k}_r\}_{r=1}^R$, $\bwks{k} = \{\wks{k}_r\}_{r=1}^R$, $\bthetaks{k} = \{\bthetaks{k}_r\}_{r=1}^R$, $\bxk{k}$, and $\yk{k}$ simply as $\bw= \{w_r\}_{r=1}^R$, $\btheta= \{\btheta_r\}_{r=1}^R$, $\bw^*= \{w^*_r\}_{r=1}^R$, $\bthetak{k}= \{\btheta^*_r\}_{r=1}^R$, $\bx$, and $y$.

By Taylor expansion:
\begin{equation}
	\tE\big[\gamma^{(r)}_{\btheta, \bw}(\bx,y) - \gamma^{(r)}_{\btheta^*, \bw^*}(\bx,y)\big] = \sum_{r'=1}^R \tE\Bigg[\frac{\partial \gamma^{(r)}_{\btheta, \bw}(\bx,y)}{\partial w_{r'}}\bigg|_{w_{r'} = \widetilde{w}_{r'}} (w_{r'} - w^*_{r'})\Bigg] + \sum_{r'=1}^R \tE\Bigg[\bigg(\frac{\partial \gamma^{(r)}_{\btheta, \bw}(\bx,y)}{\partial \btheta_{r'}}\bigg|_{\btheta_{r'} = \widetilde{\btheta}_{r'}}\bigg)^T (\btheta_{r'} - \btheta^*_{r'})\Bigg],
\end{equation}
where $\widetilde{w}_{r'}$ is at the line segment between $w_{r'}$ and $w^*_r$, and $\widetilde{\btheta}_{r'}$ is at the line segment between $\btheta_{r'}$ and $\btheta^*_r$. And
\begin{align}
	\frac{\partial \gamma^{(r)}_{\btheta, \bw}(\bx,y)}{\partial w_{r'}} = \begin{cases}
		\frac{\exp\{y\bx^T(\btheta_r - \btheta_1) - \frac{1}{2}[(\bx^T\btheta_r)^2 - (\bx^T\btheta_1)^2]\}(w_1 + \sum_{r' \neq r}w_{r'}\exp\{y\bx^T(\btheta_{r'} - \btheta_1) - \frac{1}{2}[(\bx^T\btheta_{r'})^2 - (\bx^T\btheta_1)^2]\})}{(w_1 + \sum_{r'=1}^Rw_{r'}\exp\{y\bx^T(\btheta_{r'} - \btheta_1) - \frac{1}{2}[(\bx^T\btheta_{r'})^2 - (\bx^T\btheta_1)^2]\})^2}, &r' = r;\\
		-\frac{w_r\exp\{y\bx^T(\btheta_r - \btheta_1) - \frac{1}{2}[(\bx^T\btheta_r)^2 - (\bx^T\btheta_1)^2]\}\cdot \exp\{y\bx^T(\btheta_{r'} - \btheta_1) - \frac{1}{2}[(\bx^T\btheta_{r'})^2 - (\bx^T\btheta_1)^2]\}}{(w_1 + \sum_{r'=1}^Rw_{r'}\exp\{y\bx^T(\btheta_{r'} - \btheta_1) - \frac{1}{2}[(\bx^T\btheta_{r'})^2 - (\bx^T\btheta_1)^2]\})^2}, &r' \neq r.
	\end{cases}
\end{align}
We want to upper bound $\tE\Big[\frac{\partial \gamma^{(r)}_{\btheta, \bw}(\bx,y)}{\partial w_r}\Big|z=\tilde{r}\Big]$ for all $\tilde{r}$. Since in the expression of $\tE\Big[\frac{\partial \gamma^{(r)}_{\btheta, \bw}(\bx,y)}{\partial w_r}\Big|z=\tilde{r}\Big]$, $\{w_{r'}\}_{r' \neq r}$ and $\{\btheta_{r'}\}_{r' \neq r}$ are symmetric, i.e., any class can be the reference class. WLOG, we only show how to bound $\tE\Big[\frac{\partial \gamma^{(r)}_{\btheta, \bw}(\bx,y)}{\partial w_r}\Big|z=1\Big]$ and the same arguments can be used to bound $\tE\Big[\frac{\partial \gamma^{(r)}_{\btheta, \bw}(\bx,y)}{\partial w_r}\Big|z=\tilde{r}\Big]$ for $\widetilde{r} \neq 1$.

Denote $(*) = \yk{1}\bx^T(\btheta_r - \btheta_1) - \frac{1}{2}[(\bx^T\btheta_r)^2 - (\bx^T\btheta_1)^2]$, $\widetilde{z} = (\yk{1} - \bx^T\btheta_1^*)\cdot \bx^T(\btheta_r - \btheta_1)|\bx \sim N(0, [\bx^T(\btheta_r - \btheta_1)]^2)$, where $\yk{1} \overset{d}{=} (y|z = 1)$. Then
\begin{equation}
	(*) = \widetilde{z} + (\bx^T\btheta_1^*)\cdot x^T(\btheta_r - \btheta_1) - \frac{1}{2}x^T(\btheta_r + \btheta_1)\cdot x^T(\btheta_r - \btheta_1),
\end{equation}
where 
\begin{align}
	&(\bx^T\btheta_1^*)\cdot x^T(\btheta_r - \btheta_1) - \frac{1}{2}x^T(\btheta_r + \btheta_1)\cdot x^T(\btheta_r - \btheta_1) \\
	&= \big[x^T(\btheta_r^* - \btheta_1^*) + x^T(\btheta_1^* - \btheta_1 + \btheta_r - \btheta_r^*)\big]\bigg[-\frac{1}{2}x^T(\btheta_r^* - \btheta_1^*) +-\frac{1}{2}x^T(\btheta_1^* - \btheta_1) + \frac{1}{2}x^T(\btheta_r^* - \btheta_r)\bigg] \\
	&= -\frac{1}{2}[x^T(\btheta_r^* - \btheta_1^*)]^2 + \frac{1}{2}\{x^T[(\btheta_1^* - \btheta_1) + (\btheta_r^* - \btheta_r)]\}^2.
\end{align}
Define events
\begin{align}
	\mathcal{V}_1 &= \bigg\{|x^T(\btheta_1^* - \btheta_1)| \vee |x^T(\btheta_r^* - \btheta_r)| \leq \frac{1}{4}|x^T(\btheta_r^* - \btheta_1^*)|\bigg\},\\
	\mathcal{V}_2 &= \{|\widetilde{z}| \leq \tau_1 |x^T(\btheta_r - \btheta_1)|\}, \\
	\mathcal{V}_3 &= \{|\bx^T(\btheta_r^* - \btheta_1^*)| > \tau_2 |\btheta_r^* - \btheta_1^*|\}.
\end{align}
We know that
\begin{align}
	\tP(\mathcal{V}_1^c) &\leq \tP\bigg(|x^T(\btheta_1^* - \btheta_1)| \vee |x^T(\btheta_r^* - \btheta_r)| > \frac{1}{4}|x^T(\btheta_r^* - \btheta_1^*)|\bigg) \\
	&\leq \tP\bigg(|x^T(\btheta_1^* - \btheta_1)| > \frac{1}{4}|x^T(\btheta_r^* - \btheta_1^*)|\bigg) + \tP\bigg(|x^T(\btheta_r^* - \btheta_r)| > \frac{1}{4}|x^T(\btheta_r^* - \btheta_1^*)|\bigg) \\
	&\leq 4\bigg(\frac{\twonorm{\btheta_1^* - \btheta_1}}{\twonorm{\btheta_r^* - \btheta_1^*}} + \frac{\twonorm{\btheta_r^* - \btheta_r}}{\twonorm{\btheta_r^* - \btheta_1^*}}\bigg) \\
	&\leq 8C_b,
\end{align}
where we applied Lemma A.1 in \citet{kwon2020converges} to get the second last inequality. And
\begin{equation}
	\tP(\mathcal{V}_2^c) \leq C\exp\{-C'\tau_1^2\}, \quad \tP(\mathcal{V}_3^c) \leq C\tau_2.
\end{equation}
Given $\mathcal{V}_1 \cap \mathcal{V}_2 \cap \mathcal{V}_3$, we have
\begin{equation}
	-\frac{1}{2}[x^T(\btheta_r^* - \btheta_1^*)]^2 + \frac{1}{2}\{x^T[(\btheta_1^* - \btheta_1) + (\btheta_r^* - \btheta_r)]\}^2 \leq -\frac{3}{8}[x^T(\btheta_r^* - \btheta_1^*)]^2,
\end{equation}
leading to
\begin{equation}
	\widetilde{z} \leq \tau_1|x^T(\btheta_r - \btheta_1)| \leq \tau_1(|x^T(\btheta_r^* - \btheta_1^*)| + |x^T(\btheta_r - \btheta_r^*)| + |x^T(\btheta_1 - \btheta_1^*)|) \leq \tau_1\cdot \frac{3}{2}|x^T(\btheta_r^* - \btheta_1^*)|.
\end{equation}
Hence
\begin{align}
	\Bigg|\tE\bigg[\frac{\partial \gamma^{(r)}_{\btheta, \bw}(\bx,y)}{\partial w_r}\bigg|z=1\bigg]\Bigg| &\leq \Bigg|\tE\bigg[\frac{\partial \gamma^{(r)}_{\btheta, \bw}(\bx,y)}{\partial w_r}\bigg|z=1, \mathcal{V}_1 \cap \mathcal{V}_2 \cap \mathcal{V}_3\bigg]\tP(\mathcal{V}_1 \cap \mathcal{V}_2 \cap \mathcal{V}_3)\Bigg| + \frac{R}{c_w}[\tP(\mathcal{V}_1^c) + \tP(\mathcal{V}_2^c) + \tP(\mathcal{V}_3^c)] \\
	&\leq \frac{R}{c_w}\exp\bigg\{\frac{3}{2}\tau_1|x^T(\btheta_r^* - \btheta_1^*)|-\frac{3}{8}[x^T(\btheta_r^* - \btheta_1^*)]^2\bigg\} + C\frac{R}{c_w}\big(C_b + \exp\{-C'\tau_1^2\}+\tau_2\big) \\
	&\leq \frac{R}{c_w}\exp\bigg\{\frac{3}{2}\tau_1\tau_2\twonorm{\btheta_r^* - \btheta_1^*}-\frac{3}{8}\tau_2^2\twonorm{\btheta_r^* - \btheta_1^*}^2\bigg\} + C\frac{R}{c_w}\big(C_b + \exp\{-C'\tau_1^2\}+\tau_2\big),
\end{align}
where the last inequality requires $|x^T(\btheta_r^* - \btheta_1^*)| \geq \tau_2\twonorm{\btheta_r^* - \btheta_1^*} \geq 2\tau_1$. Let $\tau_1 = c\sqrt{\log \twonorm{\btheta_r^* - \btheta_1^*}}$ and $\tau_2 = 10c\frac{\sqrt{\log \twonorm{\btheta_r^* - \btheta_1^*}}}{\twonorm{\btheta_r^* - \btheta_1^*}}$ with some constant $c > 0$, then
\begin{equation}
	\Bigg|\tE\bigg[\frac{\partial \gamma^{(r)}_{\btheta, \bw}(\bx,y)}{\partial w_r}\bigg|z=1\bigg]\Bigg| \lesssim \frac{R}{c_w}\bigg(C_b + \frac{\sqrt{\log \twonorm{\btheta_r^* - \btheta_1^*}}}{\twonorm{\btheta_r^* - \btheta_1^*}}\bigg) \lesssim \frac{R}{c_w}\bigg(C_b + \frac{\sqrt{\log \Delta}}{\Delta}\bigg). 
\end{equation}
Similarly,
\begin{equation}
	\Bigg|\tE\bigg[\frac{\partial \gamma^{(r)}_{\btheta, \bw}(\bx,y)}{\partial w_{r'}}\bigg|z=\widetilde{r}\bigg]\Bigg| \lesssim \frac{R}{c_w}\bigg(C_b + \frac{\sqrt{\log \Delta}}{\Delta}\bigg),
\end{equation}
for any $r$, $r'$, and $\widetilde{r}$. Therefore,
\begin{equation}
	\sum_{r'=1}^R \tE\Bigg[\frac{\partial \gamma^{(r)}_{\btheta, \bw}(\bx,y)}{\partial w_{r'}}\bigg|_{w_{r'} = \widetilde{w}_{r'}} (w_{r'} - w^*_{r'})\Bigg]  \leq  C\frac{R}{c_w}\bigg(C_b + \frac{\sqrt{\log \Delta}}{\Delta}\bigg)\sum_{r=1}^R |w_r - w_r^*|.
\end{equation}

On the other hand,
\begin{equation}\label{eq: gamma derivative theta}
	\frac{\partial \gamma^{(r)}_{\bw,\btheta}(\bx, y)}{\partial \btheta_{r'}} = \begin{cases}
		\gamma^{(r)}_{\bw,\btheta}(\bx, y)\bx(y - \bx^T\btheta_r) - \big(\gamma^{(r)}_{\bw,\btheta}(\bx, y)\big)^2\bx(y - \bx^T\btheta_r), \quad &r' = r;\\
		-\gamma^{(r)}_{\bw,\btheta}(\bx, y)\gamma^{(r')}_{\bw,\btheta}(\bx, y)\bx(y - \bx^T\btheta_{r'}),\quad &r' \neq r,
	\end{cases}
\end{equation}
where
\begin{equation}
	\gamma^{(r)}_{\bw, \btheta}(\bx, y) = \frac{w_r\exp\{y\bx^T(\btheta_r - \btheta_1) - \frac{1}{2}[(\bx^T\btheta_r)^2 - (\bx^T\btheta_1)^2]\}}{w_1 + \sum_{r'=2}^Rw_{r'}\exp\{y\bx^T(\btheta_{r'} - \btheta_1) - \frac{1}{2}[(\bx^T\btheta_{r'})^2 - (\bx^T\btheta_1)^2]\}},
\end{equation}
We will show how to upper bound $\tE[\gamma^{(r)}_{\bw, \btheta}(\bx, y)(y - \bx^T\btheta_r)\cdot \bx^T(\btheta_r - \btheta_r')]$, and the same arguments can be used to bound $\tE[(\gamma^{(r)}_{\bw, \btheta}(\bx, y))^2(y - \bx^T\btheta_r)\cdot \bx^T(\btheta_r - \btheta_r')]$. Then we will have an upper bound for $\tE\Big[\big(\frac{\partial \gamma^{(r)}_{\btheta, \bw}(\bx,y)}{\partial \btheta_r}\big|_{\btheta_r = \widetilde{\btheta}_r}\big)^T (\btheta_r - \btheta_r')\Big]$, and the analysis is the same for $\tE\Big[\big(\frac{\partial \gamma^{(r)}_{\btheta, \bw}(\bx,y)}{\partial \btheta_{r'}}\big|_{\btheta_{r'} = \widetilde{\btheta}_{r'}}\big)^T (\btheta_{r'} - \btheta^*_{r'})\Big]$ with $r' \neq r$. 

Let us start from  $\tE[\gamma^{(r)}_{\bw, \btheta}(\bx, y)(y - \bx^T\btheta_r)\cdot \bx^T(\btheta_r - \btheta_r')|z = 1]$. Consider $\yk{1} \overset{d}{=} (y|z = 1)$ and
\begin{equation}
	\yk{1} - \bx^T\btheta_r = (\yk{1} - \bx^T\btheta_1^*) + \bx^T(\btheta_1^*-\btheta_r^*) + \bx^T(\btheta_r^*-\btheta_r).
\end{equation}
Define events
\begin{align}
	\mathcal{V}_1 &= \bigg\{|x^T(\btheta_1^* - \btheta_1)| \vee |x^T(\btheta_r^* - \btheta_r)| \leq \frac{1}{4}|x^T(\btheta_r^* - \btheta_1^*)|\bigg\},\\
	\mathcal{V}_2 &= \{|\yk{1} - \bx^T\btheta_1^*| \leq \tau_1\}, \\
	\mathcal{V}_3 &= \{|\bx^T(\btheta_r^* - \btheta_1^*)| > \tau_2 |\btheta_r^* - \btheta_1^*|\}.
\end{align}
We know that
\begin{align}
	\tP(\mathcal{V}_1^c) &\leq \tP\bigg(|x^T(\btheta_1^* - \btheta_1)| \vee |x^T(\btheta_r^* - \btheta_r)| > \frac{1}{4}|x^T(\btheta_r^* - \btheta_1^*)|\bigg) \\
	&\leq \tP\bigg(|x^T(\btheta_1^* - \btheta_1)| > \frac{1}{4}|x^T(\btheta_r^* - \btheta_1^*)|\bigg) + \tP\bigg(|x^T(\btheta_r^* - \btheta_r)| > \frac{1}{4}|x^T(\btheta_r^* - \btheta_1^*)|\bigg) \\
	&\leq 4\bigg(\frac{\twonorm{\btheta_1^* - \btheta_1}}{\twonorm{\btheta_r^* - \btheta_1^*}} + \frac{\twonorm{\btheta_r^* - \btheta_r}}{\twonorm{\btheta_r^* - \btheta_1^*}}\bigg) \\
	&\leq 8C_b,
\end{align}
where we applied Lemma A.1 in \citet{kwon2020converges} to get the second last inequality. And
\begin{equation}
	\tP(\mathcal{V}_2^c) \leq C\exp\{-C'\tau_1^2\}, \quad \tP(\mathcal{V}_3^c) \leq C\tau_2.
\end{equation}
Note that
\begin{align}
	&\tE\big[\gamma^{(r)}_{\bw, \btheta}(\bx, y)(y - \bx^T\btheta_r)\cdot \bx^T(\btheta_r - \btheta_r')|z = 1\big] \\
	&\leq \underbrace{\tE\big[\gamma^{(r)}_{\bw, \btheta}(\bx, y)(y - \bx^T\btheta_r)\cdot \bx^T(\btheta_r - \btheta_r')|z = 1, \mathcal{V}_1\cap \mathcal{V}_2 \cap \mathcal{V}_3\big] \tP(\mathcal{V}_1\cap \mathcal{V}_2 \cap \mathcal{V}_3)}_{[1]} \\
	&\quad + \underbrace{\tE\big[\bx^T(\btheta_r - \btheta_r')(\yk{1} - \bx^T\btheta_r)\mathds{1}(\mathcal{V}_1^c \cup \mathcal{V}_2^c \cup \mathcal{V}_3^c)\big]}_{[2]}.
\end{align}

(a) Case 1: $\max_{r}\twonorm{\btheta_r - \btheta_r^*} \leq 1$.

Note that for term [2], by Lemma A.1 in \citet{kwon2020converges},
\begin{align}
	&\tE\big[\bx^T(\btheta_r - \btheta_r')(\yk{1} - \bx^T\btheta_1^*)\mathds{1}(\mathcal{V}_1^c)\big] \\
	&\leq \tE\bigg[\bx^T(\btheta_r - \btheta_r')(\yk{1} - \bx^T\btheta_1^*)\mathds{1}(\mathcal{V}_1^c)\bigg||x^T(\btheta_1^* - \btheta_1)| > \frac{1}{4}|x^T(\btheta_r^* - \btheta_1^*)|\bigg]\tP\bigg(|x^T(\btheta_1^* - \btheta_1)| > \frac{1}{4}|x^T(\btheta_r^* - \btheta_1^*)|\bigg) \\
	&\quad + \tE\bigg[\bx^T(\btheta_r - \btheta_r')(\yk{1} - \bx^T\btheta_1^*)\mathds{1}(\mathcal{V}_1^c)\bigg||x^T(\btheta_r^* - \btheta_r)| > \frac{1}{4}|x^T(\btheta_r^* - \btheta_1^*)|\bigg]\tP\bigg(|x^T(\btheta_r^* - \btheta_r)| > \frac{1}{4}|x^T(\btheta_r^* - \btheta_1^*)|\bigg)  \\
	&\leq \sqrt{\tE\bigg[(\bx^T(\btheta_r - \btheta_r'))^2\bigg||x^T(\btheta_1^* - \btheta_1)| > \frac{1}{4}|x^T(\btheta_r^* - \btheta_1^*)|\bigg]} \cdot \sqrt{\tE(\yk{1} - \bx^T\btheta_1^*)^2} \\
	&\quad \cdot \tP\bigg(|x^T(\btheta_1^* - \btheta_1)| > \frac{1}{4}|x^T(\btheta_r^* - \btheta_1^*)|\bigg) \\
	&\quad + \sqrt{\tE\bigg[(\bx^T(\btheta_r - \btheta_r'))^2\bigg||x^T(\btheta_1^* - \btheta_1)| > \frac{1}{4}|x^T(\btheta_r^* - \btheta_1^*)|\bigg]} \cdot \sqrt{\tE(\yk{1} - \bx^T\btheta_1^*)^2} \\
	&\quad \cdot \tP\bigg(|x^T(\btheta_r^* - \btheta_r)| > \frac{1}{4}|x^T(\btheta_r^* - \btheta_1^*)|\bigg) \\
	&\lesssim C_b \twonorm{\btheta_r - \btheta_r'}.
\end{align}
And by Cauchy-Schwarz inequality,
\begin{align}
	\tE\big[\bx^T(\btheta_r - \btheta_r')(\yk{1} - \bx^T\btheta_1^*)\mathds{1}(\mathcal{V}_2^c)\big] &\leq \sqrt{\tE\big[(\bx^T(\btheta_r - \btheta_r'))^2(\yk{1} - \bx^T\btheta_1^*)^2\big]}\sqrt{\tP(\mathcal{V}_2^c)} \lesssim \twonorm{\btheta_r - \btheta_r'}\cdot \exp\{-C\tau_1^2\},\\
	\tE\big[\bx^T(\btheta_r - \btheta_r')(\yk{1} - \bx^T\btheta_1^*)\mathds{1}(\mathcal{V}_3^c)\big] &\leq \sqrt{\tE\big[(\bx^T(\btheta_r - \btheta_r'))^2\big]\cdot \tP(\mathcal{V}_2^c)}\sqrt{\tE\big[(\yk{1} - \bx^T\btheta_1^*)^2\big]\cdot \tP(\mathcal{V}_2^c)} \lesssim \twonorm{\btheta_r - \btheta_r'}\cdot \tau_2.
\end{align}
Combine them together:
\begin{equation}
	\tE\big[\bx^T(\btheta_r - \btheta_r')(\yk{1} - \bx^T\btheta_1^*)\mathds{1}(\mathcal{V}_1^c \cup \mathcal{V}_2^c \cup \mathcal{V}_3^c)\big] \lesssim (C_b + \exp\{-C\tau_1^2\} + \tau_2)\twonorm{\btheta_r - \btheta_r'}.
\end{equation}
Furthermore,
\begin{align}
	\tE\big[\bx^T(\btheta_r - \btheta_r')\cdot \bx^T(\btheta_1^* - \btheta_r^*)\mathds{1}(\mathcal{V}_1^c)\big] &\leq \sqrt{\tE\big[(\bx^T(\btheta_r - \btheta_r'))^2|\mathcal{V}_1^c\big]\cdot \tP(\mathcal{V}_1^c)}\cdot \sqrt{\tE\big[(\bx^T(\btheta_1^* - \btheta_r^*))^2|\mathcal{V}_1^c\big]\cdot \tP(\mathcal{V}_1^c)} \\
	&\lesssim \twonorm{\btheta_r - \btheta_r'}\cdot C_b \cdot \sqrt{\tE\big[|x^T(\btheta_1^* - \btheta_1)|^2 \vee |x^T(\btheta_r^* - \btheta_r)|^2|\mathcal{V}_1^c\big]} \\
	&\lesssim C_b \twonorm{\btheta_r - \btheta_r'},
\end{align}
and
\begin{align}
	\tE\big[\bx^T(\btheta_r - \btheta_r')\cdot \bx^T(\btheta_1^* - \btheta_r^*)\mathds{1}(\mathcal{V}_2^c)\big] &\lesssim \exp\{-C\tau_1^2\}\cdot \twonorm{\btheta_1^* - \btheta_r^*}\cdot \twonorm{\btheta_r - \btheta_r'},\\
	\tE\big[\bx^T(\btheta_r - \btheta_r')\cdot \bx^T(\btheta_1^* - \btheta_r^*)\mathds{1}(\mathcal{V}_3^c)\big] &\lesssim \sqrt{\tE\big[(\bx^T(\btheta_r - \btheta_r'))^2|\mathcal{V}_3^c\big]\cdot \tP(\mathcal{V}_3^c)}\cdot \sqrt{\tE\big[(\bx^T(\btheta_1^* - \btheta_r^*))^2|\mathcal{V}_3^c\big]\cdot \tP(\mathcal{V}_3^c)} \\
	&\lesssim \tau_2^2\twonorm{\btheta_r - \btheta_r'}\cdot \twonorm{\btheta_1^* -\btheta_r^*}.
\end{align}
Therefore 
\begin{equation}
	\tE\big[\bx^T(\btheta_r - \btheta_r')\cdot \bx^T(\btheta_1^* - \btheta_r^*)\mathds{1}(\mathcal{V}_1^c \cup \mathcal{V}_2^c \cup \mathcal{V}_3^c)\big] \lesssim (C_b + \exp\{-C\tau_1^2\}+ \tau_2^2\twonorm{\btheta_1^* -\btheta_r^*})\twonorm{\btheta_r - \btheta_r'}.
\end{equation}
Similarly,
\begin{align}
	\tE\big[\bx^T(\btheta_r - \btheta_r')\cdot \bx^T(\btheta_r^* - \btheta_r)\mathds{1}(\mathcal{V}_1^c \cup \mathcal{V}_2^c \cup \mathcal{V}_3^c)\big] &\lesssim (C_b + \exp\{-C\tau_1^2\}+ \tau_2^2\twonorm{\btheta_r^* -\btheta_r})\twonorm{\btheta_r - \btheta_r'}\\
	&\lesssim (C_b + \exp\{-C\tau_1^2\}+ \tau_2^2)\twonorm{\btheta_r - \btheta_r'}.
\end{align}
Therefore we have
\begin{equation}
	[2] \lesssim (C_b + \exp\{-C\tau_1^2\}+ \tau_2+\tau_2^2\twonorm{\btheta_1^* -\btheta_r^*})\twonorm{\btheta_r - \btheta_r'}.
\end{equation}
For [1], given $\mathcal{V}_1\cap \mathcal{V}_2 \cap \mathcal{V}_3$, we know that
\begin{equation}
	-\frac{1}{2}[x^T(\btheta_r^* - \btheta_1^*)]^2 + \frac{1}{2}\{x^T[(\btheta_1^* - \btheta_1) + (\btheta_r^* - \btheta_r)]\}^2 \leq -\frac{3}{8}[x^T(\btheta_r^* - \btheta_1^*)]^2,
\end{equation}
leading to $(\yk{1} - \bx^T\btheta_1^*) + \bx^T(\btheta_1^*-\btheta_r^*)$
\begin{equation}
	(\yk{1} - \bx^T\btheta_1^*)\cdot \bx^T(\btheta_r-\btheta_1) \leq \tau_1|x^T(\btheta_r - \btheta_1)| \leq \tau_1(|x^T(\btheta_r^* - \btheta_1^*)| + |x^T(\btheta_r - \btheta_r^*)| + |x^T(\btheta_1 - \btheta_1^*)|) \leq \tau_1\cdot \frac{3}{2}|x^T(\btheta_r^* - \btheta_1^*)|.
\end{equation}
Similar to the previous analysis, we can show that
\begin{align}
	[1] &\lesssim \frac{R}{c_w}\exp\bigg\{\frac{3}{2}\tau_1|x^T(\btheta_r^* - \btheta_1^*)| - \frac{3}{8}|x^T(\btheta_r^* - \btheta_1^*)|^2\bigg\}(\tau_1 + 1)\cdot \twonorm{\btheta_r - \btheta_r'}\\
	&\lesssim \frac{R}{c_w}\exp\bigg\{\frac{3}{2}\tau_1\tau_2\twonorm{\btheta_r^* - \btheta_1^*} - \frac{3}{8}\tau_2^2\twonorm{\btheta_r^* - \btheta_1^*} ^2\bigg\}(\tau_1 + 1)\cdot \twonorm{\btheta_r - \btheta_r'}.
\end{align}
This implies that
\begin{align}
	&\tE\big[\gamma^{(r)}_{\bw, \btheta}(\bx, y)(y - \bx^T\btheta_r)\cdot \bx^T(\btheta_r - \btheta_r')|z = 1\big] \\
	&\leq [1] + [2] \\
	&\lesssim \Bigg[\frac{R}{c_w}\exp\bigg\{\frac{3}{2}\tau_1\tau_2\twonorm{\btheta_r^* - \btheta_1^*} - \frac{3}{8}\tau_2^2\twonorm{\btheta_r^* - \btheta_1^*} ^2\bigg\}(\tau_1 + 1) + C_b + \exp\{-C\tau_1^2\}+ \tau_2+\tau_2^2\twonorm{\btheta_1^* -\btheta_r^*}\Bigg]\cdot \twonorm{\btheta_r - \btheta_r'}. \label{eq: theta derivative}
\end{align}
Let $\tau_1 = c\sqrt{\log \twonorm{\btheta_r^* -\btheta_1^*}}$ and $\tau_2 = 10c\frac{\sqrt{\log \twonorm{\btheta_r^* -\btheta_1^*}}}{\twonorm{\btheta_r^* -\btheta_1^*}}$ with some constant $c > 0$. Then 
\begin{equation}
	\text{RHS of \eqref{eq: theta derivative}} \lesssim \bigg(\frac{R}{c_w}\frac{1}{\Delta}\sqrt{\log \Delta} + \frac{R}{c_w}C_b\bigg) \twonorm{\btheta_r - \btheta_r'}.
\end{equation}
Similarly, we can show that $\tE\big[\gamma^{(r)}_{\bw, \btheta}(\bx, y)(y - \bx^T\btheta_r)\cdot \bx^T(\btheta_r - \btheta_r')|z = r'\big]$ has the same upper bound. Therefore
\begin{equation}
	\tE\big[\gamma^{(r)}_{\bw, \btheta}(\bx, y)(y - \bx^T\btheta_r)\cdot \bx^T(\btheta_r - \btheta_r')\big]\lesssim \bigg(\frac{R}{c_w}\frac{1}{\Delta}\sqrt{\log \Delta} + \frac{R}{c_w}C_b\bigg) \twonorm{\btheta_r - \btheta_r'}.
\end{equation}
Similarly, following the same arguments, it can be shown that
\begin{equation}
	\tE\big[(\gamma^{(r)}_{\bw, \btheta}(\bx, y))^2(y - \bx^T\btheta_r)\cdot \bx^T(\btheta_r - \btheta_r')\big]\lesssim \bigg(\frac{R}{c_w}\frac{1}{\Delta}\sqrt{\log \Delta} + \frac{R}{c_w}C_b\bigg) \twonorm{\btheta_r - \btheta_r'}.
\end{equation}
Hence by \eqref{eq: gamma derivative theta},
\begin{equation}
	\Bigg|\tE\bigg[\bigg(\frac{\partial \gamma^{(r)}_{\btheta, \bw}(\bx,y)}{\partial \btheta_r}\big|_{\btheta_r = \widetilde{\btheta}_r}\bigg)^T (\btheta_r - \btheta^*_r)\bigg] \Bigg| \lesssim \bigg(\frac{R}{c_w}\frac{1}{\Delta}\sqrt{\log \Delta} + \frac{R}{c_w}C_b\bigg) \twonorm{\btheta_r - \btheta_r'}.
\end{equation}

(b) Case 2: $\max_{r}\twonorm{\btheta_r - \btheta_r^*} > 1$. 

Suppose $r_0 \in [R]$ satisfies $\twonorm{\btheta_{r_0} - \btheta_{r_0}^*} > 1$. For $r' \neq r$, we have
\begin{equation}
	\big|\tE\big[\gamma^{(r)}_{\bw, \btheta}(\bx, y) - \gamma^{(r)}_{\bw^*, \btheta^*}(\bx, y)\big| z = r'\big]\big| \leq \big|\tE\big[\gamma^{(r)}_{\bw, \btheta}(\bx, y)\big| z = r'\big]\big| + \big|\tE\big[\gamma^{(r)}_{\bw^*, \btheta^*}(\bx, y)\big| z = r'\big]\big|.
\end{equation}
WLOG, let us consider the case $r' = 1 \neq r$, and the other cases can be similarly discussed. We have
\begin{equation}
	\tE\big[\gamma^{(r)}_{\bw, \btheta}(\bx, y)\big|z = 1\big] = \tE\Bigg[\frac{w_r\exp\{y\bx^T(\btheta_r - \btheta_1) - \frac{1}{2}[(\bx^T\btheta_r)^2 - (\bx^T\btheta_1)^2]\}}{w_1 + \sum_{r'=2}^Rw_{r'}\exp\{y\bx^T(\btheta_{r'} - \btheta_1) - \frac{1}{2}[(\bx^T\btheta_{r'})^2 - (\bx^T\btheta_1)^2]\}}\Bigg].
\end{equation}
Recall that
\begin{align}
	\mathcal{V}_1 &= \bigg\{|x^T(\btheta_1^* - \btheta_1)| \vee |x^T(\btheta_r^* - \btheta_r)| \leq \frac{1}{4}|x^T(\btheta_r^* - \btheta_1^*)|\bigg\},\\
	\mathcal{V}_2 &= \{|\yk{1} - \bx^T\btheta_1^*| \leq \tau_1\}, \\
	\mathcal{V}_3 &= \{|\bx^T(\btheta_r^* - \btheta_1^*)| > \tau_2 |\btheta_r^* - \btheta_1^*|\}.
\end{align}
and 
\begin{equation}
	\tP(\mathcal{V}_1^c) \leq 8C_b,\quad \tP(\mathcal{V}_2^c) \leq C\exp\{-C'\tau_1^2\}, \quad \tP(\mathcal{V}_3^c) \leq C\tau_2.
\end{equation}
Similar to the previous analysis,
\begin{equation}
	\yk{1}\bx^T(\btheta_r - \btheta_1) - \frac{1}{2}[(\bx^T\btheta_r)^2 - (\bx^T\btheta_1)^2] = (\yk{1}-\bx^T\btheta_1^*)\bx^T(\btheta_r - \btheta_1) + (\bx^T\btheta_1^*)\cdot x^T(\btheta_r - \btheta_1) - \frac{1}{2}x^T(\btheta_r + \btheta_1)\cdot x^T(\btheta_r - \btheta_1).
\end{equation}
Conditioned on $\mathcal{V}_1 \cap \mathcal{V}_2 \cap \mathcal{V}_3$,
\begin{align}
	&(\bx^T\btheta_1^*)\cdot x^T(\btheta_r - \btheta_1) - \frac{1}{2}x^T(\btheta_r + \btheta_1)\cdot x^T(\btheta_r - \btheta_1) \\
	&= \big[x^T(\btheta_r^* - \btheta_1^*) + x^T(\btheta_1^* - \btheta_1 + \btheta_r - \btheta_r^*)\big]\bigg[-\frac{1}{2}x^T(\btheta_r^* - \btheta_1^*) +-\frac{1}{2}x^T(\btheta_1^* - \btheta_1) + \frac{1}{2}x^T(\btheta_r^* - \btheta_r)\bigg] \\
	&= -\frac{1}{2}[x^T(\btheta_r^* - \btheta_1^*)]^2 + \frac{1}{2}\{x^T[(\btheta_1^* - \btheta_1) + (\btheta_r^* - \btheta_r)]\}^2 \\
	&\leq -\frac{3}{8}[x^T(\btheta_r^* - \btheta_1^*)]^2,
\end{align}
and
\begin{equation}
	|(\yk{1}-\bx^T\btheta_1^*)\bx^T(\btheta_r - \btheta_1)| \leq \frac{1}{4}\tau_1\norm{x^T(\btheta_r^* - \btheta_1^*)},
\end{equation}
hence
\begin{align}
	\tE\big[\gamma^{(r)}_{\bw, \btheta}(\bx, y)\big|z = 1,  \mathcal{V}_1 \cap \mathcal{V}_2 \cap \mathcal{V}_3\big] &\leq \frac{w_r}{w_1}\tE\bigg[\exp\bigg\{\frac{1}{4}\tau_1\norm{x^T(\btheta_r^* - \btheta_1^*)} - \frac{3}{8}\norm{x^T(\btheta_r^* - \btheta_1^*)}^2\bigg\}\bigg] \\
	\lesssim \frac{R}{c_w}\exp\bigg\{\frac{1}{4}\tau_1\tau_2\twonorm{\btheta_r^* - \btheta_1^*} - \frac{3}{8}\tau_2\twonorm{\btheta_r^* - \btheta_1^*}^2\bigg\}.
\end{align}
Therefore,
\begin{align}
	\tE\big[\gamma^{(r)}_{\bw, \btheta}(\bx, y)\big|z = 1\big] &\lesssim \frac{R}{c_w}\exp\bigg\{\frac{1}{4}\tau_1\tau_2\twonorm{\btheta_r^* - \btheta_1^*} - \frac{3}{8}\tau_2\twonorm{\btheta_r^* - \btheta_1^*}^2\bigg\} + \tP(\mathcal{V}_1^c)+ \tP(\mathcal{V}_2^c)+ \tP(\mathcal{V}_3^c) \\
	&\lesssim \frac{R}{c_w}\exp\bigg\{\frac{1}{4}\tau_1\tau_2\twonorm{\btheta_r^* - \btheta_1^*} - \frac{3}{8}\tau_2\twonorm{\btheta_r^* - \btheta_1^*}^2\bigg\} + C_b + \exp\{-C\tau_1^2\} + \tau_2.
\end{align}
Let $\tau_1 = c\sqrt{\log \twonorm{\btheta_r^* -\btheta_1^*}}$ and $\tau_2 = 10c\frac{\sqrt{\log \twonorm{\btheta_r^* -\btheta_1^*}}}{\twonorm{\btheta_r^* -\btheta_1^*}}$ with some constant $c > 0$. Then 
\begin{equation}
	\tE\big[\gamma^{(r)}_{\bw, \btheta}(\bx, y)\big|z = 1\big] \lesssim \frac{R}{c_w}\frac{1}{\Delta} + C_b + \frac{1}{\Delta} + \frac{\sqrt{\log \Delta}}{\Delta} \lesssim \frac{R}{c_w}\frac{1}{\Delta}+ C_b+ \frac{\sqrt{\log \Delta}}{\Delta}.
\end{equation}
Similarly, we can show the same bound for $\tE[\gamma^{(r)}_{\bw^*, \btheta^*}(\bx, y)|z = 1]$. Then
\begin{equation}
	\big|\tE\big[\gamma^{(r)}_{\bw, \btheta}(\bx, y) - \gamma^{(r)}_{\bw^*, \btheta^*}(\bx, y)\big| z = 1\big]\big| \lesssim  \frac{R}{c_w}\frac{1}{\Delta}+ C_b+ \frac{\sqrt{\log \Delta}}{\Delta},
\end{equation}
and the same bound holds for $\big|\tE\big[\gamma^{(r)}_{\bw, \btheta}(\bx, y) - \gamma^{(r)}_{\bw^*, \btheta^*}(\bx, y)\big| z = r'\big]\big|$ for any $r' \neq r$. On the other hand,
\begin{equation}
	\big|\tE\big[\gamma^{(r)}_{\bw, \btheta}(\bx, y) - \gamma^{(r)}_{\bw^*, \btheta^*}(\bx, y)\big| z = r\big]\big| \leq \sum_{r' \neq r}\big|\tE\big[\gamma^{(r')}_{\bw, \btheta}(\bx, y) - \gamma^{(r')}_{\bw^*, \btheta^*}(\bx, y)\big| z = r\big]\big|\lesssim  \frac{R^2}{c_w}\frac{1}{\Delta}+ RC_b+ R\frac{\sqrt{\log \Delta}}{\Delta}.
\end{equation}
Therefore,
\begin{align}
	\big|\tE\big[\gamma^{(r)}_{\bw, \btheta}(\bx, y) - \gamma^{(r)}_{\bw^*, \btheta^*}(\bx, y)\big]\big| &\leq \sum_{r'=1}^R w_{r'}^*\big|\tE\big[\gamma^{(r)}_{\bw, \btheta}(\bx, y) - \gamma^{(r)}_{\bw^*, \btheta^*}(\bx, y)\big| z = r'\big]\big| \\
	&\lesssim \frac{R^2}{c_w}\frac{1}{\Delta}+ RC_b+ R\frac{\sqrt{\log \Delta}}{\Delta} \\
	&\lesssim \bigg(\frac{R^2}{c_w}\frac{1}{\Delta}+ RC_b+ R\frac{\sqrt{\log \Delta}}{\Delta}\bigg)\cdot \twonorm{\btheta_{r_0} - \btheta_{r_0}^*},
\end{align}
where the last inequality comes from the fact that $\twonorm{\btheta_{r_0} - \btheta_{r_0}^*} > 1$.

Combining two cases entails Assumption \ref{asmp: w appendix}.(\rom{1}) with $\kappa \asymp \frac{R}{c_w}\frac{\sqrt{\log \Delta}}{\Delta} + \frac{R}{c_w}C_b + \frac{R^2}{c_w}\frac{1}{\Delta}$.

\underline{(\Rom{3}) Part 3:} Deriving the rate of $\gamma$ in Assumption \ref{asmp: theta appendix}.(\rom{1}). 

Similar to Part 2, for notation simplicity, we drop the task index $k$ in the superscript and write $\bwk{k} = \{\wk{k}_r\}_{r=1}^R$, $\bthetak{k} = \{\bthetak{k}_r\}_{r=1}^R$, $\bwks{k} = \{\wks{k}_r\}_{r=1}^R$, $\bthetaks{k} = \{\bthetaks{k}_r\}_{r=1}^R$, $\bxk{k}$, $\yk{k}$, $\Qk{k}$, and $\qk{k}$ simply as $\bwk{k}= \{\wk{k}_r\}_{r=1}^R$, $\btheta= \{\btheta_r\}_{r=1}^R$, $\bw= \{w^*_r\}_{r=1}^R$, $\bthetak{k}= \{\btheta^*_r\}_{r=1}^R$, $\bx$, $y$, $Q$, and $q$.

Note that $\frac{\partial \Qk{k}}{\partial \btheta_r}(\btheta|\bw', \btheta') = -\tE[\gamma^{(r)}_{\bw', \btheta'}(\bx, y)\bx(\bx^T\btheta_r - y)]$, which implies that
\begin{equation}
	\bigg\|\frac{\partial Q}{\partial \btheta_r}(\btheta|\bw, \btheta) - \frac{\partial q}{\partial \btheta_r}(\btheta)\bigg\|_2 = \bigg\|\tE\big[\big(\gamma^{(r)}_{\bw, \btheta}(\bx, y)-\gamma^{(r)}_{\bw^*, \btheta^*}(\bx, y)\big)\bx(\bx^T\btheta_r - y)\big]\bigg\|_2
\end{equation}

(a) Case 1: $\max_{r}\twonorm{\btheta_r - \btheta_r^*} \leq 1$.

WLOG, let us consider $\tE\big[\big(\gamma^{(r)}_{\bw, \btheta}(\bx, y)-\gamma^{(r)}_{\bw^*, \btheta^*}(\bx, y)\big)\bx(\bx^T\btheta_r - y)\big|z = 1\big]$ with $r \neq 1$. For any $\bm{u} \in \mathcal{S}^{d-1}$,
\begin{align}
	&\big|\tE\big[\big(\gamma^{(r)}_{\bw, \btheta}(\bx, y)-\gamma^{(r)}_{\bw^*, \btheta^*}(\bx, y)\big)\bx^T\bm{u}\cdot (\bx^T\btheta_r - y)\big|z = 1\big]\big| \\
	&\leq \underbrace{\big|\tE\big[\big(\gamma^{(r)}_{\bw, \btheta}(\bx, y)-\gamma^{(r)}_{\bw^*, \btheta^*}(\bx, y)\big)\bx^T\bm{u}\cdot \bx^T(\btheta_r^* - \btheta_1^*)\big|z = 1\big]\big|}_{[1]} \\
	&\quad + \underbrace{\big|\tE\big[\big(\gamma^{(r)}_{\bw, \btheta}(\bx, y)-\gamma^{(r)}_{\bw^*, \btheta^*}(\bx, y)\big)\bx^T\bm{u}\cdot \bx^T(\btheta_r - \btheta_r^*)\big|z = 1\big]\big|}_{[2]} \\
	&\quad + \underbrace{\big|\tE\big[\big(\gamma^{(r)}_{\bw, \btheta}(\bx, y)-\gamma^{(r)}_{\bw^*, \btheta^*}(\bx, y)\big)\bx^T\bm{u}\cdot (y - \bx^T\btheta_1^*)\big|z = 1\big]\big|}_{[3]}. \label{eq: case 1 gamma theta}
\end{align}
First,
\begin{align}
	[1] &\leq \sum_{r'=1}^R \bigg|\tE\bigg[\frac{\partial \gamma^{(r)}(\bx, y)}{\partial w_{r'}}(w_{r'} - w_{r'}^*)\bx^T\bm{u}\cdot \bx^T(\btheta_r^* - \btheta_1^*)\bigg| z= 1\bigg]\bigg| \\
	&\quad +\sum_{r'=1}^R \bigg|\tE\bigg[\bigg(\frac{\partial \gamma^{(r)}(\bx, y)}{\partial \btheta_{r'}}\bigg)^T(\btheta_{r'} - \btheta_{r'}^*)\bx^T\bm{u}\cdot \bx^T(\btheta_r^* - \btheta_1^*)\bigg| z= 1\bigg]\bigg|. \label{eq: case 1 gamma theta 2}
\end{align}
Recall that
\begin{align}
	&\frac{\partial \gamma^{(r)}(\bx, y)}{\partial w_r} \\
	&= \frac{\exp\{y\bx^T(\btheta_r - \btheta_1) - \frac{1}{2}[(\bx^T\btheta_r)^2 - (\bx^T\btheta_1)^2]\}(w_1 + \sum_{r' \neq r}w_{r'}\exp\{y\bx^T(\btheta_{r'} - \btheta_1) - \frac{1}{2}[(\bx^T\btheta_{r'})^2 - (\bx^T\btheta_1)^2]\})}{(w_1 + \sum_{r'=1}^Rw_{r'}\exp\{y\bx^T(\btheta_{r'} - \btheta_1) - \frac{1}{2}[(\bx^T\btheta_{r'})^2 - (\bx^T\btheta_1)^2]\})^2}.
\end{align}
Denote $(*) = \yk{1}\bx^T(\btheta_r - \btheta_1) - \frac{1}{2}[(\bx^T\btheta_r)^2 - (\bx^T\btheta_1)^2]$, $\widetilde{z} = (\yk{1} - \bx^T\btheta_1^*)\cdot \bx^T(\btheta_r - \btheta_1)|\bx \sim N(0, [\bx^T(\btheta_r - \btheta_1)]^2)$, where $\yk{1} \overset{d}{=} (y|z = 1)$. Then
\begin{equation}
	(*) = \widetilde{z} + (\bx^T\btheta_1^*)\cdot x^T(\btheta_r - \btheta_1) - \frac{1}{2}x^T(\btheta_r + \btheta_1)\cdot x^T(\btheta_r - \btheta_1),
\end{equation}
where 
\begin{align}
	&(\bx^T\btheta_1^*)\cdot x^T(\btheta_r - \btheta_1) - \frac{1}{2}x^T(\btheta_r + \btheta_1)\cdot x^T(\btheta_r - \btheta_1) \\
	&= \big[x^T(\btheta_r^* - \btheta_1^*) + x^T(\btheta_1^* - \btheta_1 + \btheta_r - \btheta_r^*)\big]\bigg[-\frac{1}{2}x^T(\btheta_r^* - \btheta_1^*) +-\frac{1}{2}x^T(\btheta_1^* - \btheta_1) + \frac{1}{2}x^T(\btheta_r^* - \btheta_r)\bigg] \\
	&= -\frac{1}{2}[x^T(\btheta_r^* - \btheta_1^*)]^2 + \frac{1}{2}\{x^T[(\btheta_1^* - \btheta_1) + (\btheta_r^* - \btheta_r)]\}^2.
\end{align}
Define events
\begin{align}
	\mathcal{V}_1 &= \bigg\{|x^T(\btheta_1^* - \btheta_1)| \vee |x^T(\btheta_r^* - \btheta_r)| \leq \frac{1}{4}|x^T(\btheta_r^* - \btheta_1^*)|\bigg\},\\
	\mathcal{V}_2 &= \{|\widetilde{z}| \leq \tau_1 |x^T(\btheta_r - \btheta_1)|\}, \\
	\mathcal{V}_3 &= \{|\bx^T(\btheta_r^* - \btheta_1^*)| > \tau_2 |\btheta_r^* - \btheta_1^*|\}.
\end{align}
We know that
\begin{align}
	\tP(\mathcal{V}_1^c) &\leq \tP\bigg(|x^T(\btheta_1^* - \btheta_1)| \vee |x^T(\btheta_r^* - \btheta_r)| > \frac{1}{4}|x^T(\btheta_r^* - \btheta_1^*)|\bigg) \\
	&\leq \tP\bigg(|x^T(\btheta_1^* - \btheta_1)| > \frac{1}{4}|x^T(\btheta_r^* - \btheta_1^*)|\bigg) + \tP\bigg(|x^T(\btheta_r^* - \btheta_r)| > \frac{1}{4}|x^T(\btheta_r^* - \btheta_1^*)|\bigg) \\
	&\leq 4\bigg(\frac{\twonorm{\btheta_1^* - \btheta_1}}{\twonorm{\btheta_r^* - \btheta_1^*}} + \frac{\twonorm{\btheta_r^* - \btheta_r}}{\twonorm{\btheta_r^* - \btheta_1^*}}\bigg) \\
	&\leq 8C_b,
\end{align}
where we applied Lemma A.1 in \citet{kwon2020converges} to get the second last inequality. And
\begin{equation}
	\tP(\mathcal{V}_2^c) \leq C\exp\{-C'\tau_1^2\}, \quad \tP(\mathcal{V}_3^c) \leq C\tau_2.
\end{equation}
Given $\mathcal{V}_1 \cap \mathcal{V}_2 \cap \mathcal{V}_3$, we have
\begin{equation}
	-\frac{1}{2}[x^T(\btheta_r^* - \btheta_1^*)]^2 + \frac{1}{2}\{x^T[(\btheta_1^* - \btheta_1) + (\btheta_r^* - \btheta_r)]\}^2 \leq -\frac{3}{8}[x^T(\btheta_r^* - \btheta_1^*)]^2,
\end{equation}
leading to
\begin{equation}
	\widetilde{z} \leq \tau_1|x^T(\btheta_r - \btheta_1)| \leq \tau_1(|x^T(\btheta_r^* - \btheta_1^*)| + |x^T(\btheta_r - \btheta_r^*)| + |x^T(\btheta_1 - \btheta_1^*)|) \leq \tau_1\cdot \frac{3}{2}|x^T(\btheta_r^* - \btheta_1^*)|.
\end{equation}
Hence
\begin{align}
	&\Bigg|\tE\bigg[\frac{\partial \gamma^{(r)}_{\btheta, \bw}(\bx,y)}{\partial w_r}(w_{r} - w_{r}^*)\bx^T\bm{u}\cdot \bx^T(\btheta_r^* - \btheta_1^*)\bigg|z=1\bigg]\Bigg| \\
	&\leq \Bigg|\tE\bigg[\frac{\partial \gamma^{(r)}_{\btheta, \bw}(\bx,y)}{\partial w_r}(w_{r} - w_{r}^*)\bx^T\bm{u}\cdot \bx^T(\btheta_r^* - \btheta_1^*)\bigg|z=1, \mathcal{V}_1 \cap \mathcal{V}_2 \cap \mathcal{V}_3\bigg]\tP(\mathcal{V}_1 \cap \mathcal{V}_2 \cap \mathcal{V}_3)\Bigg| \\
	&\quad + \sum_{j=1}^3\Bigg|\tE\bigg[\frac{\partial \gamma^{(r)}_{\btheta, \bw}(\bx,y)}{\partial w_r}(w_{r} - w_{r}^*)\bx^T\bm{u}\cdot \bx^T(\btheta_r^* - \btheta_1^*)\cdot \mathds{1}(\mathcal{V}_j^c)\bigg|z=1\bigg]\Bigg|.
\end{align}
Note that
\begin{align}
	&\Bigg|\tE\bigg[\frac{\partial \gamma^{(r)}_{\btheta, \bw}(\bx,y)}{\partial w_r}(w_{r} - w_{r}^*)\bx^T\bm{u}\cdot \bx^T(\btheta_r^* - \btheta_1^*)\bigg|z=1, \mathcal{V}_1 \cap \mathcal{V}_2 \cap \mathcal{V}_3\bigg]\Bigg| \\
	&\lesssim \frac{R}{c_w}\exp\bigg\{\frac{3}{2}\tau_1\tau_2\twonorm{\btheta_r^* - \btheta_1^*} - \frac{3}{8}\tau_2^2\twonorm{\btheta_r^* - \btheta_1^*} ^2\bigg\}\cdot \twonorm{\btheta_r^* - \btheta_1^*}\cdot |w_r - w_r^*|.
\end{align}
Also,
\begin{align}
	&\Bigg|\tE\bigg[\frac{\partial \gamma^{(r)}_{\btheta, \bw}(\bx,y)}{\partial w_r}(w_r - w_{r}^*)\bx^T\bm{u}\cdot \bx^T(\btheta_r^* - \btheta_1^*)\cdot \mathds{1}(\mathcal{V}_1^c)\bigg|z=1\bigg]\Bigg| \\
	&\leq \frac{R}{c_w}\sqrt{\tE[(\bx^T\bm{u})^2|\mathcal{V}_1^c]\tP(\mathcal{V}_1^c)}\sqrt{\tE[(\bx^T(\btheta_r^* - \btheta_1^*))^2|\mathcal{V}_1^c]\tP(\mathcal{V}_1^c)}\cdot \norm{w_r - w_{r}^*} \\
	&\lesssim \frac{R}{c_w}\sqrt{\tE[(\bx^T(\btheta_r - \btheta_r^*))^2|\mathcal{V}_1^c] + \tE[(\bx^T(\btheta_1 - \btheta_1^*))^2|\mathcal{V}_1^c]}\cdot \tP(\mathcal{V}_1^c)\cdot \norm{w_r - w_{r}^*} \\
	&\lesssim \frac{R}{c_w}C_b \norm{w_r - w_{r}^*}.
\end{align}
\begin{equation}
	\Bigg|\tE\bigg[\frac{\partial \gamma^{(r)}_{\btheta, \bw}(\bx,y)}{\partial w_r}(w_r - w_{r}^*)\bx^T\bm{u}\cdot \bx^T(\btheta_r^* - \btheta_1^*)\cdot \mathds{1}(\mathcal{V}_2^c)\bigg|z=1\bigg]\Bigg| \lesssim \exp\{-C\tau_1^2\}\cdot \twonorm{\btheta_r^* - \btheta_1^*}\cdot \norm{w_r - w_{r}^*}.
\end{equation}
\begin{align}
	&\Bigg|\tE\bigg[\frac{\partial \gamma^{(r)}_{\btheta, \bw}(\bx,y)}{\partial w_r}(w_r - w_{r}^*)\bx^T\bm{u}\cdot \bx^T(\btheta_r^* - \btheta_1^*)\cdot \mathds{1}(\mathcal{V}_3^c)\bigg|z=1\bigg]\Bigg| \\
	&\leq \frac{R}{c_w}\sqrt{\tE[(\bx^T\bm{u})^2|\mathcal{V}_3^c]\tP(\mathcal{V}_3^c)}\sqrt{\tE[(\bx^T(\btheta_r^* - \btheta_1^*))^2|\mathcal{V}_3^c]\tP(\mathcal{V}_3^c)}\cdot \norm{w_r - w_{r}^*}\\
	&\lesssim \frac{R}{c_w}\tau_2^2 \cdot \twonorm{\btheta_r^* - \btheta_1^*}\cdot \norm{w_r - w_{r}^*}.
\end{align}

Let $\tau_1 = c\sqrt{\log \twonorm{\btheta_r^* - \btheta_1^*}}$ and $\tau_2 = 10c\frac{\sqrt{\log \twonorm{\btheta_r^* - \btheta_1^*}}}{\twonorm{\btheta_r^* - \btheta_1^*}}$ with some constant $c > 0$, then
\begin{align}
	\Bigg|\tE\bigg[\frac{\partial \gamma^{(r)}_{\btheta, \bw}(\bx,y)}{\partial w_r}(w_{r} - w_{r}^*)\bx^T\bm{u}\cdot \bx^T(\btheta_r^* - \btheta_1^*)\bigg|z=1\bigg]\Bigg| &\lesssim \bigg(\frac{R}{c_w}\frac{1}{\Delta} + \frac{R}{c_w}C_b + \frac{1}{\Delta} + \frac{R}{C_w}\frac{\log \Delta}{\Delta}\bigg)\cdot \norm{w_r - w_{r}^*} \\
	&\lesssim \bigg(\frac{R}{c_w}C_b + \frac{R}{C_w}\frac{\log \Delta}{\Delta}\bigg)\cdot \norm{w_r - w_{r}^*}.
\end{align}
Similarly, we can show that
\begin{equation}
	\Bigg|\tE\bigg[\frac{\partial \gamma^{(r)}_{\btheta, \bw}(\bx,y)}{\partial w_{r'}}(w_{r'} - w_{r'}^*)\bx^T\bm{u}\cdot \bx^T(\btheta_{r'}^* - \btheta_1^*)\bigg|z=1\bigg]\Bigg|\lesssim \bigg(\frac{R}{c_w}C_b + \frac{R}{C_w}\frac{\log \Delta}{\Delta}\bigg)\cdot \norm{w_{r'} - w_{r'}^*},
\end{equation}
for $r' \neq r$.

On the other hand, recall that
\begin{equation}
	\frac{\partial \gamma^{(r)}_{\bw,\btheta}(\bx, y)}{\partial \btheta_r} = 
		\gamma^{(r)}_{\bw,\btheta}(\bx, y)\bx(y - \bx^T\btheta_r) - \big(\gamma^{(r)}_{\bw,\btheta}(\bx, y)\big)^2\bx(y - \bx^T\btheta_r), 
\end{equation}
where
\begin{equation}
	\gamma^{(r)}_{\bw, \btheta}(\bx, y) = \frac{w_r\exp\{y\bx^T(\btheta_r - \btheta_1) - \frac{1}{2}[(\bx^T\btheta_r)^2 - (\bx^T\btheta_1)^2]\}}{w_1 + \sum_{r'=2}^Rw_{r'}\exp\{y\bx^T(\btheta_{r'} - \btheta_1) - \frac{1}{2}[(\bx^T\btheta_{r'})^2 - (\bx^T\btheta_1)^2]\}}.
\end{equation}
We have
\begin{align}
	&\big|\tE\big[\gamma^{(r)}_{\bw, \btheta}(\bx, y)\bx^T(\btheta_r - \btheta_r^*)\cdot (\yk{1}-\bx^T\btheta_r)\cdot \bx^T\bm{u}\cdot \bx^T(\btheta_r^* - \btheta_1^*)\big|z = 1\big]\big| \\
	&\leq \big|\tE\big[\gamma^{(r)}_{\bw, \btheta}(\bx, y)\bx^T(\btheta_r - \btheta_r^*)\cdot (\yk{1}-\bx^T\btheta_1^*)\cdot\bx^T\bm{u}\cdot \bx^T(\btheta_r^* - \btheta_1^*)\big|z = 1\big]\big| \\
	&\quad +\big|\tE\big[\gamma^{(r)}_{\bw, \btheta}(\bx, y)\bx^T(\btheta_r - \btheta_r^*)\cdot \bx^T(\btheta_1^* - \btheta_r^*)\cdot\bx^T\bm{u}\cdot \bx^T(\btheta_r^* - \btheta_1^*)\big|z = 1\big]\big| \\
	&\quad + \big|\tE\big[\gamma^{(r)}_{\bw, \btheta}(\bx, y)\bx^T(\btheta_r - \btheta_r^*)\cdot \bx^T(\btheta_r^* - \btheta_r)\cdot\bx^T\bm{u}\cdot \bx^T(\btheta_r^* - \btheta_1^*)\big|z = 1\big]\big|.
\end{align}
Similar to the previous analysis,
\begin{align}
	&\big|\tE\big[\gamma^{(r)}_{\bw, \btheta}(\bx, y)\bx^T(\btheta_r - \btheta_r^*)\cdot (\yk{1}-\bx^T\btheta_1^*)\cdot\bx^T\bm{u}\cdot \bx^T(\btheta_r^* - \btheta_1^*)\big|z = 1\big]\big| \\
	&\leq \big|\tE\big[\gamma^{(r)}_{\bw, \btheta}(\bx, y)\bx^T(\btheta_r - \btheta_r^*)\cdot (\yk{1}-\bx^T\btheta_1^*)\cdot\bx^T\bm{u}\cdot \bx^T(\btheta_r^* - \btheta_1^*)\big|z = 1, \mathcal{V}_1 \cap \mathcal{V}_2 \cap \mathcal{V}_3\big]\big| \\
	&\quad + \sum_{j=1}^3 \big|\tE\big[\gamma^{(r)}_{\bw, \btheta}(\bx, y)\bx^T(\btheta_r - \btheta_r^*)\cdot (\yk{1}-\bx^T\btheta_1^*)\cdot\bx^T\bm{u}\cdot \bx^T(\btheta_r^* - \btheta_1^*)\big|z = 1, \mathcal{V}_j^c\big]\big|\cdot \tP(\mathcal{V}_j^c) \\
	&\lesssim \frac{R}{c_w}\exp\bigg\{\frac{3}{2}\tau_1\tau_2\twonorm{\btheta_r^* - \btheta_1^*} - \frac{3}{8}\tau_2\twonorm{\btheta_r^* - \btheta_1^*}^2\bigg\} \twonorm{\btheta_r^* - \btheta_1^*}\cdot \twonorm{\btheta_r - \btheta_r^*} \\
	&\quad + \frac{R}{c_w}C_b + \exp\{-C\tau_1^2\}\cdot \twonorm{\btheta_r^* - \btheta_1^*}\cdot \twonorm{\btheta_r - \btheta_r^*} + \frac{R}{c_w}\tau_2^2\twonorm{\btheta_r^* - \btheta_1^*}\cdot \twonorm{\btheta_r - \btheta_r^*}.
\end{align}
Let $\tau_1 = c\sqrt{\log \twonorm{\btheta_r^* - \btheta_1^*}}$ and $\tau_2 = 10c\frac{\sqrt{\log \twonorm{\btheta_r^* - \btheta_1^*}}}{\twonorm{\btheta_r^* - \btheta_1^*}}$ with some constant $c > 0$, then
\begin{equation}
	\big|\tE\big[\gamma^{(r)}_{\bw, \btheta}(\bx, y)\bx^T(\btheta_r - \btheta_r^*)\cdot (\yk{1}-\bx^T\btheta_1^*)\cdot\bx^T\bm{u}\cdot \bx^T(\btheta_r^* - \btheta_1^*)\big|z = 1\big]\big| \lesssim \bigg(\frac{R}{c_w}\frac{\log \Delta}{\Delta} + \frac{R}{c_w}C_b\bigg)\cdot \twonorm{\btheta_r - \btheta_r^*}.
\end{equation}
Similarly,
\begin{align}
	\big|\tE\big[\gamma^{(r)}_{\bw, \btheta}(\bx, y)\bx^T(\btheta_r - \btheta_r^*)\cdot \bx^T(\btheta_1^* - \btheta_r^*)\cdot\bx^T\bm{u}\cdot \bx^T(\btheta_r^* - \btheta_1^*)\big|z = 1\big]\big| &\lesssim \bigg[\frac{R}{c_w}\frac{(\log \Delta)^{3/2}}{\Delta} + \frac{R}{c_w}C_b\bigg]\cdot \twonorm{\btheta_r - \btheta_r^*},\\
	\big|\tE\big[\gamma^{(r)}_{\bw, \btheta}(\bx, y)\bx^T(\btheta_r - \btheta_r^*)\cdot \bx^T(\btheta_r^* - \btheta_r)\cdot\bx^T\bm{u}\cdot \bx^T(\btheta_r^* - \btheta_1^*)\big|z = 1\big]\big| &\lesssim \bigg(\frac{R}{c_w}\frac{\log \Delta}{\Delta} + \frac{R}{c_w}C_b\bigg)\cdot \twonorm{\btheta_r - \btheta_r^*}.
\end{align}
Hence,
\begin{equation}
	\big|\tE\big[\gamma^{(r)}_{\bw, \btheta}(\bx, y)\bx^T(\btheta_r - \btheta_r^*)\cdot (\yk{1}-\bx^T\btheta_r)\cdot \bx^T\bm{u}\cdot \bx^T(\btheta_r^* - \btheta_1^*)\big|z = 1\big]\big|\lesssim \bigg[\frac{R}{c_w}\frac{(\log \Delta)^{3/2}}{\Delta} + \frac{R}{c_w}C_b\bigg]\cdot \twonorm{\btheta_r - \btheta_r^*}.
\end{equation}
Similarly,
\begin{equation}
	\big|\tE\big[(\gamma^{(r)}_{\bw, \btheta}(\bx, y))^2\bx^T(\btheta_r - \btheta_r^*)\cdot (\yk{1}-\bx^T\btheta_r)\cdot \bx^T\bm{u}\cdot \bx^T(\btheta_r^* - \btheta_1^*)\big|z = 1\big]\big|\lesssim \bigg[\frac{R}{c_w}\frac{(\log \Delta)^{3/2}}{\Delta} + \frac{R}{c_w}C_b\bigg]\cdot \twonorm{\btheta_r - \btheta_r^*}.
\end{equation}
Therefore,
\begin{equation}
	\bigg|\tE\bigg[\bigg(\frac{\partial \gamma^{(r)}_{\bw, \btheta}(\bx, y)}{\partial \btheta_r}\bigg)^T(\btheta_r - \btheta_r^*) \cdot \bx^T\bm{u}\cdot \bx^T(\btheta_r^* - \btheta_1^*)\big|z = 1\bigg]\bigg|\lesssim \bigg[\frac{R}{c_w}\frac{(\log \Delta)^{3/2}}{\Delta} + \frac{R}{c_w}C_b\bigg]\cdot \twonorm{\btheta_r - \btheta_r^*}.
\end{equation}
Similarly,
\begin{equation}
	\bigg|\tE\bigg[\bigg(\frac{\partial \gamma^{(r)}_{\bw, \btheta}(\bx, y)}{\partial \btheta_{r'}}\bigg)^T(\btheta_r - \btheta_r^*) \cdot \bx^T\bm{u}\cdot \bx^T(\btheta_r^* - \btheta_1^*)\big|z = 1\bigg]\bigg|\lesssim \bigg[\frac{R}{c_w}\frac{(\log \Delta)^{3/2}}{\Delta} + \frac{R}{c_w}C_b\bigg]\cdot \twonorm{\btheta_r - \btheta_r^*},
\end{equation}
for $r' \neq r$. Recall \eqref{eq: case 1 gamma theta} and \eqref{eq: case 1 gamma theta 2},
\begin{align}
	[1] &\leq \sum_{r'=1}^R \bigg|\tE\bigg[\frac{\partial \gamma^{(r)}(\bx, y)}{\partial w_{r'}}(w_{r'} - w_{r'}^*)\bx^T\bm{u}\cdot \bx^T(\btheta_r^* - \btheta_1^*)\bigg| z= 1\bigg]\bigg| \\
	&\quad +\sum_{r'=1}^R \bigg|\tE\bigg[\bigg(\frac{\partial \gamma^{(r)}(\bx, y)}{\partial \btheta_{r'}}\bigg)^T(\btheta_{r'} - \btheta_{r'}^*)\bx^T\bm{u}\cdot \bx^T(\btheta_r^* - \btheta_1^*)\bigg| z= 1\bigg]\bigg| \\
	&\lesssim \bigg[\frac{R}{c_w}\frac{(\log \Delta)^{3/2}}{\Delta} + \frac{R}{c_w}C_b\bigg] \cdot \sum_{r'=1}^R (\norm{w_{r'}-w_{r'}^*} + \twonorm{\btheta_{r'} - \btheta_{r'}^*}).
\end{align}
Similarly, the same bound holds for terms [2] and [3] in \eqref{eq: case 1 gamma theta} as well, therefore
\begin{equation}
	\big|\tE\big[\big(\gamma^{(r)}_{\bw, \btheta}(\bx, y)-\gamma^{(r)}_{\bw^*, \btheta^*}(\bx, y)\big)\bx^T\bm{u}\cdot (\bx^T\btheta_r - y)\big|z = 1\big]\big| \lesssim \bigg[\frac{R}{c_w}\frac{(\log \Delta)^{3/2}}{\Delta} + \frac{R}{c_w}C_b\bigg] \cdot \sum_{r'=1}^R (\norm{w_{r'}-w_{r'}^*} + \twonorm{\btheta_{r'} - \btheta_{r'}^*}).
\end{equation}
With similar arguments, we have
\begin{equation}
	\big|\tE\big[\big(\gamma^{(r)}_{\bw, \btheta}(\bx, y)-\gamma^{(r)}_{\bw^*, \btheta^*}(\bx, y)\big)\bx^T\bm{u}\cdot (\bx^T\btheta_r - y)\big|z = \tilde{r}\big]\big| \lesssim  \bigg[\frac{R}{c_w}\frac{(\log \Delta)^{3/2}}{\Delta} + \frac{R}{c_w}C_b\bigg] \cdot \sum_{r'=1}^R (\norm{w_{r'}-w_{r'}^*} + \twonorm{\btheta_{r'} - \btheta_{r'}^*}),
\end{equation}
for $\tilde{r} \neq 1$. Therefore,
\begin{align}
	\bigg\|\frac{\partial Q}{\partial \btheta_r}(\btheta|\bw, \btheta) - \frac{\partial q}{\partial \btheta_r}(\btheta)\bigg\|_2 &= \bigg\|\tE\big[\big(\gamma^{(r)}_{\bw, \btheta}(\bx, y)-\gamma^{(r)}_{\bw^*, \btheta^*}(\bx, y)\big)\bx(\bx^T\btheta_r - y)\big]\bigg\|_2 \\
	&= \sup_{\twonorm{\bm{u}}\leq 1}\sum_{r'=1}^R \tE\big[\big(\gamma^{(r)}_{\bw, \btheta}(\bx, y)-\gamma^{(r)}_{\bw^*, \btheta^*}(\bx, y)\big)(\bx^T\bm{u})(\bx^T\btheta_r - y)\big|z = r'\big]\cdot \tP(z = r') \\
	&\leq \bigg[C\frac{R}{c_w}\frac{(\log \Delta)^{3/2}}{\Delta} + C\frac{R}{c_w}C_b\bigg] \cdot \sum_{r'=1}^R (\norm{w_{r'}-w_{r'}^*} + \twonorm{\btheta_{r'} - \btheta_{r'}^*}).
\end{align}

(b) Case 2: $\max_{r}\twonorm{\btheta_r - \btheta_r^*} > 1$. 

Similar to case 2 of Part 2, it can be shown that
\begin{equation}
	\bigg\|\frac{\partial Q}{\partial \btheta_r}(\btheta|\bw, \btheta) - \frac{\partial q}{\partial \btheta_r}(\btheta)\bigg\|_2 \leq \bigg[C\frac{R^2}{c_w} + C\frac{R}{c_w}\frac{(\log \Delta)^{3/2}}{\Delta} + C\frac{R}{c_w}C_b\bigg] \cdot \sum_{r'=1}^R (\norm{w_{r'}-w_{r'}^*} + \twonorm{\btheta_{r'} - \btheta_{r'}^*}).
\end{equation}
Therefore $\gamma \asymp \frac{R^2}{c_w} + \frac{R}{c_w}\frac{(\log \Delta)^{3/2}}{\Delta} + \frac{R}{c_w}C_b$ in Assumption \ref{asmp: theta appendix}.(\rom{1}).

\underline{(\Rom{4}) Part 4:} Deriving the rate of $\mathcal{W}$ in Assumption \ref{asmp: w appendix}.(\rom{2}). 
Let 
\begin{align}
	V &= \sup_{\substack{|w_r - \wks{k}_r| \leq \frac{c_w}{2R} \\ \twonorm{\btheta_r - \bthetaks{k}_r} \leq \xi}}\bigg|\frac{1}{n}\sum_{i=1}^n\tP(\zk{k}=r|\bxk{k}_i, \yk{k}_i;\bw, \btheta) - \tE_{\bxk{k}}\big[\tP(\zk{k}=r|\bxk{k}, \yk{k};\bw, \btheta)\big]\bigg| \\
	&= \sup_{\substack{|w_r - \wks{k}_r| \leq \frac{c_w}{2R} \\ \twonorm{\btheta_r - \bthetaks{k}_r} \leq \xi}}\bigg|\frac{1}{n}\sum_{i=1}^n\gamma^{(r)}_{\btheta, \bw}(\bxk{k}_i, \yk{k}_i) - \tE\big[\gamma^{(r)}_{\btheta, \bw}(\bxk{k}, \yk{k})\big]\bigg|.
\end{align}
By bounded difference inequality (Corollary 2.21 in \citet{wainwright2019high}), w.p. at least $1-\delta$,
\begin{equation}
	V \leq \tE V + \sqrt{\frac{\log(1/\delta)}{n}}.
\end{equation}
And by classical symmetrization arguments (e.g., see Proposition 4.11 in \citet{wainwright2019high}),
\begin{equation}
	\tE V \leq \frac{2}{n}\tE_{\bxk{k}}\tE_{\bm{\epsilon}}\sup_{\substack{|w_r - \wks{k}_r| \leq \frac{c_w}{2R} \\ \twonorm{\btheta_r - \bthetaks{k}_r} \leq \xi}} \bigg|\sum_{i=1}^n \epsilonk{k}_i\gamma^{(k)}_{\btheta, \bw}(\bxk{k}_i, \yk{k}_i)\bigg|.
\end{equation}
Let $g_{ir}^{(k)} = (\btheta_r - \btheta_1)^T\bxk{k}_i\cdot \yk{k}_i - \frac{1}{2}[((\bxk{k}_i)^T\btheta_r)^2 - ((\bxk{k}_i)^T\btheta_1)^2] + \log w_r - \log w_1$, $\varphi(\bx) = \frac{\exp\{x_r\}}{1+\sum_{r=2}^R \exp\{x_r\}}$, where $\varphi$ is $1$-Lipschitz (w.r.t. $\ell_2$-norm) and $\gamma^{(r)}_{\btheta, \bw}(\bx, y) = \varphi(\{g^{(k)}_{ir}\}_{r=2}^R)$. Then by Lemma \ref{lem: vec contraction},
\begin{align}
	&\frac{2}{n}\tE_{\bxk{k}}\tE_{\bm{\epsilon}}\sup_{\substack{|w_r - \wks{k}_r| \leq \frac{c_w}{2R} \\ \twonorm{\btheta_r - \bthetaks{k}_r} \leq \xi}} \bigg|\sum_{i=1}^n \epsilonk{k}_i\gamma^{(k)}_{\btheta, \bw}(\bxk{k}_i, \yk{k}_i))\bigg| \\
	&\lesssim \frac{1}{n}\tE_{\bxk{k}}\tE_{\bm{\epsilon}}\sup_{\substack{|w_r - \wks{k}_r| \leq \frac{c_w}{2R} \\ \twonorm{\btheta_r - \bthetaks{k}_r} \leq \xi}} \bigg|\sum_{i=1}^n \sum_{r=2}^R \epsilonk{k}_{ir}g^{(k)}_{ir}\bigg| \\
	&\lesssim \frac{1}{n}\sum_{r=2}^R \tE_{\bxk{k}}\tE_{\bm{\epsilon}}\sup_{\substack{|w_r - \wks{k}_r| \leq \frac{c_w}{2R} \\ \twonorm{\btheta_r - \bthetaks{k}_r} \leq \xi}} \bigg|\sum_{i=1}^n \epsilonk{k}_{ir}g^{(k)}_{ir}\bigg| \\
	&\lesssim \sum_{r=2}^R\Bigg\{\frac{1}{n}\tE_{\bxk{k}}\tE_{\bm{\epsilon}} \sup_{\substack{|w_r - \wks{k}_r| \leq \frac{c_w}{2R} \\ \twonorm{\btheta_r - \bthetaks{k}_r} \leq \xi}} \bigg|\sum_{i=1}^n \epsilonk{k}_{ir}(\btheta_r - \btheta_1)^T\bxk{k}_i\cdot \yk{k}_i\bigg| \\
	&\quad\quad\quad + \frac{1}{n}\tE_{\bxk{k}}\tE_{\bm{\epsilon}} \sup_{\substack{|w_r - \wks{k}_r| \leq \frac{c_w}{2R} \\ \twonorm{\btheta_r - \bthetaks{k}_r} \leq \xi}} \bigg|\sum_{i=1}^n \epsilonk{k}_{ir}[((\bxk{k}_i)^T\btheta_r)^2 - ((\bxk{k}_i)^T\btheta_1)^2]\bigg| \\
	&\quad\quad\quad +  \frac{1}{n}\tE_{\bxk{k}}\tE_{\bm{\epsilon}} \sup_{\substack{|w_r - \wks{k}_r| \leq \frac{c_w}{2R} \\ \twonorm{\btheta_r - \bthetaks{k}_r} \leq \xi}} \bigg|\sum_{i=1}^n \epsilonk{k}_{ir}(\log w_r - \log w_1)\bigg|\Bigg\}\\
	&\lesssim \sum_{r=2}^R\Bigg\{\frac{1}{n}\tE_{\bxk{k}}\tE_{\bm{\epsilon}} \sup_{\substack{|w_r - \wks{k}_r| \leq \frac{c_w}{2R} \\ \twonorm{\btheta_r - \bthetaks{k}_r} \leq \xi}} \bigg|\sum_{i=1}^n \epsilonk{k}_{ir}(\btheta_r - \bthetaks{k}_r)^T\bxk{k}_i\cdot \yk{k}_i\bigg| \\
	&\quad\quad\quad + \frac{1}{n}\tE_{\bxk{k}}\tE_{\bm{\epsilon}} \sup_{\substack{|w_r - \wks{k}_r| \leq \frac{c_w}{2R} \\ \twonorm{\btheta_r - \bthetaks{k}_r} \leq \xi}} \bigg|\sum_{i=1}^n \epsilonk{k}_{ir}(\btheta_1 - \bthetaks{k}_1)^T\bxk{k}_i\cdot \yk{k}_i\bigg| \\
	&\quad\quad\quad + \frac{1}{n}\tE_{\bxk{k}}\tE_{\bm{\epsilon}} \bigg|\sum_{i=1}^n \epsilonk{k}_{ir}(\bthetaks{k}_r - \bthetaks{k}_1)^T\bxk{k}_i\cdot \yk{k}_i\bigg|  \\
	&\quad\quad\quad + \frac{1}{n}\tE_{\bxk{k}}\tE_{\bm{\epsilon}} \sup_{\substack{|w_r - \wks{k}_r| \leq \frac{c_w}{2R} \\ \twonorm{\btheta_r - \bthetaks{k}_r} \leq \xi}} \bigg|\sum_{i=1}^n \epsilonk{k}_{ir}(\btheta_r + \btheta_1)^T\bxk{k}_i\cdot (\btheta_r - \btheta_1)^T\bxk{k}_i\bigg|\\
	&\quad\quad\quad  +  \frac{1}{n}\tE_{\bxk{k}}\tE_{\bm{\epsilon}} \sup_{\substack{|w_r - \wks{k}_r| \leq \frac{c_w}{2R} \\ \twonorm{\btheta_r - \bthetaks{k}_r} \leq \xi}} \bigg|\sum_{i=1}^n \epsilonk{k}_{ir}(\log w_r - \log w_1)\bigg|\Bigg\}\\
	&\lesssim RM\xi\sqrt{\frac{d}{n}} + [RM^2+R\log(Rc_w^{-1})]\sqrt{\frac{1}{n}}, 
\end{align}
which implies 
\begin{equation}
	V \lesssim RM\xi\sqrt{\frac{d}{n}} + [RM^2+R\log(Rc_w^{-1})]\sqrt{\frac{1}{n}} + \sqrt{\frac{\log(1/\delta)}{n}} \asymp \mathcal{W}(n, \delta, \xi).
\end{equation}
w.p. at least $1-\delta$.

\underline{(\Rom{5}) Part 5:} Deriving the rate of $\mathcal{E}_1$ in Assumption \ref{asmp: theta appendix}.(\rom{2}).

We first introduce the following useful lemma.
\begin{lemma}[Theorem 4 in \citet{maurer2021concentration}]\label{lem: maurer concentration subexp}
	Let $f: \mathcal{X}^n \rightarrow \mathbb{R}$ and $X = (X_1, \ldots, X_n)$ be a vector of independent random variables with values in a space $\mathcal{X}$. Then for any $t > 0$ we have
	\begin{equation}
		\tP(f(X) - \tE f(X) > t) \leq \exp\left\{-\frac{t^2}{4e^2\infnorma{\sum_{i=1}^n \|f_i(X)\|_{\psi_1}^2} + 2e\max_i \infnorma{\|f_i(X)\|_{\psi_1}}t}\right\},
	\end{equation}
	where $f_i(X)$ as a random function of $x$ is defined to be $(f_i(X))(x) \coloneqq f(x_1, \ldots, x_{i-1}, X_{i}, x_{i+1}, \ldots, X_n) - \tE_{X_i}[f(x_1, \ldots, x_{i-1}, X_{i}, x_{i+1}, \ldots, X_n)]$, the sub-Gaussian norm $\|Z\|_{\psi_1} \coloneqq \sup_{d\geq 1}\{\|Z\|_d/d\}$, and $\|Z\|_d = (\tE |Z|^d)^{1/d}$.
\end{lemma}

Let
\begin{align}
	U &= \sup_{\substack{|w_r - \wks{k}_r| \leq \frac{c_w}{2R} \\ \twonorm{\btheta_r - \bthetaks{k}_r} \leq r_{\btheta}^*}}\bigg\|\frac{1}{n}\sum_{i=1}^n\gamma^{(r)}_{\btheta, \bw}(\bxk{k}_i)\bxk{k}_i\yk{k}_i - \tE\big[\gamma^{(r)}_{\btheta, \bw}(\bxk{k})\bxk{k}\yk{k}\big]\bigg\|_2 \\
	&= \sup_{\twonorm{\bm{u}}\leq 1} \sup_{\substack{|w_r - \wks{k}_r| \leq \frac{c_w}{2R} \\ \twonorm{\btheta_r - \bthetaks{k}_r} \leq r_{\btheta}^*}}\bigg|\frac{1}{n}\sum_{i=1}^n\gamma^{(r)}_{\btheta, \bw}(\bxk{k}_i)(\bxk{k}_i)^T\bm{u}\cdot \yk{k}_i - \tE\big[\gamma^{(r)}_{\btheta, \bw}(\bxk{k})(\bxk{k})^T\bm{u}\cdot \yk{k}\big]\bigg| \\
	&\leq 2\max_{j=1:N}\underbrace{\sup_{\substack{|w_r - \wks{k}_r| \leq \frac{c_w}{2R} \\ \twonorm{\btheta_r - \bthetaks{k}_r} \leq r_{\btheta}^*}}\bigg|\frac{1}{n}\sum_{i=1}^n\gamma^{(r)}_{\btheta, \bw}(\bxk{k}_i)(\bxk{k}_i)^T\bm{u}_j\cdot \yk{k}_i - \tE\big[\gamma^{(r)}_{\btheta, \bw}(\bxk{k})(\bxk{k})^T\bm{u}_j\cdot \yk{k}_i\big]\bigg|}_{U_j},
\end{align}
where $\{\bm{u}_j\}_{j=1}^N$ is a $1/2$-cover of the unit ball $\mathcal{B}(\bm{0}, 1)$ in $\mathbb{R}^d$ w.r.t. $\ell_2$-norm, with $N \leq 5^d$ (by Example 5.8 in \citet{wainwright2019high}). We first bound $U_j - \tE U_j$ as below. Fix $(\bxk{k}_1, \yk{k}_i), \ldots, (\bxk{k}_{i-1}, \yk{k}_{i-1}), (\bxk{k}_{i+1}, \yk{k}_{i+1}), \ldots, (\bxk{k}_n, \yk{k}_n)$ and define $s_{ir}^{(k)}(\bxk{k}_{i}, \yk{k}_{i}) = V_j - \tE\big[V_j|(\bxk{k}_1, \yk{k}_i), \ldots, (\bxk{k}_{i-1}, \yk{k}_{i-1}), (\bxk{k}_{i+1}, \yk{k}_{i+1}), \ldots, (\bxk{k}_n, \yk{k}_n)\big]$. Then
\begin{align}
	|s_{ir}^{(k)}(\bxk{k}_i, \yk{k}_i)| &\leq \underbrace{\frac{1}{n}\sup_{\substack{|w_r - \wks{k}_r| \leq \frac{c_w}{2R} \\ \twonorm{\btheta_r - \bthetaks{k}_r} \leq r_{\btheta}^*}} \Big|\gamma^{(r)}_{\btheta, \bw}(\bxk{k}_i, \yk{k}_i)(\bxk{k}_i)^T\bm{u}_j\cdot \yk{k}_i\Big|}_{W_1}\\
	&\quad + \underbrace{\frac{2}{n}\tE\Bigg|\sup_{\substack{|w_r - \wks{k}_r| \leq \frac{c_w}{2R} \\ \twonorm{\btheta_r - \bthetaks{k}_r} \leq r_{\btheta}^*}} \gamma^{(r)}_{\btheta, \bw}(\bxk{k}, \yk{k})(\bxk{k})^T\bm{u}_j\cdot \yk{k}\Bigg|}_{W_2},
\end{align}
where $[\tE(W_1 + W_2)^d]^{1/d} \leq (\tE W_1^d)^{1/d} + (\tE W_2^d)^{1/d}$, and $(\tE W_1^d)^{1/d}, (\tE W_2^d)^{1/d} \leq CMd/n$ with some constant$C > 0$. Then by Lemma \ref{lem: maurer concentration},
\begin{equation}
	\tP(U_j - \tE U_j \geq t) \lesssim \exp\bigg\{-\frac{Cnt^2}{M^2}\bigg\}.
\end{equation}
By a similar procedure used in deriving $\mathcal{W}(n, \delta, \xi)$, we can show that
\begin{align}
	\tE U_j &\lesssim \frac{1}{n}\tE_{\bxk{k}}\tE_{\bm{\epsilon}}\sup_{\substack{|w_r - \wks{k}_r| \leq \frac{c_w}{2R} \\ \twonorm{\btheta_r - \bthetaks{k}_r} \leq r_{\btheta}^*}} \bigg|\sum_{i=1}^n  \gamma^{(r)}_{\btheta,\bw}(\bxk{k}_i, \yk{k}_i)(\bxk{k}_i)^T\bm{u}_j\cdot \yk{k}_i\cdot \epsilonk{k}_i \bigg| \\
	&\lesssim \frac{1}{n}\tE_{\bxk{k}}\tE_{\bm{\epsilon}} \sup_{\substack{|w_r - \wks{k}_r| \leq \frac{c_w}{2R} \\ \twonorm{\btheta_r - \bthetaks{k}_r} \leq r_{\btheta}^*}} \bigg|\sum_{i=1}^n (\bxk{k}_i)^T\bm{u}_j\cdot \yk{k}_i\cdot \epsilonk{k}_{i1}\bigg|\\
	&\quad+ \sum_{r=2}^R\Bigg\{\frac{1}{n}\tE_{\bxk{k}}\tE_{\bm{\epsilon}} \sup_{\substack{|w_r - \wks{k}_r| \leq \frac{c_w}{2R} \\ \twonorm{\btheta_r - \bthetaks{k}_r} \leq r_{\btheta}^*}} \bigg|\sum_{i=1}^n \epsilonk{k}_{ir}(\btheta_r - \btheta_1)^T\bxk{k}_i\cdot \yk{k}_i\bigg| \\
	&\quad\quad\quad + \frac{1}{n}\tE_{\bxk{k}}\tE_{\bm{\epsilon}} \sup_{\substack{|w_r - \wks{k}_r| \leq \frac{c_w}{2R} \\ \twonorm{\btheta_r - \bthetaks{k}_r} \leq r_{\btheta}^*}} \bigg|\sum_{i=1}^n \epsilonk{k}_{ir}\yk{k}_i\big[(\btheta_r^T\bxk{k}_i)^2 - (\btheta_1^T\bxk{k}_i)^2\big]\bigg| \\
	&\quad\quad\quad +  \frac{1}{n}\tE_{\bxk{k}}\tE_{\bm{\epsilon}} \sup_{\substack{|w_r - \wks{k}_r| \leq \frac{c_w}{2R} \\ \twonorm{\btheta_r - \bthetaks{k}_r} \leq r_{\btheta}^*}} \bigg|\sum_{i=1}^n \epsilonk{k}_{ir}\yk{k}_i(\log w_r - \log w_1)\bigg|\Bigg\}\\
	&\lesssim [RM^2r_{\btheta}^* + RM(M+r_{\btheta}^*)^2]\sqrt{\frac{d}{n}} + MR\log\Big(\frac{R}{c_w}\Big)\sqrt{\frac{1}{n}}\\
	&\lesssim RM^3\sqrt{\frac{d}{n}} + MR\log\Big(\frac{R}{c_w}\Big)\sqrt{\frac{1}{n}},
\end{align}
which implies that
\begin{equation}
	\tP\bigg(U_j \geq RM^3\sqrt{\frac{d}{n}} + MR\log\Big(\frac{R}{c_w}\Big)\sqrt{\frac{1}{n}} + t\bigg) \lesssim \exp\bigg\{-\frac{Cn^2t^2}{nM^2 + Mtn}\bigg\} = \exp\bigg\{-\frac{Cnt^2}{M^2 + Mt}\bigg\}.
\end{equation}
Therefore
\begin{equation}
	\tP\bigg(\max_{j=1:N}U_j \geq CRM^3\sqrt{\frac{d}{n}} + CMR\log\Big(\frac{R}{c_w}\Big)\sqrt{\frac{1}{n}} + t\bigg) \lesssim N\exp\bigg\{-\frac{Cnt^2}{M^2+M^2t}\bigg\},
\end{equation}
which implies that
\begin{equation}
	U \lesssim RM^3\sqrt{\frac{d}{n}} + MR\log (Rc_w^{-1})\sqrt{\frac{1}{n}} + M\sqrt{\frac{\log(1/\delta)}{n}},
\end{equation}
w.p. at least $1-\delta$. On the other hand, similarly, we have
\begin{align}
	&\sup_{\substack{|w_r - \wks{k}_r| \leq \frac{c_w}{2R} \\ \twonorm{\btheta_r - \bthetaks{k}_r} \leq r_{\btheta}^*}}\bigg\|\frac{1}{n}\sum_{i=1}^n\gamma^{(r)}_{\btheta, \bw}(\bxk{k}_i) \bxk{k}_i(\bxk{k}_i)^T\btheta_r - \tE\big[\gamma^{(r)}_{\btheta, \bw}(\bxk{k})\bxk{k}(\bxk{k})^T\btheta_r\big]\bigg\|_2 \\
	&\lesssim RM^3\sqrt{\frac{d}{n}} + MR\log (Rc_w^{-1})\sqrt{\frac{1}{n}} + M\sqrt{\frac{\log(1/\delta)}{n}},
\end{align}
hence
\begin{equation}
	\mathcal{E}_1(n, \delta) \asymp RM^3\sqrt{\frac{d}{n}} + MR\log (Rc_w^{-1})\sqrt{\frac{1}{n}} + M\sqrt{\frac{\log(1/\delta)}{n}}.
\end{equation}

\underline{(\Rom{6}) Part 6:} Deriving the rate of $\mathcal{E}_2$ in Assumption \ref{asmp: theta appendix}.(\rom{3}).
Let 
\begin{align}
	Z &= \sup_{\substack{\norm{w_r - \wks{k}_r} \leq \frac{c_w}{2R} \\ \twonorm{\btheta_r - \bthetaks{k}_r} \leq r_{\btheta}^* \\ 0 <  \etak{k}_r \leq \bar{\eta}}} \frac{1}{n|S|}\Bigg\|\sum_{k \in S}\etak{k}_r\cdot \sum_{i=1}^n\big(\gamma^{(k)}_{\btheta, \bw}(\bxk{k}_i)\bxk{k}_i\yk{k}_i - \tE[\gamma^{(k)}_{\btheta, \bw}(\bxk{k})\bxk{k}\yk{k}]\big)\Bigg\|_2 \\
	&= \sup_{\twonorm{\bm{u}}\leq 1}\sup_{\substack{\norm{w_r - \wks{k}_r} \leq \frac{c_w}{2R} \\ \twonorm{\btheta_r - \bthetaks{k}_r} \leq r_{\btheta}^* \\ 0 <  \etak{k}_r \leq \bar{\eta}}} \frac{1}{n|S|}\Bigg|\sum_{k \in S}\etak{k}_r\cdot \sum_{i=1}^n\big(\gamma^{(k)}_{\btheta, \bw}(\bxk{k}_i, \yk{k}_i)(\bxk{k}_i)^T\bm{u}\cdot \yk{k}_i - \tE[\gamma^{(k)}_{\btheta, \bw}(\bxk{k})(\bxk{k})^T\bm{u}\cdot \yk{k}]\big)\Bigg| \\
	&\leq \sup_{j'_1,\ldots, j_k'=1:N'}\sup_{j=1:N}\\
	&\quad\quad\underbrace{\frac{2}{n|S|}\sup_{\substack{\norm{w_r - \wks{k}_r} \leq \frac{c_w}{2R} \\ \twonorm{\btheta_r - \bthetaks{k}_r} \leq r_{\btheta}^*}}\Bigg|\sum_{k \in S}\eta_{j'_k}\cdot \sum_{i=1}^n\big(\gamma^{(k)}_{\btheta, \bw}(\bxk{k}_i, \yk{k}_i)(\bxk{k}_i)^T\bm{u}_j\cdot \yk{k}_i - \tE[\gamma^{(k)}_{\btheta, \bw}(\bxk{k}, \yk{k})(\bxk{k})^T\bm{u}_j\cdot \yk{k}]\big)\Bigg|}_{Z(j,j'_1,\ldots, j_k')},
\end{align}
where $\{\bm{u}_j\}_{j=1}^N$ is a $1/2$-cover of the unit ball $\mathcal{B}(\bm{0}, 1)$ in $\mathbb{R}^d$ w.r.t. $\ell_2$-norm with $N \leq 5^d$ and $\{\eta_{j'}\}_{j'=1}^{N'}$ is a $1/2$-cover of $[0,1]$ with $N' \leq 2$. We first bound $Z(j,j'_1,\ldots, j_k') - \tE Z(j,j'_1,\ldots, j_k')$ as follows. Fix $(\bxk{k}_1, \yk{k}_i), \ldots, (\bxk{k}_{i-1}, \yk{k}_{i-1}), (\bxk{k}_{i+1}, \yk{k}_{i+1}), \ldots, (\bxk{k}_n, \yk{k}_n)$ and define $v_{ir}^{(k)}(\bxk{k}_i, \yk{k}_i) = Z(j,j'_1,\ldots, j_k') - \tE[Z(j,j'_1,\ldots, j_k')|\{(\bxk{k}_i, \yk{k}_i)\}_{k \in S, i \in [n]}\backslash \{(\bxk{k}_i, \yk{k}_i)\}]$. Then
\begin{align}
	|v_{ir}^{(k)}(\bxk{k}_i)| &\leq \underbrace{\frac{\eta_{j_k}}{n|S|}\sup_{\substack{|w_r - \wks{k}_r| \leq \frac{c_w}{2R} \\ \twonorm{\btheta_r - \bthetaks{k}_r} \leq r_{\btheta}^*}} \Big|\gamma^{(r)}_{\btheta, \bw}(\bxk{k}_i, \yk{k}_i)(\bxk{k}_i)^T\bm{u}_j\cdot \yk{k}_i\Big|}_{W_1}  \\
	&\quad + \underbrace{\frac{2\eta_{j_k}}{n|S|}\tE\Bigg|\sup_{\substack{|w_r - \wks{k}_r| \leq \frac{c_w}{2R} \\ \twonorm{\btheta_r - \bthetaks{k}_r} \leq r_{\btheta}^*}} \gamma^{(r)}_{\btheta, \bw}(\bxk{k},\yk{k})(\bxk{k})^T\bm{u}_j\cdot \yk{k}\Bigg|}_{W_2}.
\end{align}
Via the same procedure used to bound $U_j$, it can be shown that
\begin{align}
	\tP(Z(j,j'_1,\ldots, j_k') - \tE Z(j,j'_1,\ldots, j_k') \geq t) \lesssim \exp\bigg\{-\frac{C(n|S|)^2t^2}{n|S|M^2\bar{\eta}^2 +\bar{\eta}Mtn|S|}\bigg\} = \exp\bigg\{-\frac{Cn|S|t^2}{M^2\bar{\eta}^2 +\bar{\eta}Mt}\bigg\}, \\
	\tE Z(j,j'_1,\ldots, j_k') \lesssim \bar{\eta}RM^3\sqrt{\frac{d}{n|S|}} + \bar{\eta}[RM^3+RM\log(Rc_w^{-1})]\sqrt{\frac{1}{n}},
\end{align}
leading to
\begin{equation}
	\tP\bigg(Z(j,j'_1,\ldots, j_k') \geq \bar{\eta}RM^3\sqrt{\frac{d}{n}} + \bar{\eta}[RM^3+RM\log(Rc_w^{-1})]\sqrt{\frac{1}{n}} + t\bigg) \lesssim \exp\bigg\{-\frac{Cn|S|t^2}{M^2\bar{\eta}^2 +\bar{\eta}Mt}\bigg\}.
\end{equation}
Therefore
\begin{align}
	&\tP\bigg(\max_{j'_1,\ldots, j_k'=1:N'}\max_{j=1:N}Z(j,j'_1,\ldots, j_k') \geq C\bar{\eta}RM^3\sqrt{\frac{d}{n}} + C[RM^3+RM\log(Rc_w^{-1})]\log(Rc_w^{-1})\sqrt{\frac{1}{n}} + t\bigg) \\
	&\quad\lesssim N(N')^K\exp\bigg\{-\frac{Cn|S|t^2}{M^2\bar{\eta}^2 +\bar{\eta}Mt}\bigg\},
\end{align}
which implies that
\begin{equation}
	Z\leq \max_{j'_1,\ldots, j_k'=1:N'}\max_{j=1:N}Z(j,j'_1,\ldots, j_k') \lesssim \bar{\eta}RM^3\sqrt{\frac{d}{n}} + \bar{\eta}[RM^3+RM\log(Rc_w^{-1})]\sqrt{\frac{1}{n}} + \bar{\eta}M\sqrt{\frac{\log(1/\delta)}{n|S|}},
\end{equation}
w.p. at least $1-\delta$. Similarly,
\begin{align}
	&\sup_{\substack{\norm{w_r - \wks{k}_r} \leq \frac{c_w}{2R} \\ \twonorm{\btheta_r - \bthetaks{k}_r} \leq r_{\btheta}^* \\ 0 < \etak{k}_r \leq \bar{\eta}}} \frac{1}{n|S|}\Bigg\|\sum_{k \in S}\etak{k}_r\cdot \sum_{i=1}^n\big(\gamma^{(k)}_{\btheta, \bw}(\bxk{k}_i)\btheta_r - \tE[\gamma^{(k)}_{\btheta, \bw}(\bxk{k})\btheta_r]\big)\Bigg\|_2 \\
	&\lesssim \bar{\eta}RM^3\sqrt{\frac{d}{n}} + \bar{\eta}[RM^3+RM\log(Rc_w^{-1})]\sqrt{\frac{1}{n}} + \bar{\eta}M\sqrt{\frac{\log(1/\delta)}{n|S|}},
\end{align}
w.p. at least $1-\delta$. Hence
\begin{equation}
	\mathcal{E}_2(n, |S|, \delta) \asymp  RM^3\sqrt{\frac{d}{n}} + [RM^3+RM\log(Rc_w^{-1})]\sqrt{\frac{1}{n}} + M\sqrt{\frac{\log(1/\delta)}{n|S|}}.
\end{equation}

\subsection{Proof of Proposition \ref{prop: mor rj appendix}}
Recall that 
\begin{align}
	A_t &= \bigg[9\tilde{\kappa}_0\Big(\frac{\tilde{\kappa}_0}{119}\Big)^{t-1} + \frac{118}{119}(t-1)\tilde{\kappa}_0^{t-1}\bigg](r_w^* +r_{\btheta}^*) + \frac{1}{1-\tilde{\kappa}_0/119}\bar{\eta}\mathcal{E}_2\Big(n, |S|, \frac{\delta}{3R}\Big) \\
	&\quad+ \frac{18}{1-\tilde{\kappa}_0/119} \min\bigg\{3h, \frac{6}{1-\tilde{\kappa}_0}\Big[\mathcal{W}\Big(n, \frac{\delta}{3RK}, r_{\btheta}^*\Big) + 2\bar{\eta}\mathcal{E}_1\Big(n, \frac{\delta}{3R}\Big)\Big]\bigg\}\\
	&\quad + \frac{30}{(1-\tilde{\kappa}_0)(1-\tilde{\kappa}_0/119)}\epsilon\bigg[\mathcal{W}\Big(n, \frac{\delta}{3RK}, r_{\btheta}^*\Big) + 2\bar{\eta}\mathcal{E}_1\Big(n, \frac{\delta}{3R}\Big)\bigg],
\end{align}
and
\begin{equation}
	A_{t} + \frac{18}{1-\tilde{\kappa}_0/119}\mathcal{W}\Big(n, \frac{\delta}{3RK}, r_{\btheta, t}^*\Big) = r_{\btheta, t+1}^*,
\end{equation}
for $t \geq 1$ with $r_{\btheta, 1}^* \coloneqq r_{\btheta}^*$.

By Assumption \ref{asmp: mor appendix}.(\rom{4}), there exists $\wtkappa_0' \in (0,1)$ such that $CRM\sqrt{\frac{p}{n}} \leq \wtkappa_0'$ with a large $C$. Hence by plugging in the explicit rates obtained in Proposition \ref{prop: mor appendix},
\begin{align}
	r_{\btheta, t+1}^* &\leq \wtkappa_0'r_{\btheta, t}^* + Ct(\wtkappa_0)^{t-1}(r_w^* \vee r_{\btheta}^*) + C\bar{\eta}RM^3\sqrt{\frac{d}{n|S|}} + C[(\bar{\eta}M)\vee 1][RM^2 + R\log(Rc_w^{-1})]\sqrt{\frac{1}{n}} \\
	&\quad + C[(\bar{\eta}M)\vee 1]\sqrt{\frac{\log(RK/\delta)}{n}} + C\min\bigg\{h, \bar{\eta}RM^3\sqrt{\frac{d}{n}}\bigg\} + \epsilon\bar{\eta}RM^2[(\bar{\eta}M)\vee 1]\sqrt{\frac{d}{n}},
\end{align}
implying that
\begin{align}
	r_{\btheta, T}^* &\lesssim (\wtkappa_0')^{T-1}r_{\btheta}^* + T^2(\wtkappa_0')^{T-1}(r_w^* \vee r_{\btheta}^*) + \bar{\eta}RM^3\sqrt{\frac{d}{n|S|}} + [(\bar{\eta}M)\vee 1][RM^2 + R\log(Rc_w^{-1})]\sqrt{\frac{1}{n}} \\
	&\quad + [(\bar{\eta}M)\vee 1]\sqrt{\frac{\log(RK/\delta)}{n}} + \min\bigg\{h, \bar{\eta}RM^3\sqrt{\frac{d}{n}}\bigg\} + \epsilon\bar{\eta}RM^2[(\bar{\eta}M)\vee 1]\sqrt{\frac{d}{n}} \\
	&\lesssim T^2(\wtkappa_0 \vee \wtkappa_0')^{T-1}(r_w^* \vee r_{\btheta}^*) + \bar{\eta}RM^3\sqrt{\frac{d}{n|S|}} + [(\bar{\eta}M)\vee 1][RM^2 + R\log(Rc_w^{-1})]\sqrt{\frac{1}{n}} \\
	&\quad + [(\bar{\eta}M)\vee 1]\sqrt{\frac{\log(RK/\delta)}{n}} + \min\bigg\{h, \bar{\eta}RM^3\sqrt{\frac{d}{n}}\bigg\} + \epsilon\bar{\eta}RM^2[(\bar{\eta}M)\vee 1]\sqrt{\frac{d}{n}} \\
	&\lesssim T^2(\wtkappa_0 \vee \wtkappa_0')^{T-1}(r_w^* \vee r_{\btheta}^*) + R^2M^3c_w^{-1}\sqrt{\frac{d}{n|S|}} + R^2Mc_w^{-1}[M^2 + \log(Rc_w^{-1})]\sqrt{\frac{1}{n}} \\
	&\quad + MRc_w^{-1}\sqrt{\frac{\log(RK/\delta)}{n}} + \min\bigg\{h, R^2M^3c_w^{-1}\sqrt{\frac{d}{n}}\bigg\} + \epsilon RM^3c_w^{-1}\sqrt{\frac{d}{n}},
\end{align}
where $\kappa_0 = 119\sqrt{\frac{3C_b}{1+2C_b}+ CR^3c_w^{-2}\frac{(\log \Delta)^{3/2}}{\Delta}} + CR^3c_w^{-2}C_b + CR^4c_w^{-2}\frac{1}{\Delta} +  \tilde{\kappa}_0'$, and $\tilde{\kappa}_0'$ satisfies $1> \tilde{\kappa}_0' > CMR\sqrt{\frac{d}{n}}$ for some $C > 0$.

\subsection{Proof of Corollary \ref{cor: mor appendix}}
By the rate of $\mathcal{W}(n, \frac{\delta}{3RK}, r_{\btheta, T}^*)$ in Proposition \ref{prop: mor appendix} and the upper bound of $r_{\btheta, T}^*$ in Proposition \ref{prop: mor rj appendix},
\begin{align}
	\mathcal{W}\Big(n, \frac{\delta}{3RK}, r_{\btheta, T}^* \Big) &\asymp RM r_{\btheta, T}^*\sqrt{\frac{d}{n}} + [RM^2+R\log(Rc_w^{-1})]\sqrt{\frac{1}{n}} + \sqrt{\frac{\log(RK/\delta)}{n}} \\
	&\lesssim T^2(\wtkappa_0 \vee \wtkappa_0')^{T-1}(r_w^* \vee r_{\btheta}^*) + R^2M^3c_w^{-1}\sqrt{\frac{d}{n|S|}} + R^2Mc_w^{-1}[M^2 + \log(Rc_w^{-1})]\sqrt{\frac{1}{n}} \\
	&\quad + MRc_w^{-1}\sqrt{\frac{\log(RK/\delta)}{n}} + \min\bigg\{h, R^2M^3c_w^{-1}\sqrt{\frac{d}{n}}\bigg\} + \epsilon RM^3c_w^{-1}\sqrt{\frac{d}{n}}.
\end{align}
Applying Theorem \ref{thm: generic appendix}, we have
\begin{align}
	&\max_{k \in S}\max_{r \in [R]}(\twonorm{\hthetakt{k}{T}_r - \bthetaks{k}_r} + \norm{\hwkt{k}{T}_r - \wks{k}_r}) \\
	&\leq 20T(\wtkappa_0)^{T-1}(r_w^* \vee r_{\btheta}^*) + \bigg[\frac{119}{15}\wtkappa_0(\wtkappa_0/119)^{T-1} + \frac{118}{119}(T-1)(\wtkappa_0)^T\bigg](r_w^* + r_{\btheta}^*)\\
	&\quad + \frac{1}{1-\kappa_0}\Big[\bar{\eta}\mathcal{E}_2\Big(n, |S|, \frac{\delta}{3R}\Big) + \mathcal{W}\Big(n, \frac{\delta}{3RK}, r_{\btheta, J}^*\Big)\Big] \\
	&\quad + \frac{18}{1-\wtkappa_0/119}\cdot\min\bigg\{3h, \frac{6}{1-\wtkappa_0}\Big[\mathcal{W}\Big(n, \frac{\delta}{3RK}, r_{\btheta}^*\Big) + 2\bar{\eta}\mathcal{E}_1\Big(n, \frac{\delta}{3RK}\Big)\Big]\bigg\} \\
	&\quad + \frac{30}{(1-\wtkappa_0)(1-\wtkappa_0/119)}\epsilon\cdot  \Big[\mathcal{W}\Big(n, \frac{\delta}{3RK}, r_{\btheta}^*\Big) + 2\bar{\eta}\mathcal{E}_1\Big(n, \frac{\delta}{3RK}\Big)\Big] \\
	&\leq T^2(\wtkappa_0 \vee \wtkappa_0')^{T-1}(r_w^* \vee r_{\btheta}^*) + R^2M^3c_w^{-1}\sqrt{\frac{d}{n|S|}} + R^2Mc_w^{-1}[M^2 + \log(Rc_w^{-1})]\sqrt{\frac{1}{n}} \\
	&\quad + MRc_w^{-1}\sqrt{\frac{\log(RK/\delta)}{n}} + \min\bigg\{h, R^2M^3c_w^{-1}\sqrt{\frac{d}{n}}\bigg\} + \epsilon RM^3c_w^{-1}\sqrt{\frac{d}{n}}. \label{eq: MOR cor 1}
\end{align}
Note that conditioned on the event $\mathcal{V}$ defined in the proof of Theorem \ref{thm: generic appendix}, 
\begin{equation}
	\etak{k}_r = (1+2C_b)^{-1}(\hwkt{k}{0}_r)^{-1} \lesssim Rc_w^{-1},
\end{equation}
for all $k \in S$ and $r \in [R]$. Plugging it in equation \eqref{eq: MOR cor 1} implies the desired upper bound in Corollary 2.

\subsection{Proof of Theorem \ref{thm: exhaustive search appendix}}
Recall that our best permutation $\pi_k^* \in \mathcal{P}^R$ on task $k$ can be up to a permutation on $[K]$. WLOG, consider $\pi_k^*$ satisfying that $\pi_k^*(r) = $ ``the majority class" $\tilde{r}$ if $\#\{k \in S: \pi_k(r) = \tilde{r}\} > \frac{1}{2}|S|$, for all $k \in S$. Define ``the majority class" of $\{\pi_k(r)\}_{k \in S}$ as the $\tilde{r} \in [R]$ which satisfies $\#\{k \in S: \pi_k(r) = \tilde{r}\} \geq \max_{r' \neq \tilde{r}}\#\{k \in S: \pi_k(r) = r'\}$. Denote the majority class of $\{\pi_k(r)\}_{k \in S}$ as $m_r \in [R]$ and $S_r = \{k \in S: \pi_k(r) = m_r\}$. WLOG, suppose $\bm{\pi}^* = \{\pi_k^*\}_{k=1}^K$ satisties $\pi_k^*(r) = r$ for any $r$ and $k \in S$. Consider any $\bm{\pi} = \{\pi_k\}_{k=1}^K$ with $\pi_k(r) = \pi_k^*(r)$ for all $k \in S^c$ and $\bm{\pi} \neq \bm{\pi}^*$. It suffices to show that $\text{score}(\bm{\pi}, K) > \text{score}(\bm{\pi}^*, K)$.

For convenience, denote $\xi = \max_{k \in S}\min_{\pi_k}\max_{r \in [R]}\twonorm{\htheta_{\pi_k(r)}^{(k)[0]} - \bthetaks{k}_r}$. We have
\begin{align}
	\text{score}(\bm{\pi}, K) - \text{score}(\bm{\pi}^*, K) &= \underbrace{\sum_{k \neq k' \in S}\sum_{r=1}^R \twonorm{\htheta_{\pi_k(r)}^{(k)[0]} - \htheta_{\pi_{k'}(r)}^{(k')[0]}}}_{[1]} + \underbrace{2\sum_{k \in S, k' \in S^c}\sum_{r=1}^R \twonorm{\htheta_{\pi_k(r)}^{(k)[0]} - \htheta_{\pi_{k'}(r)}^{(k')[0]}}}_{[2]} \\
	&\quad - \underbrace{\sum_{k \neq k' \in S}\sum_{r=1}^R \twonorm{\htheta_r^{(k)[0]} - \htheta_r^{(k')[0]}}}_{[1]'} - \underbrace{2\sum_{k \neq k' \in S}\sum_{r=1}^R \twonorm{\htheta_r^{(k)[0]} - \htheta_{\pi_{k'}(r)}^{(k')[0]}}}_{[2]'}.
\end{align}
Note that
\begin{align}
	[1] - [1]' &= \sum_{k \neq k' \in S}\sum_{r:\pi_k(r) \neq \pi_{k'}(r)} \Big(\twonorm{\htheta_{\pi_k(r)}^{(k)[0]} - \htheta_{\pi_{k'}(r)}^{(k')[0]}} - \twonorm{\htheta_r^{(k)[0]} - \htheta_r^{(k')[0]}}\Big) \\
	&\geq \sum_{k \neq k' \in S}\sum_{r:\pi_k(r) \neq \pi_{k'}(r)} \Big(\twonorm{\bthetaks{k}_{\pi_k(r)} - \bthetaks{k'}_{\pi_{k'}(r)}} - \twonorm{\bthetaks{k}_r - \bthetaks{k'}_r}-4\xi\Big) \\
	&\geq \sum_{k \neq k' \in S}\sum_{r:\pi_k(r) \neq \pi_{k'}(r)} \Big(\twonorm{\bthetaks{k}_{\pi_k(r)} - \bthetaks{k}_{\pi_{k'}(r)}} - \twonorm{\bthetaks{k}_{\pi_{k'}(r)}- \bthetaks{k'}_{\pi_{k'}(r)}} - \twonorm{\bthetaks{k}_r - \bthetaks{k'}_r}-4\xi\Big) \\
	&\geq \sum_{k \neq k' \in S}\sum_{r:\pi_k(r) \neq \pi_{k'}(r)}(\Delta - 2h-4\xi) \\
	&= \sum_{r=1}^R \sum_{k \neq k' \in S, \pi_k(r) \neq \pi_{k'}(r)}(\Delta - 2h-4\xi).
\end{align}
For $r$ with $|S_r| > \frac{1}{2}|S|$:
\begin{equation}
	\sum_{k \neq k': \pi_k(r) \neq \pi_{k'}(r)}(\Delta -2h-4\xi) \geq |S_r|(|S| - |S_r|)(\Delta -2h-4\xi).
\end{equation}
For $r$ with $|S_r| \leq \frac{1}{2}|S|$: denote $\#\{K \in S: \pi_k(r) = r'\}$ as $v_{r'}$, where $\sum_{r'=1}^R v_{r'} = |S|$ and $|v_{r'}| \leq \frac{1}{2}|S|$ for all $r'$. Then
\begin{align}
	\sum_{k \neq k': \pi_k(r) \neq \pi_{k'}(r)} (\Delta -2h-4\xi) &\geq (\Delta -2h-4\xi)\bigg[|S|(|S| - 1) - \sum_{r'=1}^R v_{r'}(v_{r'}-1)\bigg] \\
	&= (\Delta -2h-4\xi)\bigg[|S|^2 - \sum_{r'=1}^R v_{r'}^2\bigg] \\
	&\geq (\Delta -2h-4\xi)\bigg[|S|^2 - 2\cdot \Big(\frac{1}{2}|S|\Big)^2\bigg] \\
	&\geq (\Delta -2h-4\xi) \cdot \frac{1}{2}|S|^2.
\end{align}
Hence
\begin{equation}
	[1] - [1]' \geq (\Delta -2h-4\xi)\cdot \Bigg[\sum_{r: |S_r| > |S|/2} |S_r|(|S| - |S_r|) + \sum_{r: |S_r| \leq |S|/2}\frac{1}{2}|S|^2\Bigg].
\end{equation}
And
\begin{align}
	[2] - [2]' &\geq -2\sum_{k \in S, k' \in S^c}\sum_{r=1}^R \twonorm{\hthetakt{k}{0}_{\pi_k(r)} - \hthetakt{k}{0}_r} \\
	&\geq -2|S^c|\sum_{k \in S}\sum_{r: \pi_k(r) \neq r} (h + 2\xi) \\
	&\geq -2|S^c|\Bigg[\sum_{r: |S_r| > |S|/2}\sum_{k \in S, \pi_k(r) \neq r}(h+2\xi) + \sum_{r: |S_r| \leq |S|/2}\sum_{k \in S, \pi_k(r) \neq r}(h+2\xi)\Bigg] \\
	&\geq -2|S^c|(h+2\xi)\Bigg[\sum_{r: |S_r| > |S|/2}(|S| - |S_r|) + \sum_{r: |S_r| \leq |S|/2}|S|\Bigg].
\end{align}
Therefore,
\begin{align}
	\text{score}(\bm{\pi}) - \text{score}(\bm{\pi}^*) &\geq \sum_{r:|S_r| > |S|/2}(|S| - |S_r|)[|S_r|(\Delta - 2h - 4\xi) - 2|S^c|(h+2\xi)] \\
	&\quad + \sum_{r:|S_r| \leq |S|/2}\bigg[\frac{1}{2}|S|^2(\Delta - 2h - 4\xi) - 2|S^c||S|(h + 2\xi)\bigg] \\
	&\geq \sum_{r:|S_r| > |S|/2}(|S| - |S_r|)\bigg[\frac{1}{2}|S|\Delta - h(|S| + 2|S^c|) - \xi(2|S| + 4|S^c|)\bigg] \\
	&\quad + \sum_{r:|S_r| \leq |S|/2}|S|\bigg[\frac{1}{2}|S|(\Delta - 2h - 4\xi) - 2|S^c|(h + 2\xi)\bigg] \\
	&\geq \sum_{r:|S_r| > |S|/2}(|S| - |S_r|)\cdot \frac{1}{2}|S|\bigg[\Delta - h\bigg(2+4\cdot \frac{|S^c|}{|S|}\bigg) - \xi\bigg(4+8\cdot \frac{|S^c|}{|S|}\bigg)\bigg] \\
	&\quad + \sum_{r:|S_r| \leq |S|/2}\frac{1}{2}|S|^2 \bigg[\Delta - h\bigg(2+4\cdot \frac{|S^c|}{|S|}\bigg) - \xi\bigg(4+8\cdot \frac{|S^c|}{|S|}\bigg)\bigg] \\
	&> 0,
\end{align}
which completes the proof, where in the last inequality we used the fact that $|S^c|/|S| \leq \frac{\epsilon}{1-\epsilon}$.

\subsection{Proof of Theorem \ref{thm: stepwise search appendix}}
Consider the $\widetilde{K}$-th round where $\widetilde{K} \in S$. WLOG, suppose $\iota$ is the identity permutation on $[R]$. Denote $\widetilde{S} = [\widetilde{K}] \cap S$, $\widetilde{S}^c = [\widetilde{K}] \cap S^c$, hence $[K] = \widetilde{S} \cup \widetilde{S}^c$. WLOG, suppose $\pi_1 = \pi_2 = \ldots = \pi_{\widetilde{K}-1}$ are the identity permutations on $[R]$. Denote $\bm{\pi} = \{\pi_k\}_{k=1}^{\widetilde{K}-1} \cup \pi_{\widetilde{K}}$ and $\widetilde{\bm{\pi}} = \{\pi_k\}_{k=1}^{\widetilde{K}-1} \cup \widetilde{\pi}_{\widetilde{K}}$, where $\widetilde{\pi}_{\widetilde{K}}$ is the identity permutation on $[R]$ and $\pi_{\widetilde{K}}$ can be any non-identity permutation. We claim that it suffices to show that $\text{score}(\bm{\pi}) > \text{score}(\widetilde{\bm{\pi}})$ for any $\widetilde{K} \geq K_0$, because if this is the case, then $\widetilde{\pi}_{\widetilde{K}}$ will be chosen in the  $\widetilde{K}$-th round of the ``for" loop. Hence all chosen permutations in $\widetilde{S}$ have the same alignment. By induction, the output permutations from ``Permutation Alignment Algorithm 2 (Stepwise search)" are identity permutations on $[R]$ among tasks in $S$, which completes our proof. In the remaining part of this proof, we show $\text{score}(\bm{\pi}) > \text{score}(\widetilde{\bm{\pi}})$ for any $\widetilde{K} \geq K_0$.

In fact,
\begin{align}
	\text{score}(\bm{\pi}) - \text{score}(\widetilde{\bm{\pi}}) &= \underbrace{\sum_{k \in \widetilde{S}}\sum_{r: \pi_{\widetilde{K}}(r) \neq r}\twonorm{\hthetakt{\widetilde{K}}{0}_{\pi_{\widetilde{K}}(r)} - \hthetakt{k}{0}_r}}_{[1]} + \underbrace{\sum_{k \in \widetilde{S}^c}\sum_{r: \pi_{\widetilde{K}}(r) \neq r}\twonorm{\hthetakt{\widetilde{K}}{0}_{\pi_{\widetilde{K}}(r)} - \hthetakt{k}{0}_r}}_{[2]}  \\
	&\quad - \underbrace{\sum_{k \in \widetilde{S}}\sum_{r: \pi_{\widetilde{K}}(r) \neq r}\twonorm{\hthetakt{\widetilde{K}}{0}_r - \hthetakt{k}{0}_r}}_{[1]'} - \underbrace{\sum_{k \in \widetilde{S}^c}\sum_{r: \pi_{\widetilde{K}}(r) \neq r}\twonorm{\hthetakt{\widetilde{K}}{0}_r - \hthetakt{k}{0}_r}}_{[2]'}.
\end{align}
And
\begin{align}
	[1] - [1]' &= \sum_{k \in \widetilde{S}}\sum_{r: \pi_{\widetilde{K}}(r) \neq r}\Big(\twonorm{\hthetakt{\widetilde{K}}{0}_{\pi_{\widetilde{K}}(r)} - \hthetakt{k}{0}_r} - \twonorm{\hthetakt{\widetilde{K}}{0}_r - \hthetakt{k}{0}_r}\Big) \\
	&\geq \sum_{k \in \widetilde{S}}\sum_{r: \pi_{\widetilde{K}}(r) \neq r}\Big(\twonorm{\hthetakt{\widetilde{K}}{0}_{\pi_{\widetilde{K}}(r)} - \hthetakt{k}{0}_r} - \twonorm{\bthetaks{\widetilde{K}}_r - \bthetaks{k}_r} -2\xi\Big) \\
	&\geq \sum_{k \in \widetilde{S}}\sum_{r: \pi_{\widetilde{K}}(r) \neq r}\Big(\twonorm{\hthetakt{\widetilde{K}}{0}_{\pi_{\widetilde{K}}(r)} - \hthetakt{\widetilde{K}}{0}_r} -2h -4\xi\Big) \\
	&\geq \sum_{r: \pi_{\widetilde{K}}(r) \neq r}|\widetilde{S}|\Big(\twonorm{\hthetakt{\widetilde{K}}{0}_{\pi_{\widetilde{K}}(r)} - \hthetakt{\widetilde{K}}{0}_r} -2h -4\xi\Big),
\end{align}
and
\begin{align}
	[2] - [2]' &\geq \sum_{k \in \widetilde{S}^c}\sum_{r: \pi_{\widetilde{K}}(r) \neq r}\Big(\twonorm{\hthetakt{\widetilde{K}}{0}_{\pi_{\widetilde{K}}(r)} - \hthetakt{k}{0}_r} - \twonorm{\hthetakt{\widetilde{K}}{0}_r - \hthetakt{k}{0}_r}\Big) \\
	&\geq -\sum_{k \in \widetilde{S}^c}\sum_{r: \pi_{\widetilde{K}}(r) \neq r}\twonorm{\hthetakt{\widetilde{K}}{0}_{\pi_{\widetilde{K}}(r)} - \hthetakt{\widetilde{K}}{0}_r} \\
	&= -\sum_{r: \pi_{\widetilde{K}}(r) \neq r}|\widetilde{S}^c|\cdot \twonorm{\hthetakt{\widetilde{K}}{0}_{\pi_{\widetilde{K}}(r)} - \hthetakt{\widetilde{K}}{0}_r}.
\end{align}
Therefore,
\begin{align}
	\text{score}(\bm{\pi}) - \text{score}(\widetilde{\bm{\pi}}) &= [1] - [1]' + [2] - [2]' \\
	&\geq \sum_{r: \pi_{\widetilde{K}}(r) \neq r}\Big[(|\widetilde{S}|-|\widetilde{S}^c|)\twonorm{\hthetakt{\widetilde{K}}{0}_{\pi_{\widetilde{K}}(r)} - \hthetakt{\widetilde{K}}{0}_r} - 2|\widetilde{S}|h - 4|\widetilde{S}|\xi\Big] \\
	&\geq \sum_{r: \pi_{\widetilde{K}}(r) \neq r}\Big[(|\widetilde{S}|-|\widetilde{S}^c|)\Delta - 2|\widetilde{S}|h - (6|\widetilde{S}| - 2|\widetilde{S}^c|)\xi\Big]\\
	&= \sum_{r: \pi_{\widetilde{K}}(r) \neq r}\widetilde{K}\bigg[\bigg(2\frac{|\widetilde{S}|}{\widetilde{K}}-1\bigg)\Delta - 2\frac{|\widetilde{S}|}{\widetilde{K}}\cdot h - \bigg(8\frac{|\widetilde{S}|}{\widetilde{K}} - 2\bigg)\xi\bigg] \\
	&\geq \sum_{r: \pi_{\widetilde{K}}(r) \neq r}\widetilde{K}\cdot \bigg(\frac{K_0 - K\epsilon}{K_0 + K\epsilon}\cdot \Delta - 2h - 6\xi\bigg) \\
	&> 0,
\end{align}
where the second last inequality is due to the fact that $\frac{K_0}{K_0 + K\epsilon} \leq |\widetilde{S}/\widetilde{K}| \leq 1$.


\end{document}